%% file: main.tex
\documentclass{article}

\PassOptionsToPackage{numbers, compress}{natbib}

 \usepackage[eandd,final]{neurips_2026}

\usepackage[utf8]{inputenc} 
\usepackage[T1]{fontenc}    
\usepackage{hyperref}       
\usepackage{url}            
\usepackage{booktabs}       
\usepackage{amsfonts}       
\usepackage{nicefrac}       
\usepackage{microtype}      
\usepackage{xcolor}         
\usepackage{graphicx}
\usepackage{amsmath}
\usepackage{multirow}
\usepackage{subcaption}
\DeclareMathOperator*{\argmax}{arg\,max}

\usepackage{makecell}
\usepackage{pifont}
\newcommand{\cmark}{\ding{51}}%
\newcommand{\xmark}{\ding{55}}%
\usepackage{tikz}
\usepackage{pgfplots}
\pgfplotsset{compat=1.18}

\title{RA-ClipScore: Making Generative Model Evaluation More Interpretable}

\author{%
  Yifan Lu$^{1, 3}$ \quad
  Taras Kucherenko$^{2}$\thanks{Most of the work was performed while Taras Kucherenko was at Electronic Arts (EA).} \quad
  Hedvig Kjellström$^{1}$ \quad
  Judith Bütepage$^{3}$\thanks{This work was conducted as part of academic research at KTH Royal Institute of Technology. Taras Kucherenko and Judith Bütepage provided supervisory input only and were not involved in the implementation and conduction of the experiments.}
  \\
  $^{1}$Robotics, Perception and Learning, KTH Royal Institute of Technology \\
  $^{2}$KBLab, National Library of Sweden \\
  $^{3}$Nava, Electronic Arts (EA) \\
  \texttt{yifanlu@kth.se} \quad
  \texttt{taras.kucherenko@kb.se} \quad
  \texttt{hedvig@kth.se} \quad
  \texttt{jbutepage@ea.com}
}

\begin{document}

\maketitle

\input{text/abstract}  
\input{text/intro}
\input{text/related_works}
\input{text/method}
\input{text/experiments}
\input{text/conclusions}

\input{text/acknowledgments}



\bibliographystyle{plainnat}
\bibliography{main}

\input{text/suppl}

\clearpage
\newpage
\input{checklist.tex}

\end{document}

%% file: text/abstract.tex
\begin{abstract}
\vspace{-2mm}
Generative models can produce images nearly indistinguishable from real data, yet rigorous and interpretable evaluation remains challenging. Conventional metrics such as FID provide only scalar scores with limited diagnostic insight. Widely adopted CLIP-based metrics enable semantic evaluation beyond simple training class labels, but inherit limitations from CLIP’s training paradigm that restrict attribute-wise analysis. We propose RA-CLIPScore, a novel metric that mitigates these issues and extends CLIP-based evaluation to spatial distribution alignment, measuring whether generated objects adhere to the positional priors found in the training data. RA-CLIPScore introduces dual prompts to decouple competing attributes and leverages local patch tokens to capture fine-grained regional semantics. We evaluate image generative models on their ability to match both attribute and spatial distributions of the training data. Extensive experiments show that RA-CLIPScore provides more robust and interpretable evaluations than prior methods, particularly under distribution misalignment or partially irrelevant textual attributes. We further demonstrate how it reveals previously undocumented regional tendencies in generative models. Code is available at \url{https://github.com/wwwfan628/ra_clipscore}.
\end{abstract}

%% file: text/intro.tex
\section{Introduction}
\label{sec:intro}
\vspace{-2mm}
Generative models have advanced at an unprecedented pace in recent years. Architectures such as Variational Autoencoders (VAE)~\cite{kingma2013auto}, Generative Adversarial Networks (GAN)~\cite{karras2019style,karras2020analyzing,karras2021alias,sauer2021projected}, Diffusion Models (DM)~\cite{ho2020denoising,song2020denoising,nichol2021glide,Rombach2022ldm} and Normalizing Flows~\cite{rezende2015variational,dinh2016density} can now produce images nearly indistinguishable from real samples. However, even state-of-the-art generative models can produce unreliable or biased outputs, making rigorous and interpretable evaluation a critical challenge for ensuring the practical utility of synthetic data.

\begin{figure}[htbp]
\centering
\includegraphics[width=0.9\textwidth]{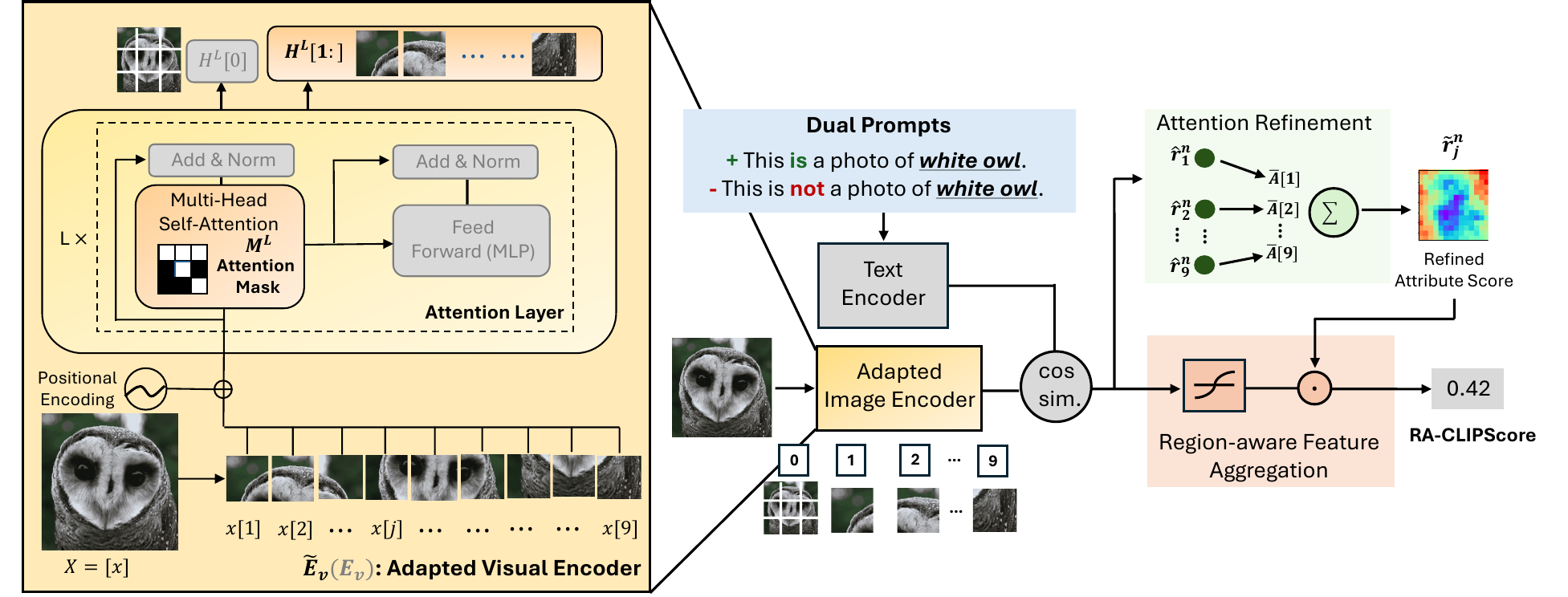}
\caption{\small{Our proposed RA-CLIPScore. We introduce \textit{dual prompts} and \textit{regional feature extraction} to improve interpretability of CLIP-based evaluation metrics while extending to spatial domain. Dual prompts decouple correlations among attributes, whereas regional feature extraction enhances fine-grained sensitivity through an adapted visual encoder, attention refinement, and region-aware feature aggregation. Grey blocks (left) denote components unchanged from the original visual encoder.}}
\label{fig:racs_pipeline}
\end{figure}

The evaluation of generative models in the image domain has attracted significant research interest. Classical metrics such as the Fréchet Inception Distance (FID)~\cite{heusel2017gans}, Kernel Inception Distance (KID)~\cite{sutherland2018demystifying} and precision/recall~\cite{kynkaanniemi2019improved,sajjadi2018assessing} quantify distributional dissimilarity or sample diversity. Despite their popularity, these metrics produce only scalar scores, offering limited interpretability and diagnostic insight. They cannot reveal whether performance differences arise from failures on implausible attributes (e.g. baby with beard)~\cite{kim2023attribute}, nor can they identify spatial biases such as generating objects in fixed poses or restricted regions. Understanding attribute and spatial failures can guide improved training strategies, such as targeted data resampling, and inform model selection for downstream applications. This motivates the need for semantically grounded metrics beyond statistics to explain why and where models differ. 

With the rise of large-scale vision-language pretraining, CLIP~\cite{radford2021learning} has emerged as a powerful foundation model and is increasingly used to replace Inception-V3~\cite{szegedy2016rethinking} as a feature extractor for traditional generative evaluation metrics, since Inception-V3 is agnostic to features unrelated to ImageNet~\cite{kynkaanniemi2022role,betzalel2022study,morozov2021self,parmar2022aliased}. Building on this trend, CLIPScore~\cite{hessel2021clipscore} and Heterogeneous CLIPScore (HCS)~\cite{kim2023attribute} leverage CLIP’s joint vision–language embedding space to measure semantic alignment between generated images and captions. However, both of them rely on CLIP’s original feature embedding. Revisiting CLIP’s training paradigm, we identify two core limitations that hinder fine-grained attribute-wise evaluation: (1) its softmax-based contrastive training introduces competition among textual categories, imposing a mutual exclusivity that limits the representation of overlapping attributes; (2) each image is encoded as a single global token without explicit regional tokens, thereby reducing sensitivity to fine-grained details.

To overcome these issues, we propose Region-and-Attribute-Aware CLIPScore (RA-CLIPScore). As shown in Figure \ref{fig:racs_pipeline}, RA-CLIPScore adopts modified dual prompts~\cite{sun2022dualcoop} to disentangle attributes by modeling both their presence and absence, effectively treating attribute detection as a multi-label recognition task to mitigate competition introduced by softmax training. It further leverages CLIP’s local patch tokens to capture fine-grained regional features through attention refinement and feature aggregation modules. By incorporating localized information, RA-CLIPScore extends CLIP-based evaluation to spatial distribution alignment, enabling analysis of whether generative models exhibit biases toward specific regions or poses. Detecting systematic spatial biases is critical to avoid spurious correlations in downstream systems. For example, if synthetic images are used to train a medical classifier, consistently generating pathological features (e.g. cancer cells) at the center of images may cause the model to learn this spatial bias instead of relevant biological patterns, failing to detect cells in other regions. Notably, RA-CLIPScore achieves these capabilities without additional training and with minimal computational overhead. Extensive experiments demonstrate that RA-CLIPScore provides more robust and interpretable attribute-wise evaluation than previous methods~\cite{hessel2021clipscore,kim2023attribute}, particularly in practical settings. It also captures spatial biases not addressed by prior metrics~\cite{heusel2017gans,kynkaanniemi2019improved,sajjadi2018assessing,sutherland2018demystifying}, offering additional insights into generative models.


Our contributions are summarized as follows: We introduce RA-CLIPScore, a robust metric that overcomes two key limitations of CLIP’s contrastive pretraining paradigm to enable precise attribute-wise and spatial evaluation. Central to this metric is a novel aggregation scheme that jointly exploits local and global tokens in a single forward pass, providing high-resolution spatial insights without the computational overhead or additional training required by prior work~\cite{sun2022dualcoop, lin2024tagclip}. With this efficient framework, we conduct the first systematic study of patch-level spatial-distributional evaluation of 2D image generators, uncovering distinct regional generation tendencie, such as localized object placement that represent a form of diversity currently ignored by standard benchmarks.

%% file: text/related_works.tex
\section{Related Work}
\label{sec:related_works}
\vspace{-2mm}
\noindent\textbf{Traditional Metrics.}
Evaluating generative models must account for multiple aspects including fidelity, diversity, and novelty. Fidelity metrics such as the Fréchet Inception Distance (FID)~\cite{heusel2017gans}, Inception Score (IS)~\cite{salimans2016improved}, and Kernel Inception Distance (KID)~\cite{sutherland2018demystifying} measure the similarity between generated and real images, typically via statistical distances such as Fréchet Distance or Maximum Mean Discrepancy (MMD). Diversity-oriented metrics, including precision/recall~\cite{kynkaanniemi2019improved,sajjadi2018assessing} and density/coverage~\cite{naeem2020reliable}, evaluate how comprehensively the generated samples span the underlying data distribution. To assess perceptual quality, Learned Perceptual Image Patch Similarity (LPIPS) \cite{zhang2018unreasonable} computes distances in deep feature space to mirror human visual judgment. Despite their broad utility, these metrics produce only scalar scores that capture isolated aspects of performance and offer limited diagnostic insight into why one model outperforms another. In contrast, RA-CLIPScore reveals human-understandable semantic and spatial distribution shifts instead of a black-box scalar score, improving explainability and diagnostic power.


\noindent\textbf{MLLM/VQA-based Evaluation.}
Recent methods leverage multimodal large language models (MLLMs) and visual question answering (VQA) to assess text-image alignment and compositional reasoning~\cite{li2024genai,huang2023t2i,hu2023tifa}. While offering stronger semantic reasoning and open-ended explanations, these approaches often incur substantial computational costs. As shown in Supp.~\ref{sec:sup-mat_mllm} Tab.~\ref{tab:sup-mat_mllm_cost}, RA-CLIPScore requires substantially fewer parameters and FLOPs. We further provide a qualitative comparison of their respective strengths and limitations in Supp.~\ref{sec:sup-mat_mllm} Tab.~\ref{tab:sup-mat_mllm_comparison}. Rather than replacing MLLM/VQA-based methods, RA-CLIPScore complements them by enabling efficient, reproducible, and spatially interpretable evaluation of large-scale image datasets.

\noindent\textbf{CLIPScore and Variants.}
\citet{hessel2021clipscore} introduced CLIPScore, a reference-free evaluation metric that leverages the pretrained CLIP~\cite{radford2021learning} to measure semantic alignment between images and captions. It computes the cosine similarity between CLIP embeddings of a generated image and its corresponding caption:

\vspace{-6.5mm}
\begin{equation}
\text{CLIPScore}(x,t)
= 100 \cdot \cos\!\big(\mathbf{E}_{v}(x), \mathbf{E}_{t}(t)\big),
\label{eq:clipscore}
\end{equation}

\vspace{-1.725mm}
\noindent where $\mathbf{E}_{v}$ and $\mathbf{E}_{t}$ denote CLIP’s image and text encoders, while $x$ and $t$ represent the input image and textual caption. While effective, CLIPScore provides only a global similarity score and lacks the granularity to identify which visual attributes contribute to misalignment or where these discrepancies occur. In contrast, our RA-CLIPScore enables finer-grained semantic and spatial analysis.

\citet{kim2023attribute} proposed Heterogeneous CLIPScore (HCS), extending CLIPScore for attribute-wise evaluation by introducing separate reference centers for image and text embeddings:

\vspace{-4mm}
\begin{equation}
\text{HCS}(x,t_i)
= 100 \cdot \cos\!\big(\mathbf{E}_v(x) - \mathbf{C}_{\text{img}},\, \mathbf{E}_t(t_i) - \mathbf{C}_{\text{text}}\big),
\label{eq:hcs}
\end{equation}

\vspace{-2mm}
\noindent where
$
\mathbf{C}_{\text{img}}
= \frac{1}{N_{\text{img}}}\sum_{i=1}^{N_{\text{img}}} \mathbf{E}_{v}(x_i)$ and 
$\mathbf{C}_{\text{text}}
= \frac{1}{N_{\text{text}}}\sum_{i=1}^{N_{\text{text}}} \mathbf{E}_{t}(t_i)$ represent the average embedding of the training images and textual attributes, respectively. Note that $t_i$ denotes an individual attribute, in contrast to $t$ in Eq.~\eqref{eq:clipscore}, which represents a holistic caption. 

Although HCS enables attribute-level evaluation, it remains constrained by the original CLIP training paradigm. Furthermore, $\mathbf{C}_{\text{img}}$ and $\mathbf{C}_{\text{text}}$ introduce contextual dependencies on both the training distribution and selected attribute set, making the metric sensitive to distributional shifts and vocabulary choices. Its effectiveness relies on the assumption that the chosen attributes adequately represent the image manifold. When this assumption is violated, HCS becomes unreliable. In Supp.\ref{sec:sup-mat_rethinking}, we theoretically prove that in a two-attribute setting, the metric collapses into a forced zero-sum relationship, where $\text{HCS}(x, t_p) = -\text{HCS}(x, t_q)$. Empirically, we also observe that removing a single attribute can lead to drastic changes. In contrast, RA-CLIPScore avoids these pitfalls by addressing CLIP’s limitations at the representation level, enabling robust attribute-wise evaluation and extending CLIP-based metrics to the spatial domain for the first time.

%% file: text/method.tex
\vspace{-2mm}
\section{Methodology}
\label{sec:method}


\vspace{-2mm}
To address the limitations, namely competition among textual categories introduced by softmax-based training and limited sensitivity to localized or region-specific cues~\cite{dosovitskiy2020image}, we propose RA-CLIPScore (Sec.\ref{subsec:racs}), introducing two key components: dual prompts and local patch tokens modules. Furthermore, we incorporate RA-CLIPScore into Single-Attribute Divergence (SaD), Pair-Attribute Divergence (PaD)~\cite{kim2023attribute} (Sec.\ref{subsec:sad_pad}) and Regional Single-Attribute Divergence (R-SaD) (Sec.\ref{subsec:r_sad}), to enable a quantitative analysis between generated and real datasets at both attribute and spatial levels.

\vspace{-2mm}
\subsection{Region-and-Attribute-Aware CLIPScore}
\label{subsec:racs}
\vspace{-2mm}
\subsubsection{Dual Prompts}
\label{subsubsec:dual_prompts}
\vspace{-2mm}
To mitigate limitations arising from CLIP’s softmax-based contrastive training, which enforces competition among attributes, we adapt dual prompts~\cite{sun2022dualcoop} to reformulate attribute evaluation as independent binary recognition tasks, thereby disentangling attribute correlations from representation level. The original dual-prompt strategy was designed for CLIP-based multi-class classification and requires learnable context tokens. Since our objective is evaluation rather than training, we instead use fixed prompts consistent with CLIP’s pretraining.

For each attribute $t_i$, we construct a pair of prompts:
\vspace{-1mm}
\begin{align}
\text{Prompt}_{t_i}^+ = \text{``This is a photo of } t_i\text{.''}\quad 
\text{Prompt}_{t_i}^- = \text{``This is a photo without } t_i\text{.''}
\label{eq:prompts}
\end{align}
Given an image $x^n$, where the superscript $n=1,\dots,N_{\text{img}}$ indexes images in a dataset, the coarse attribute score for its $j$-th visual token $x^n[j]$ is defined as
\vspace{-1mm}
\begin{align}
\hat{r}_j^n(t_i)
=
\sigma\!\Big(
    \langle &\mathbf{\tilde{E}}_v(x^n[j]),  \mathbf{E}_t(\text{Prompt}_{t_i}^+)\rangle, 
    \langle \mathbf{\tilde{E}}_v(x^n[j]), \mathbf{E}_t(\text{Prompt}_{t_i}^-)\rangle
\Big),
\label{eq:dual_prompt}
\end{align}

\vspace{-4mm}
where $\sigma(\cdot)$ denotes the softmax operation, $\mathbf{\tilde{E}}_v$ represents the adapted CLIP image encoder described below that integrates local patch information, and $\mathbf{E}_t$ denotes CLIP’s text encoder. This formulation eliminates direct competition between attributes, enabling each to be assessed independently as a binary decision, thereby yielding a more stable and interpretable evaluation of attributes. Although CLIP has known limitations in understanding negation~\cite{park2025know, nie2024out, sun2022dualcoop}, our dual-prompt design only requires a meaningful relative semantic direction between positive and counterfactual prompts, as empirically validated in Supp.~\ref{subsec:sup-mat_dualprompt}.


\vspace{-2mm}
\subsubsection{Regional Feature Extraction}
\label{subsubsec:local_patch}
\vspace{-2mm}
To clarify our adaptations, we first describe CLIP’s original visual encoder $\mathbf{E}_v$ (Figure~\ref{fig:racs_pipeline}) and briefly recall the attention mechanism. CLIP employs either a modified ResNet~\cite{he2016deep} with an attention-pooling head or a Vision Transformer (ViT)~\cite{dosovitskiy2020image} as its visual backbone. The encoder consists of multiple attention layers, each comprising a Multi-Head Self-Attention (MHSA) module followed by a feed-forward MLP. The global token $H^l[0]$ aggregates information from the entire image $x$, while the remaining tokens $H^l[j],\, j \ge 1$, correspond to local patches $x[j]$. After $L$ layers, the final global token $H^L[0]$ encodes the overall image semantics. The process is described mathematically in Eq.\eqref{eq:attn_ori}. For clarity, we illustrate how the final layer $L$ is derived from layer $L{-}1$: 

\vspace{-5mm}
{
\setlength{\thinmuskip}{2mu}
\setlength{\medmuskip}{2mu}
\setlength{\thickmuskip}{3mu}
\begin{align}
&\hat{H}^L = H^{L-1} + A^L \big(H^{L-1} W_V^L \big),\quad
H^L = \hat{H}^L + \mathrm{MLP}^L(\hat{H}^L),\quad
\mathbf{E}_v(X) = H^L[0]
\label{eq:attn_ori}
\end{align}
}

\vspace{-2mm}
\noindent where $A^L$ denotes the attention weight matrix, $W_V^L$ is value projection matrix. The output of CLIP’s visual encoder is $\mathbf{E}_v(X)$. Here $X=[x^1, \ldots, x^{N}]$ denotes a batch of input images, and $H^{L-1}$ represents the hidden features at layer $L-1$. 

CLIP only retains the \texttt{[CLS]} token $H^L[0]$, which aggregates global context. The remaining patch tokens $H^L[1:]$, which contain localized spatial information, are discarded. We now describe how our adapted encoder $\mathbf{\tilde{E}}_v$ incorporates these patch tokens to form region-sensitive representations without additional computational overhead, achieved within a single forward pass.

Prior work~\cite{ghiasi2022vision,lin2024tagclip,zhou2021deepvit} shows that spatial information is preserved in intermediate CLIP feature maps but largely diminishes in the final layer. The global \texttt{[CLS]} token primarily aggregates global information in the final layer.  Motivated by this observation, we utilize local visual features from layer $L\!-\!1$ while retaining the global \texttt{[CLS]} token after the final attention layer. To achieve this, local patch tokens are projected directly into CLIP’s unified vision–language embedding space, bypassing the final attention operation to preserve locality, as defined in Eq.\eqref{eq:attn_new}. These dense features are then combined with the global token from Eq.\eqref{eq:attn_ori} to form a fine-grained spatially aware representation.

\vspace{-5mm}
{
\setlength{\thinmuskip}{1mu}
\setlength{\medmuskip}{1mu}
\setlength{\thickmuskip}{2mu}
 \begin{equation}
 \hspace{-2.25mm}\hat{H}_{\text{dense}}^L = H^{L-1} + H^{L-1} W_V^L, \quad H_{\text{dense}}^L = \hat{H}_{\text{dense}}^L + \mathrm{MLP}^L(\hat{H}_{\text{dense}}^L), \quad
 \mathbf{\tilde{E}}_v(X) = [H^L[0],\, H_{\text{dense}}^L[1:]]
 \label{eq:attn_new}
 \end{equation}}
 
\vspace{-1mm}
In PyTorch~\cite{paszke2019pytorch}, $A^L$ is implemented as in Eq.\eqref{eq:attn_weight}, where $W_Q^L$, $W_K^L$ denote the query and key projection matrices, $d_k$ is the feature dimension, and $M^L$ is the attention mask that blocks interactions with specific positions. The above process can be realized by setting $M^L$ as shown in Eq.~\ref{eq:attn_weight}. The first row contains all $0$s, ensuring that tokens can still attend to the global \texttt{[CLS]} token, while assigning $-\inf$ forces the corresponding attention weights to zero, making the attention submatrix for local patch tokens effectively an identity matrix, i.e. bypassing the attention operation for local patch tokens.

\vspace{-5mm}
{
\setlength{\thinmuskip}{1mu}
\setlength{\medmuskip}{1mu}
\setlength{\thickmuskip}{1mu}
\begin{equation}
\hspace{-1.0em}
 A^L = \sigma\!\left(\frac{(H^{L-1} W_Q^L)(H^{L-1} W_K^L)^{\!\top}}{\sqrt{d_k}} + M^L\right),
 M^L=\begin{bmatrix}
0 & 0 &   \dots & 0\\
-\inf & 0 &   \dots & -\inf\\
-\inf & -\inf &   \dots & -\inf\\
-\inf & -\inf &   \dots & 0\\
\end{bmatrix}
\label{eq:attn_weight}
\end{equation}
}

\vspace{-2mm}
This modification preserves the global token after the final attention layer while retaining local patch tokens from layer $L-1$, enabling interpretable attribute evaluation at both global and regional scales without any additional computational overhead. In contrast, prior work either uses local patch tokens from the last layer despite reduced spatial information~\cite{sun2022dualcoop}, requires additional training~\cite{abdelfattah2023cdul}, or obtains global \texttt{[CLS]} embeddings and local patch tokens through separate forward passes~\cite{lin2024tagclip}, increasing complexity and computational cost.

\vspace{-2mm}
\subsubsection{Refinement via Attention}
\vspace{-2mm}
The initial patch-level scores from Eq.\eqref{eq:dual_prompt} may contain noise due to limited contextual information~\cite{xu2022multi,gao2021ts}, reducing their reliability for evaluation. We leverage the self-attention mechanism, which captures pairwise dependencies among local patch tokens, to refine these coarse scores:
\vspace{-1mm}
\begin{equation}
\tilde{r}_{j}^n(t_i)
= \frac{1}{|\psi|}\sum_{l \in \psi} A^l[j] \cdot \hat{r}_j^n(t_i), \quad \forall j,\, j \neq 0
\label{eq:regional_score}
\end{equation}

\vspace{-4mm}
\noindent where $A^l[j]$ denotes attention weights between the $j$-th patch and other patches within the attention layer $l, l\in\psi$. We define $\psi=\{1,2,\dots,L{-}1\}$ to include all attention layers except the final one, which primarily globalizes the \texttt{[CLS]} token $H^L[0]$. The refinement is applied only to local patch tokens ($j\ne 0$) that bypass the final attention layer, and not to the global \texttt{[CLS]} token. As $A^l[j]$ is available from the forward pass, this refinement introduces negligible computational overhead.

\vspace{-2mm}
\subsubsection{Regional Feature Aggregation}
\vspace{-2mm}
To obtain an overall assessment for a given attribute $t_i$, we aggregate the refined regional scores in Eq.\eqref{eq:regional_score} via a weighted summation. The weights are proportional to the similarity between the local patch tokens and the positive prompt from Eq.\eqref{eq:prompts}:

\vspace{-6mm}
\begin{align}
\mathrm{RA\text{-}}&\mathrm{CLIPScore}(x^n, t_i) =  
\sum_{j}
\sigma\!\left(
    \langle \mathbf{\tilde{E}}_v(x^n[j]),\, \mathbf{E}_t(\text{Prompt}^+)\rangle
\right) \cdot\, \tilde{r}_j^n(t_i),
\label{eq:raclipscore}
\end{align}

\vspace{-4mm}
\noindent where $\sigma(\cdot)$ denotes softmax normalization. This weighted aggregation emphasizes regions with a greater confidence in containing the target attribute while maintaining spatial completeness. We empirically validate our local feature extraction pipeline in Supp.~\ref{subsec:sup-mat_local_features}, demonstrating substantial improvements over the original CLIP patch tokens on the COCO segmentation task, which requires fine-grained spatial understanding.

\vspace{-2mm}
\subsection{Single-Attribute Divergence (SaD) and Pair-Attribute Divergence (PaD)}
\label{subsec:sad_pad}
\vspace{-2mm}

Discrepancies between the training and generated datasets are quantified using Kullback–Leibler (KL) divergence. Given a training dataset $\chi_{\text{train}} = \{x^1, \dots, x^N\}$, a generated dataset $\Upsilon_{\text{gen}} = \{y^1, \dots, y^M\}$, and an attribute list $\tau = \{t_1, \dots, t_C\}$, we compute RA-CLIPScore for all images in both $\chi_{\text{train}}$ and $\Upsilon_{\text{gen}}$ according to Eq.\eqref{eq:raclipscore}. 

For a single attribute $t$ or an attribute pair $t=(t_p,t_q)$, we obtain two sets of scores:

\vspace{-4mm}
\begin{equation}
\begin{aligned}
\mathcal{R}(t) = \{r^1(t), \dots, r^N(t)\}, \mathcal{R}'(t) =\{r'^1(t), \dots, r'^M(t)\}
\end{aligned}
\end{equation}

\vspace{-1mm}
\noindent where each element in $\mathcal{R}(t)$ and $\mathcal{R}'(t)$ denotes the RA-CLIPScore of a single image from $\chi_{\text{train}}$ or $\Upsilon_{\text{gen}}$ with respect to $t$:

\vspace{-6mm}
\begin{align}
\label{eq:rn}
    r^n(t) = \mathrm{RA\text{-}CLIPScore}(x^n, t), \quad 
    r'^m(t) = \mathrm{RA\text{-}CLIPScore}(y^m, t) 
\end{align}

\vspace{-3mm}

\noindent For an attribute pair $t=(t_p,t_q)$, RA-CLIPScore is computed independently for each attribute, without using pair-specific prompts. Specifically, each image $x^n$ is represented by a two-dimensional score vector $\mathrm{RA\text{-}CLIPScore}(x^n,t)=[\mathrm{RA\text{-}CLIPScore}(x^n,t_p),\mathrm{RA\text{-}CLIPScore}(x^n,t_q)]^\top$. We approximate the empirical score distributions $\mathcal{R}(t)$ and $\mathcal{R}'(t)$ with Gaussian densities, denoted as $p_{\chi_{\text{train}}}(\mathcal{R}(t_i))$ and $p_{\Upsilon_{\text{gen}}}(\mathcal{R}'(t_i))$. The divergence between the two distributions is then measured using KL divergence. The input $t$ can represent either a single attribute $t=t_i$ or a pair of attributes $t=(t_p,t_q)$. We define Single-attribute Divergence (SaD) in Eq.\eqref{eq:sad} to quantify distributional shifts along individual semantic dimensions. To detect implausible attribute combinations (e.g. \textit{baby} with \textit{beard}), we compute Pair-attribute Divergence (PaD) in Eq.\eqref{eq:pad} over the joint distribution of attribute pairs, capturing deviations in attribute co-occurrence patterns between generated and real data.

\vspace{-6mm}
{
\setlength{\thinmuskip}{1mu}
\setlength{\medmuskip}{1mu}
\setlength{\thickmuskip}{1mu}
\begin{align}
\mathrm{SaD}(t_i)
&=
\mathrm{KL}\!\Big(
p_{\Upsilon_{\text{gen}}}(\mathcal{R}'(t_i))
\,\Big\Vert\,  
p_{\chi_{\text{train}}}(\mathcal{R}(t_i))
\Big), \label{eq:sad} \\
\mathrm{PaD}(t_p, t_q)
&=
\mathrm{KL}\!\Big(
    p_{\Upsilon_{\text{gen}}}(\mathcal{R}'((t_p, t_q))
\,\Big\Vert\,  
p_{\chi_{\text{train}}}(\mathcal{R}((t_p, t_q))
\Big) \label{eq:pad}
\end{align}
}


\vspace{-4mm}
\subsection{Extension to Spatial Evaluation: Regional Single-Attribute Divergence (R-SaD)}
\label{subsec:r_sad}
\vspace{-2mm}
Existing metrics~\cite{kynkaanniemi2019improved,sajjadi2018assessing,naeem2020reliable} typically focus on appearance diversity (e.g. color, texture) or category diversity, but do not explicitly measure spatial diversity such as object position or layout variation. As synthetic data are increasingly used in downstream tasks, detecting spatial bias becomes critical. GenEval~\cite{ghosh2023geneval} evaluates prompt-spatial alignment while \citet{uchida2025objectness} focus on 3D scene evaluation, both differ fundamentally from patch-level spatial distribution evaluation for 2D image generative models. Building on Eq.\eqref{eq:regional_score}, we extend our evaluation to spatial domain to fill this gap, providing insights into where image generative models exhibit attribute-specific biases with negligible computational overhead. Regional Single-attribute Divergence (R-SaD) is formulated as:

\vspace{-6mm}
\begin{align}
\mathrm{R\text{-}SaD}_j(t_i)
&=
    \frac{1}{N}\sum_{n \in \chi_{\text{train}}}
    \tilde{r}_{j}^n(t_i)  -
    \frac{1}{M}\sum_{m \in \Upsilon_{\text{gen}}}
    \tilde{r}'_{j}{}^{m}(t_i),
\label{eq:regional_sad}
\end{align}

\vspace{-3mm}
where $j$ indexes spatial patch tokens.  
$\tilde{r}_{j}^n(t_i)$ and $\tilde{r}'_{j}{}^m(t_i)$ denote the refined regional scores from Eq.\eqref{eq:regional_score}, computed on the $j$-th patch for attribute $t_i$ in the real dataset $\chi_{\text{train}}$ and generated dataset $\Upsilon_{\text{gen}}$.

We adopt mean-difference measure, as it offers more interpretable insights. While KL divergence is always non-negative, mean difference can take both positive and negative values, allowing us to distinguish biases. A positive R-SaD value indicates that the attribute appears more prominently in a specific region of the real dataset, whereas a negative value reflects a higher occurrence of the attribute in the generated images within the corresponding spatial region.

%% file: text/experiments.tex
\vspace{-2mm}
\section{Experiments and Results}
\label{sec:experiments}
\vspace{-2mm}
We conduct experiments to evaluate effectiveness and robustness of our metric using real data in controlled settings (Sec.\ref{subsec:toy_exp}), enabling direct comparison with ground-truth annotations for sanity check. The contribution of each component is analyzed through an ablation study (Sec.\ref{subsec:ablation_study}), demonstrating that all components are necessary for stable and interpretable evaluation. Using SaD, PaD, and R-SaD, we assess generative image models and present quantitative and spatial evaluation results in Sec.\ref{subsec:evaluation_sota} and Sec.\ref{subsec:spatial_eval}. Finally, we briefly introduce a user study examining the correlation between human perception and our metric in Sec.\ref{subsec:userstudy}, with additional details provided in the Supp.\ref{subsec:sup-mat_userstudy}. All results reported in this section use $256 \times 256$ images and ViT-B/16~\cite{dosovitskiy2020image} as the CLIP backbone, experiments with modified ResNet-50~\cite{he2016deep} yield comparable results. All experiments are conducted on a single NVIDIA A100 GPU with 40 GB on-board HBM2 VRAM.

\vspace{-2mm}
\subsection{Toy Experiments}
\label{subsec:toy_exp}


\vspace{-2mm}
\subsubsection{CelebA Attribute Classification}
\label{subsubsec:toy_celeba_classification}
\vspace{-2mm}
To demonstrate that RA-CLIPScore provides improved robustness over HCS~\cite{kim2023attribute}, we reproduce the experimental setup of ~\cite{kim2023attribute} that aims to classify images from CelebA dataset~\cite{liu2015faceattributes} using predicted attribute scores. Please refer to \cite{kim2023attribute} for details on the setup. We report accuracy and F1 scores in Tab.\ref{tab:celeba_attribute_classification} under evaluation settings that incorporate irrelevant attributes alongside the relevant image–attribute pairs to assess robustness. Accuracy and F1 scores are reported on the relevant attributes. The complete lists of relevant and irrelevant attributes are provided in Supp.~\ref{subsec:sup-mat_celeba_attributes}.

\begin{table}[b]
\centering
\begin{minipage}[t]{0.48\linewidth}
\vspace{0pt}
\centering
\setlength{\tabcolsep}{1.5pt}
\renewcommand{\arraystretch}{0.9}
\begin{tabular}{lccc}
\toprule
\textbf{Method} & \textbf{Acc.} & \textbf{F1 (Macro)} & \textbf{F1 (Micro)} \\
\midrule
\multirow{1}{*}{HCS~\cite{kim2023attribute}}
 & 79.30 & 42.12 & 54.20 \\
\midrule
RA-CLIPScore & \textbf{79.75} & \textbf{45.03} & \textbf{55.20}\\
\midrule
CLIPScore~\cite{hessel2021clipscore} & 79.44 & 44.04 & 54.52\\
\bottomrule
\end{tabular}
\caption{\small{CelebA attribute classification results reported in Accuracy (Acc.) and F1 scores under evaluation settings containing irrelevant image–attribute pairs to assess robustness.}}
\vspace{-6pt}
\label{tab:celeba_attribute_classification}
\end{minipage}
\hfill
\begin{minipage}[t]{0.48\linewidth}
\vspace{0pt}
\centering
\setlength{\tabcolsep}{1.5pt}
\renewcommand{\arraystretch}{0.9}
\begin{tabular}{@{}lcccc@{}}
\toprule
\textbf{Setting} &
\makecell{\textbf{Dual}\\\textbf{Prompts}} &
\makecell{\textbf{Local}\\\textbf{Feat.}} &
\makecell{\textbf{Attn.}\\\textbf{Refine.}} &
\makecell{\textbf{Global}\\\textbf{Token Src.}} \\
\midrule
w/o DP      & \xmark & \cmark & \cmark & Layer $L$ \\
w/o LT      & \cmark & \xmark & \xmark & Layer $L$ \\
w/o AT      & \cmark & \cmark & \xmark & Layer $L$ \\
w/o GE      & \cmark & \cmark & \cmark & Layer $L\!-\!1$ \\
Ours (Full) & \cmark & \cmark & \cmark & Layer $L$ \\
\bottomrule
\end{tabular}
\captionof{table}{\small Ablation study configurations. Each variant removes one component at a time.}
\label{tab:ablation_study_settings}
\end{minipage}

\end{table}


RA-CLIPScore outperforms the baselines by enabling better attribute disentanglement and more fine-grained evaluation. In contrast, HCS exhibits degraded performance in this setting (see Sec.~\ref{sec:related_works} and Supp.~\ref{sec:sup-mat_rethinking} for further discussion), while CLIPScore achieves slightly lower performance due to its emphasis on global image-text alignment. These results highlight the robustness of RA-CLIPScore in the presence of semantic mismatches, which are common in real-world evaluation scenarios. We further extend the evaluation to action classification on the Stanford 40 Actions dataset~\cite{yao2011human}, using a similar experimental setup (Supp.~\ref{subsec:sup-mat_action}). Additionally, we investigate learnable prompt contexts using CoOp~\cite{zhou2022learning}, showing that supervised prompt optimization substantially improves attribute classification performance but introduces task-specific training requirements, unlike the training-free formulation of RA-CLIPScore (Supp.~\ref{subsec:sup-mat_coop}).

\begin{figure}[t]
    \centering
    \begin{minipage}[t]{0.4\textwidth}
       \centering
       \begin{subfigure}[t]{0.45\linewidth}
           \centering
           \includegraphics[width=\linewidth]{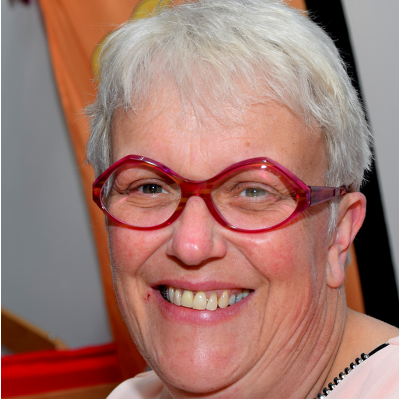}
           \caption{Original}
        \end{subfigure}
        \hfill
       \begin{subfigure}[t]{0.45\linewidth}
           \centering
           \includegraphics[width=\linewidth]{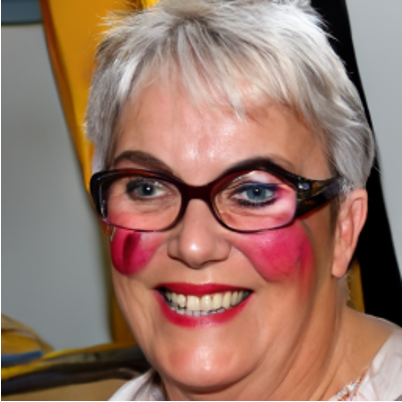}
           \caption{Makeup}
           \label{fig:qual_makeup}
       \end{subfigure}

        \vspace{0.6em}

       \begin{subfigure}[t]{0.45\linewidth}
           \centering
           \includegraphics[width=\linewidth]{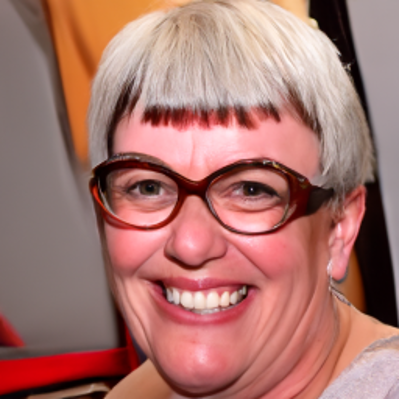}
           \caption{Bangs}
           \label{fig:qual_bangs}
       \end{subfigure}
        \hfill
       \begin{subfigure}[t]{0.45\linewidth}
           \centering
           \includegraphics[width=\linewidth]{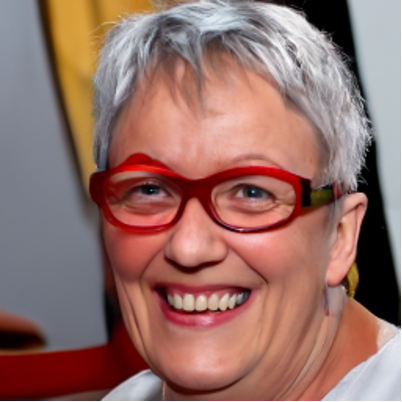}
           \caption{Person}
       \end{subfigure}
   \end{minipage}
    \begin{minipage}[t]{0.5\textwidth}
        \centering
        \begin{subfigure}[t]{0.49\linewidth}
            \centering
            \includegraphics[width=\linewidth]{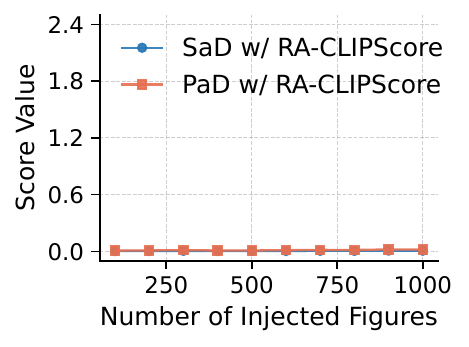}
            \caption{Original}
            \label{fig:quant_ffhq}
        \end{subfigure}
        \begin{subfigure}[t]{0.49\linewidth}
            \centering
            \includegraphics[width=\linewidth]{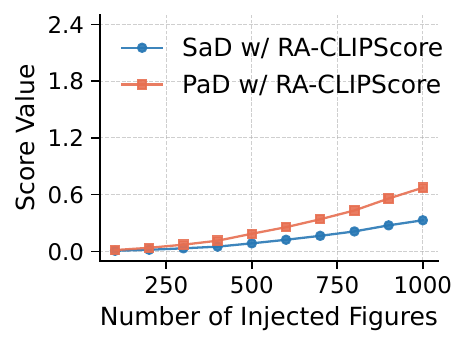}
            \caption{Makeup}
            \label{fig:quant_makeup}
        \end{subfigure}

        \vspace{0.4em}

        \begin{subfigure}[t]{0.49\linewidth}
            \centering
            \includegraphics[width=\linewidth]{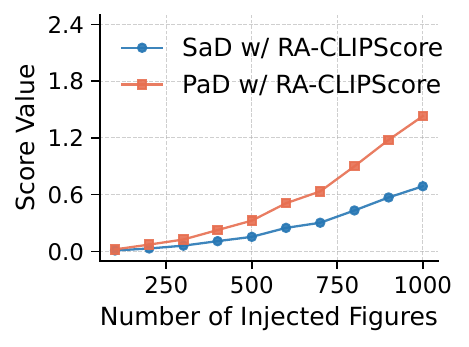}
            \caption{Bangs}
            \label{fig:quant_bangs}
        \end{subfigure}
        \begin{subfigure}[t]{0.49\linewidth}
            \centering
            \includegraphics[width=\linewidth]{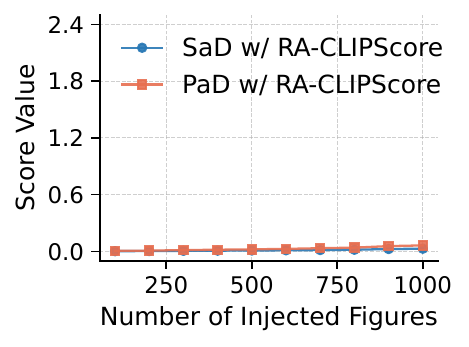}
            \caption{Person}
            \label{fig:quant_person}
        \end{subfigure}
    \end{minipage}

    \caption{
    \small{Left: an example FFHQ image and its DiffuseIT translations under different prompts. Right: SaD and PaD responses as the fraction of injected images increases.}
    }
    \label{fig:result_injection}
\end{figure}

\vspace{-2mm}
\subsubsection{Injection Experiment}
\label{subsubsec:toy_injection}
\vspace{-2mm}
To verify that SaD and PaD respond to attribute variations, we inject translated images into the original FFHQ dataset~\cite{karras2019style} and measure score changes as the injection ratio increases. The translated biased images are generated using DiffuseIT~\cite{kwon2022diffusion} with two prompts modifying specific attributes, namely \textit{makeup} and \textit{bangs}. As controls, we also inject images translated with the neutral prompt \textit{person} and reinsert unmodified FFHQ images. The list of used attributes is provided in Supp.\ref{subsec:sup-mat_injec_exp}.

We present translated samples and corresponding experiment results in Fig. \ref{fig:result_injection}. Both SaD and PaD increase monotonically when biased images (translated with \textit{makeup} or \textit{bangs}) are injected, while remaining nearly constant when unbiased images (translated with \textit{person} or original FFHQ samples) are injected. This confirms that SaD and PaD reliably detect attribute-specific changes without false positive responses under unbiased conditions. Similar trends are observed on manipulated COCO~\cite{lin2014microsoft} datasets, demonstrating effectiveness of our metrics beyond facial domain (details provided in Supp.\ref{subsec:sup-mat_injec_exp}).

\vspace{-2mm}
\subsubsection{Dataset Comparison}
\label{subsubsec:toy_dataset_compare}
\vspace{-2mm}
To further assess whether SaD and PaD capture joint attribute dependencies rather than marginal distributions, we construct three datasets with inverse combinations of \textit{eyeglasses} and \textit{gender}. In Datasets A and B, only men wear eyeglasses, while all women are without eyeglasses. Dataset C reverses this pattern, with only women wearing eyeglasses and all men without eyeglasses. We evaluate whether SaD and PaD detect this reversal in attribute coupling. For a fair comparison, SaD and PaD are computed for HCS following exactly the same evaluation pipeline as RA-CLIPScore. Specifically, we replace the RA-CLIPScore in Eq.~\eqref{eq:rn} with the HCS score computed by Eq.~\eqref{eq:hcs}, while keeping the Gaussian estimation and KL divergence computation unchanged. 

Results are summarized in Tab.\ref{tab:inverse_attr_coupling}. For SaD, we report results for \textit{eyeglasses} ($EG$) and other remaining attributes ($\neg EG$). Our metric shows a strong response (over $100\times$ increase) to changes in eyeglasses attribute alone, whereas HCS exhibits a weaker response ($\approx30\times$ increase) and also increases for unaffected attributes ($\approx2\times$ increase). This demonstrates improved attribute disentanglement and sensitivity. A similar trend is observed for PaD, where our metric more effectively detects changes in joint attribute distributions. We further visualize bar charts of the top attributes or attribute pairs with the highest SaD and PaD scores in Supp.\ref{subsec:sup-mat_dataset_compare}. For SaD, \textit{eyeglasses} is the most distinctive attribute. For PaD, the top responses correspond to the joint distributions of \textit{gender} and \textit{eyeglasses}, confirming that PaD effectively captures reversals in attribute dependencies.

We assess robustness to prompt wording under the same experimental setup using $4$ alternative prompt templates in Supp.~\ref{subsec:sup-mat_prompt_robustness}. Despite variations in the overall rankings, SaD and PaD consistently identify prominent attribute-level discrepancies, indicating robustness to prompt formulation.
\vspace{-3mm}
\subsection{Ablation Study}
\label{subsec:ablation_study}
\vspace{-3mm}

To assess the contribution of each component, we conduct an ablation study on $20$ CelebA attributes~\cite{liu2015faceattributes} on a mixed dataset of human faces from CelebA and furniture from~\cite{mukherjee_furniture_dataset}. We group these attributes into global (11), local (5), and other (4) categories according to their semantic characteristics. We evaluate four ablation variants by individually removing dual prompts (w/o DP), local tokens (w/o LT), global embeddings (w/o GE), or attention refinement (w/o AR). Experimental settings are summarized in Tab.~\ref{tab:ablation_study_settings}. The w/o DP variant removes the dual-prompt formulation and uses only the positive prompt for scoring. Specifically, Eq.~\eqref{eq:dual_prompt} becomes $\hat{r}_j^n(t_i)=\langle\mathbf{\tilde{E}}_v(x^n[j]),  \mathbf{E}_t(\text{Prompt}_{t_i}^+)\rangle$, while Eq.~\eqref{eq:regional_score}–\eqref{eq:raclipscore} remain unchanged. For each attribute, we compute scores separately for images where the attribute is present (positive) and absent (negative). Score distributions are estimated using Gaussian kernel density estimation (KDE) for each group. We then evaluate how well the scores separate positive and negative samples using Cohen’s d~\cite{cohen1988statistical}, where higher values indicate better separation (see Supp.\ref{subsec:sup-mat_ablation} for details).

\begin{figure}[b]
\centering

\begin{subfigure}{0.24\linewidth}
\centering
\includegraphics[width=\linewidth]{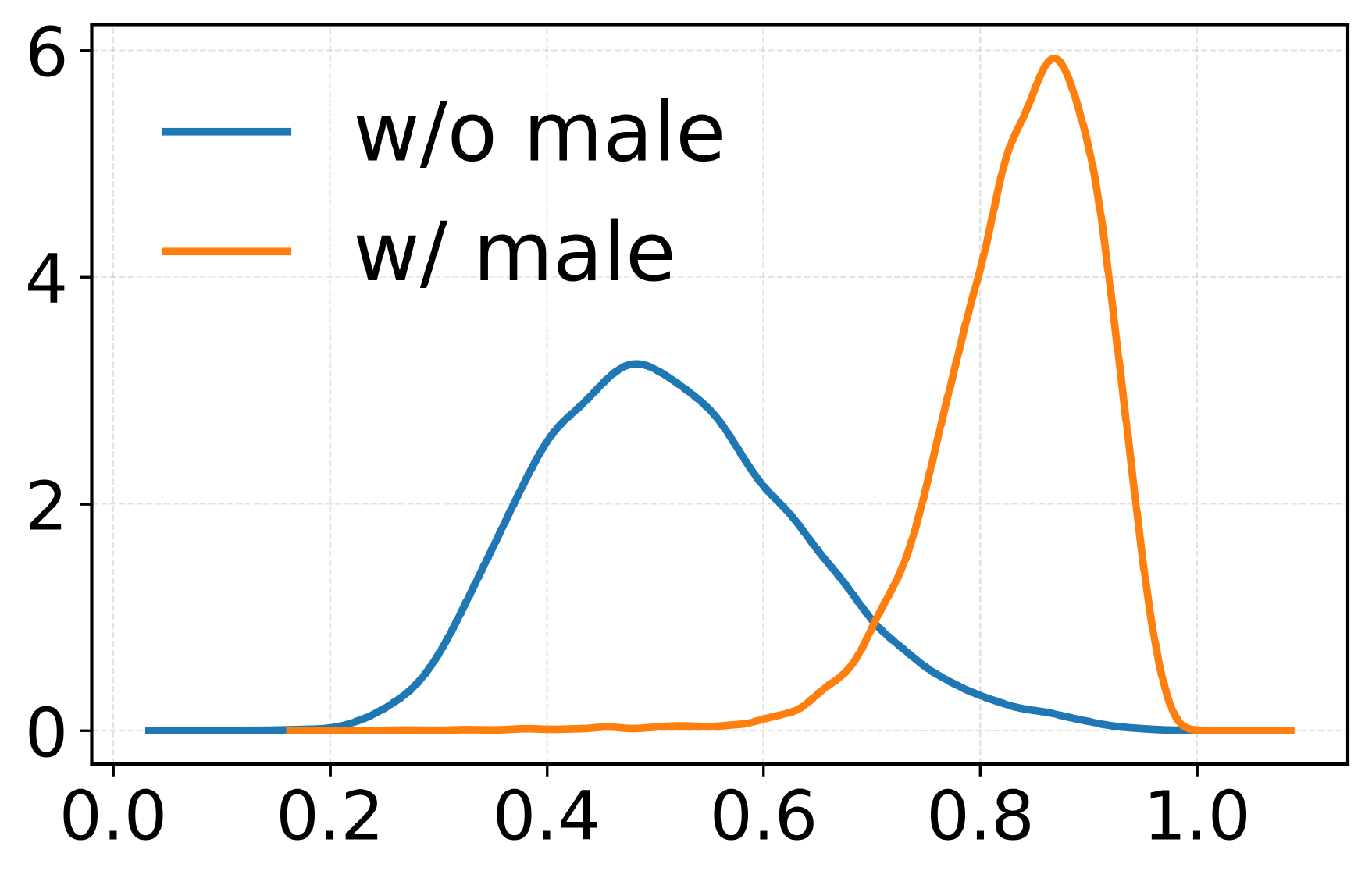}
\parbox{\linewidth}{\centering \small Male (Full)}
\end{subfigure}
\begin{subfigure}{0.24\linewidth}
\centering
\includegraphics[width=\linewidth]{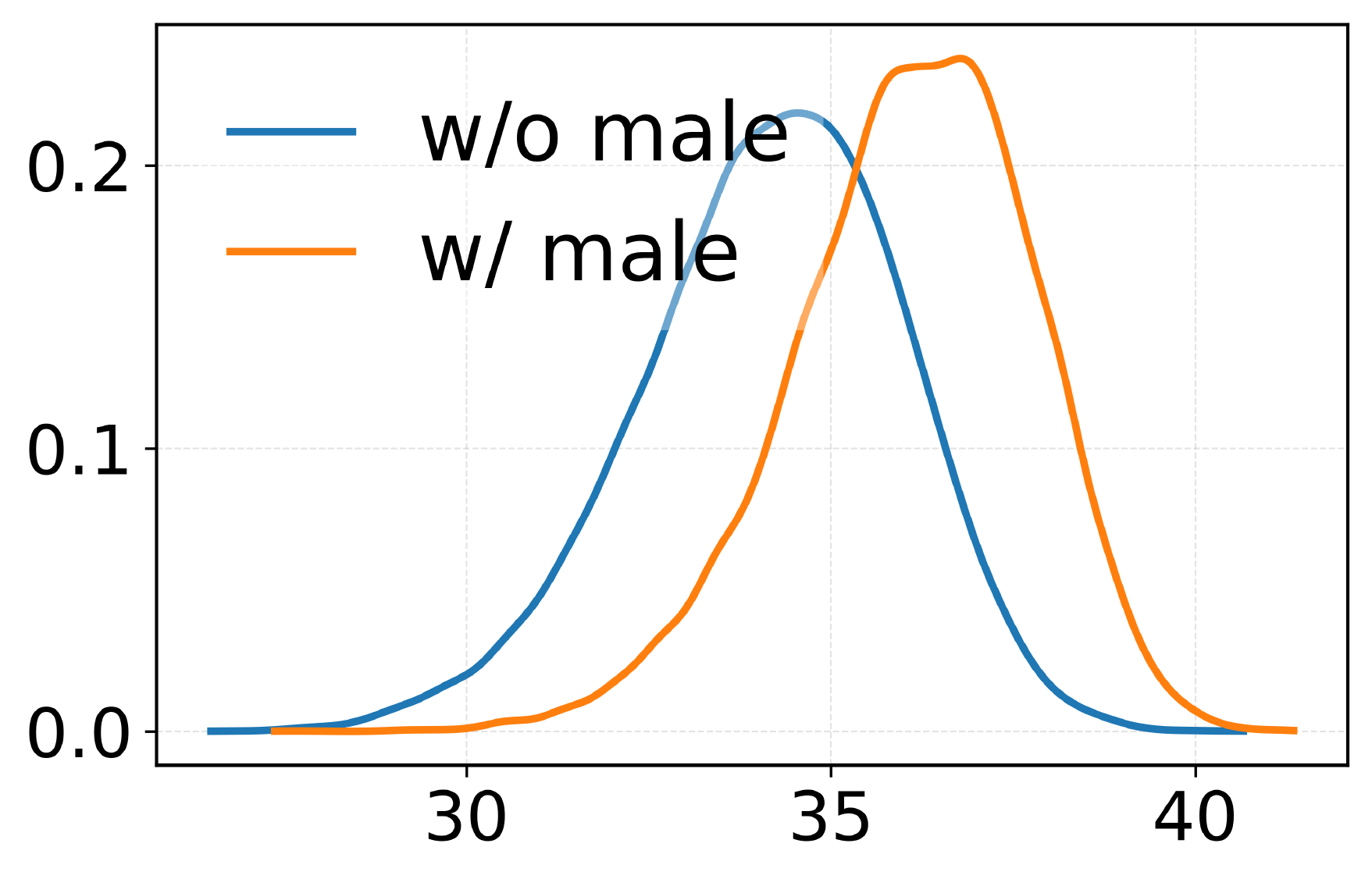}
\parbox{\linewidth}{\centering \small Male (w/o DP)}
\end{subfigure}
\begin{subfigure}{0.24\linewidth}
\centering
\includegraphics[width=\linewidth]{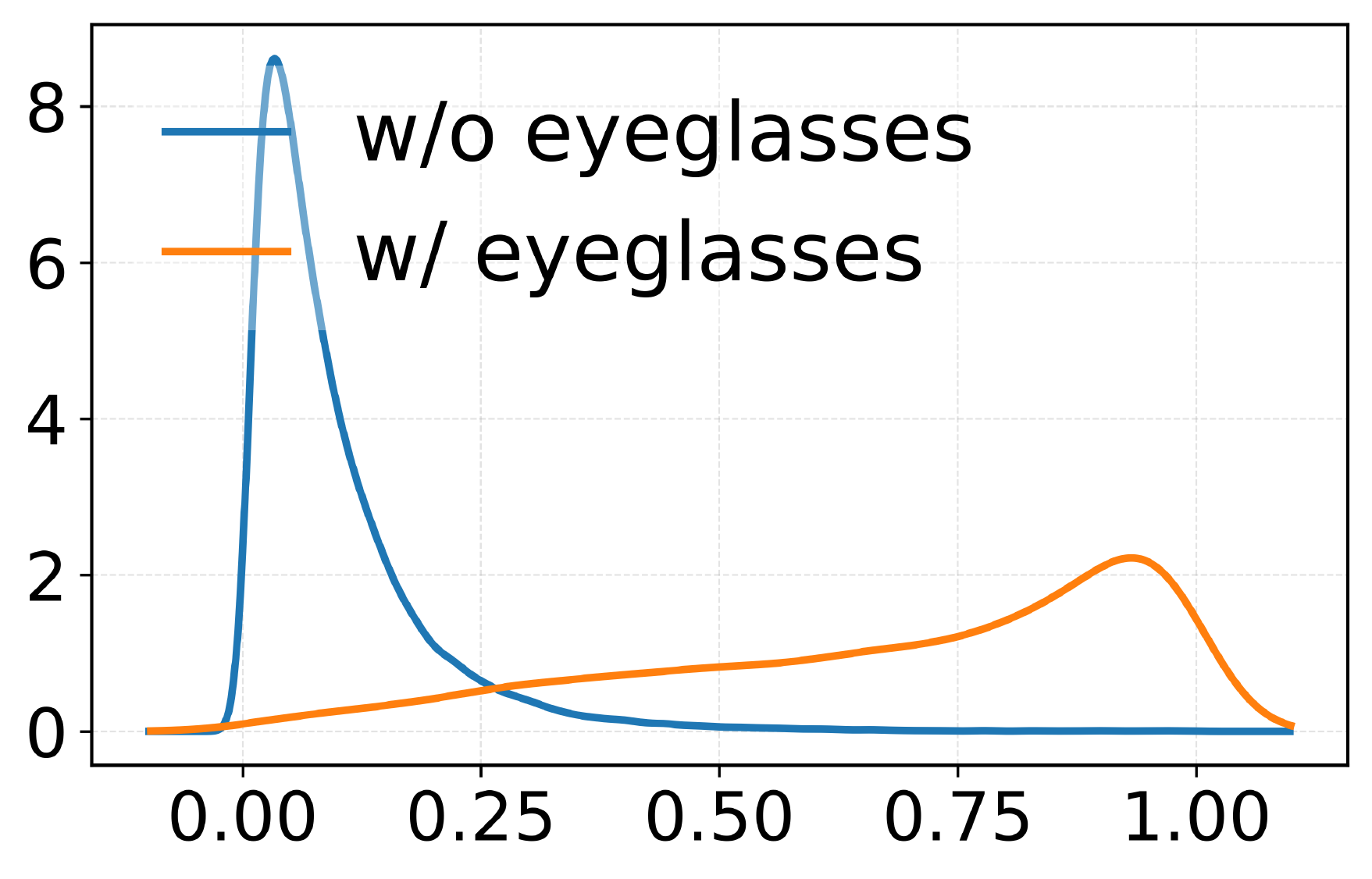}
\parbox{\linewidth}{\centering \small Eyeglasses (Full)}
\end{subfigure}
\begin{subfigure}{0.24\linewidth}
\centering
\includegraphics[width=\linewidth]{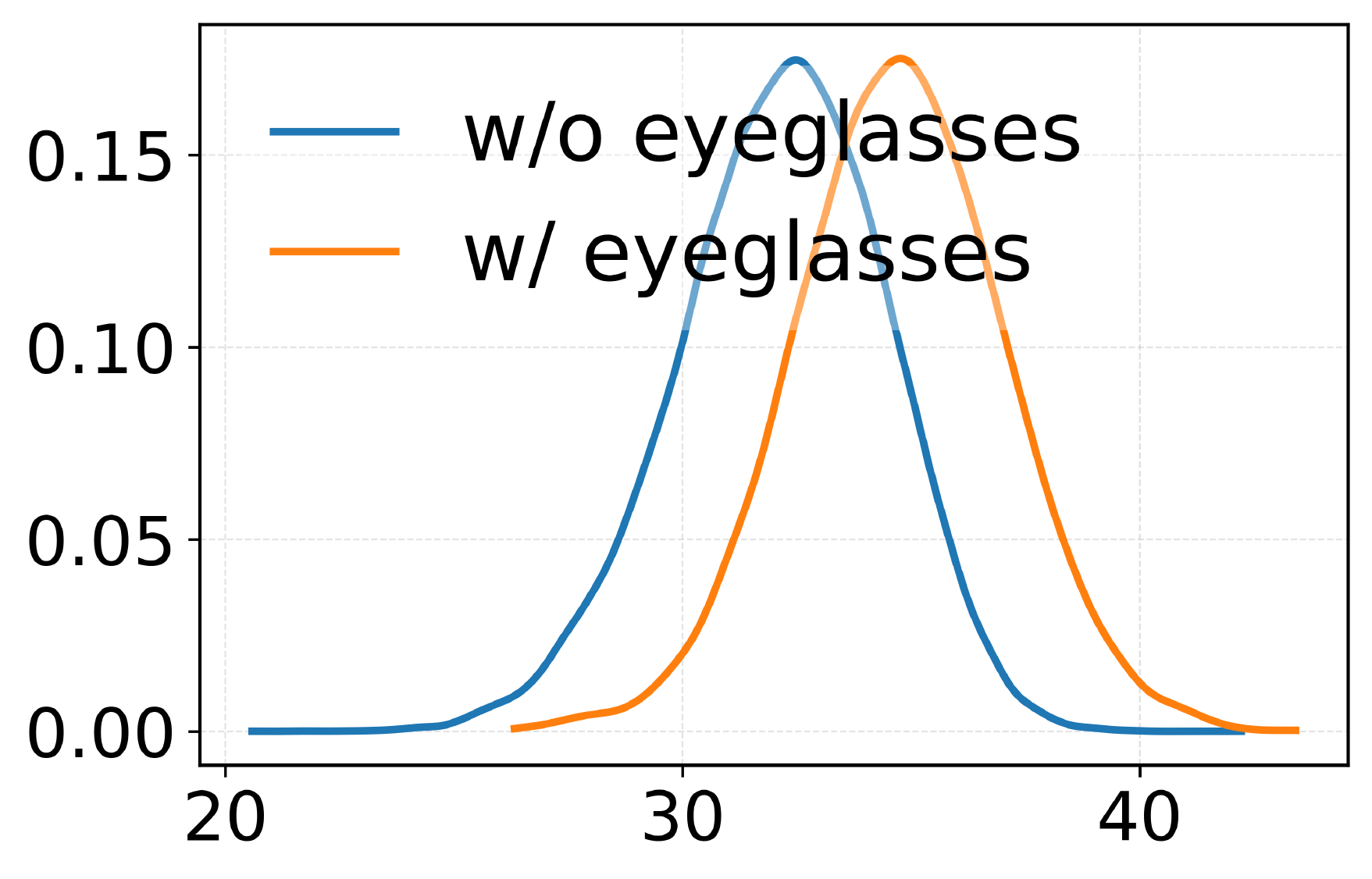}
\parbox{\linewidth}{\small Eyeglasses (w/o DP)}
\end{subfigure}
\vspace{-2pt}
\caption{\small Ablation study comparing full method with w/o DP. Orange curves show score distributions for images with the attribute present, and blue curves for images without the attribute. Dual prompts improve distribution separation for both attributes.}
\label{fig:ablation_dp}
\end{figure}

We report the mean and standard deviation of Cohen's $d$ across attributes within each category in Tab.~\ref{tab:ablation_study}. The full RA-CLIPScore achieves the highest average discriminability across all three attribute groups, outperforming both CLIPScore and HCS. Removing global embeddings primarily degrades performance on global attributes, while removing local tokens has a greater impact on local attributes, confirming the complementary roles of global and local representations. Removing dual prompts or attention refinement also reduces average discriminability across all groups, highlighting their contributions to attribute separation. Fig.~\ref{fig:ablation_dp} further illustrates how dual prompts improve the separation between positive and negative score distributions. Additional analyses of representative attributes from each category, namely \textit{gender} (global), \textit{eyeglasses} (local), and \textit{smiling} (other), are provided in Supp.~\ref{subsec:sup-mat_ablation}.

\begin{table}[b]
\centering
\setlength{\tabcolsep}{3pt}
\renewcommand{\arraystretch}{1.1}

\resizebox{\columnwidth}{!}{%
\begin{tabular}{lccccccc}
\toprule
\textbf{Attribute}
& \textbf{CLIPScore}
& \textbf{HCS}
& \textbf{Ours (Full)}
& \textbf{w/o DP}
& \textbf{w/o LT}
& \textbf{w/o GE}
& \textbf{w/o AR} \\
\midrule
Global (11)
& $0.83 \pm 0.47$
& $0.72 \pm 0.63$
& $\mathbf{1.13 \pm 0.91}$
& $0.76 \pm 0.56$
& $1.11 \pm 0.82$
& $0.90 \pm 0.77$
& $0.83 \pm 0.70$ \\
Local (5)
& $1.47 \pm 0.48$
& $1.75 \pm 0.54$
& $\mathbf{2.03 \pm 0.63}$
& $1.73 \pm 0.40$
& $1.62 \pm 0.56$
& $1.87 \pm 0.58$
& $1.89 \pm 0.54$ \\
Other (4)
& $0.80 \pm 0.45$
& $0.73 \pm 0.52$
& $\mathbf{0.92 \pm 0.78}$
& $0.76 \pm 0.55$
& $0.89 \pm 0.72$
& $0.75 \pm 0.66$
& $0.72 \pm 0.64$ \\
\bottomrule
\end{tabular}%
}
\caption{Ablation study on $20$ CelebA attributes, grouped into global, local, and other attributes. Results are reported as mean $\pm$ standard deviation of Cohen's $d$ across attributes. Higher values indicate better separation between positive and negative samples. Best results are highlighted in bold.}
\label{tab:ablation_study}
\end{table}

\begin{table}[t]
\begin{minipage}[t]{0.49\linewidth}
\vspace{0pt}
\begin{tabular}{@{}llcc@{}}
\toprule
\textbf{Method} & \textbf{Metric} & \textbf{A vs.\ B} & \textbf{A vs.\ C} \\
\midrule
\multirow{3}{*}{RA-CLIPScore}
 & SaD$_{EG}$      & 4.52  & 693.97 \\
 & SaD$_{\neg EG}$ & 1.11  & 1.10   \\
 & PaD             & 2.80  & 79.00  \\
\midrule
\multirow{3}{*}{HCS}
 & SaD$_{EG}$      & 3.89  & 93.86  \\
 & SaD$_{\neg EG}$ & 3.43  & 5.94   \\
 & PaD             & 20.97 & 54.98  \\
\bottomrule
\end{tabular}
\caption{\small Results of dataset comparison under inverse attribute coupling (\textit{eyeglasses}$\times$\textit{gender}). RA-CLIPScore shows higher sensitivity.}
\label{tab:inverse_attr_coupling}
\end{minipage}
\hfill
\begin{minipage}[t]{0.48\linewidth}
\vspace{0pt}  %
\resizebox{\linewidth}{!}{
\begin{tabular}{@{}lcccc@{}}
\toprule
 & \makecell{\textbf{Style}\\\textbf{GAN2}}
 & \makecell{\textbf{Style}\\\textbf{GAN3}}
 & \makecell{\textbf{LDM}\\\textbf{(50)}}
 & \makecell{\textbf{LDM}\\\textbf{(200)}} \\
\midrule
FID~\cite{heusel2017gans} $\downarrow$ & \textbf{2.94} & 5.20 & 8.85 & 9.53 \\
KID~\cite{sutherland2018demystifying} ($\times10^{-4}$) $\downarrow$ & \textbf{5.0} & 8.2 & 60 & 69 \\
Precision~\cite{kynkaanniemi2019improved,sajjadi2018assessing} $\uparrow$ & \textbf{0.792} & 0.769 & 0.703 & 0.700 \\
Recall~\cite{kynkaanniemi2019improved,sajjadi2018assessing} $\uparrow$ & 0.665 & \textbf{0.700} & 0.666 & 0.663 \\
Coverage~\cite{naeem2020reliable} $\uparrow$ & \textbf{0.948} & 0.947 & 0.832 & 0.827 \\
Density~\cite{naeem2020reliable} $\uparrow$ & \textbf{1.11} & 1.02 & 0.72 & 0.67 \\
SaD ($\times10^{-3}$) $\downarrow$ & \textbf{2.25} & 2.98 & 5.64 & 6.44 \\
PaD ($\times10^{-3}$) $\downarrow$ & \textbf{4.9} & 6.8 & 12.3 & 14.0 \\
\bottomrule
\end{tabular}}
\captionof{table}{\small Quantitative evaluation results of image generative models on FFHQ.  
}
\label{tab:sota_evaluation}
\end{minipage}
\vspace{-3mm}
\end{table}


\vspace{-2mm}
\subsection{Evaluation of Generative Image Models}
\label{subsec:evaluation_sota}
\vspace{-3mm}
We evaluate several generative image models~\cite{karras2019style,karras2020analyzing,Rombach2022ldm} trained on the FFHQ dataset~\cite{karras2019style}. For each model, $50{,}000$ images are generated using official checkpoints and compared with $50{,}000$ real FFHQ samples. For attribute-level evaluation, we use $20$ face-related attributes derived from CelebA annotations~\cite{liu2015faceattributes}. All evaluations are conducted at a resolution of $256\times256$. The full attribute list is provided in Supp.\ref{subsec:sup-mat_eval}. 

The results in Tab.\ref{tab:sota_evaluation} report conventional image-quality metrics 
alongside our attribute-aware metrics (SaD and PaD). Arrows indicate whether higher ($\uparrow$) or lower ($\downarrow$) values are better. Bar charts highlighting the most discrepant attributes are provided in Supp.~\ref{subsec:sup-mat_eval}. Overall, our metrics show strong alignment with traditional quality measures. Across most metrics, StyleGAN2 performs best, followed by StyleGAN3, while diffusion-based models (LDM) underperform in the face domain. This is expected, as StyleGAN family is specialized for high-fidelity face synthesis, whereas diffusion models prioritize broader generalization. Interestingly, LDM(50) outperforms LDM(200) despite using fewer sampling steps. From bar charts (Supp.\ref{subsec:sup-mat_eval}), SaD reveals that \textit{bald head} contributes most to the higher divergence in diffusion models, and larger sampling steps worsen this gap. We also observe that excessive denoising steps may overemphasize fine details (e.g. \textit{wearing earrings}), leading to distributions that deviate from real data.We additionally evaluate SaD on ImageNet, observing rankings consistent with FID and KID while revealing substantial class-wise variation in attribute-level discrepancies (Supp.~\ref{subsec:sup-mat_imagenet_eval}).

\vspace{-2mm}
\subsection{Spatial Evaluation of Generative Image Models}
\label{subsec:spatial_eval}
\vspace{-2mm}

We further evaluate the spatial and semantic consistency of image generative models~\cite{sauer2022stylegan,Rombach2022ldm,brock2018large} using the proposed R-SaD on ImageNet~\cite{deng2009imagenet}. Each model generates class-conditional images using official checkpoints, and the generated samples are compared with the corresponding real ImageNet subsets. The $1{,}000$ class names serve as textual attributes for spatial evaluation. All evaluations are conducted at a resolution of $256\times256$. 

\begin{figure}[t]
\centering
\begin{minipage}[t]{0.48\columnwidth}
\vspace{0pt}
\centering
\begin{subfigure}[t]{0.32\linewidth}
    \includegraphics[width=\linewidth]{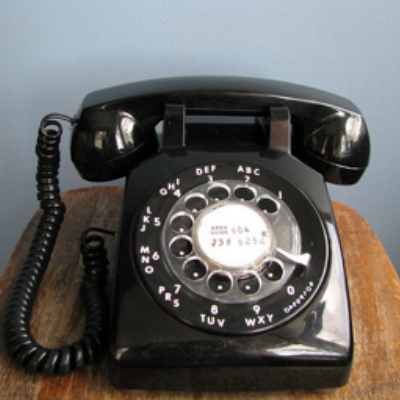}
    \parbox{\linewidth}{\centering \small \mbox{ImageNet}}
\end{subfigure}\hfill
\begin{subfigure}[t]{0.32\linewidth}
    \includegraphics[width=\linewidth]{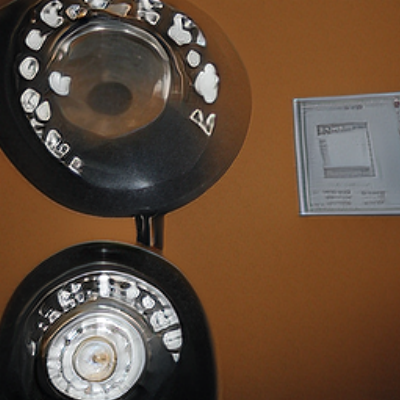}
    \parbox{\linewidth}{\centering \small \mbox{StyleGAN-XL}}
\end{subfigure}\hfill
\begin{subfigure}[t]{0.32\linewidth}
    \includegraphics[width=\linewidth]{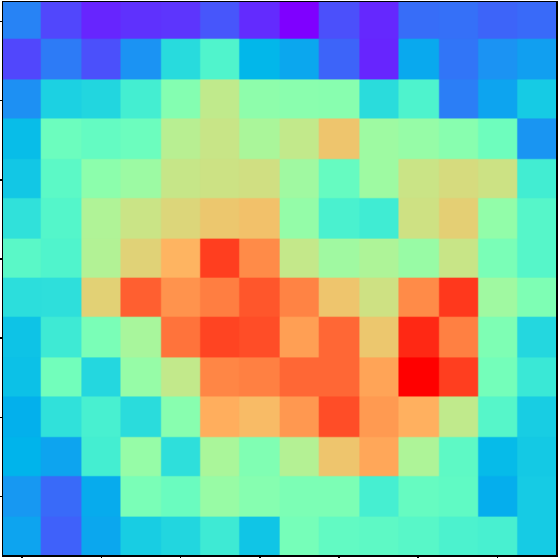}
    \parbox{\linewidth}{\centering \small \mbox{R-SaD}}
\end{subfigure}
\vspace{0.4em}
\begin{subfigure}[t]{0.32\linewidth}
    \includegraphics[width=\linewidth]{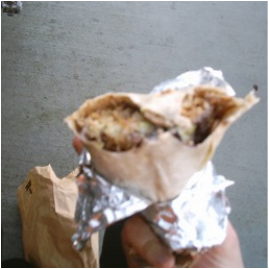}
    \parbox{\linewidth}{\centering \small \mbox{ImageNet}}
\end{subfigure}\hfill
\begin{subfigure}[t]{0.32\linewidth}
    \includegraphics[width=\linewidth]{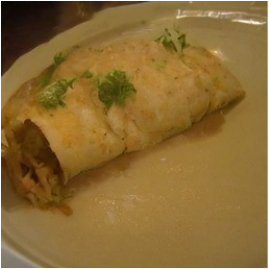}
    \parbox{\linewidth}{\centering \small \mbox{LDM}}
\end{subfigure}\hfill
\begin{subfigure}[t]{0.32\linewidth}
    \includegraphics[width=\linewidth]{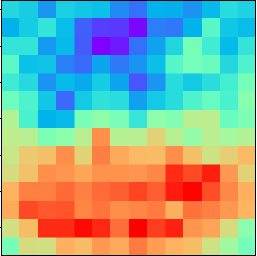}
    \parbox{\linewidth}{\centering \small \mbox{R-SaD}}
\end{subfigure}
\vspace{0.4em}
\begin{subfigure}[t]{0.32\linewidth}
    \includegraphics[width=\linewidth]{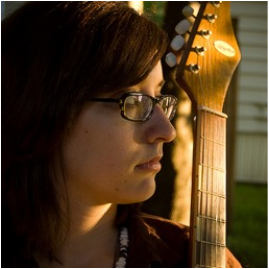}
    \parbox{\linewidth}{\centering \small \mbox{ImageNet}}
\end{subfigure}\hfill
\begin{subfigure}[t]{0.32\linewidth}
    \includegraphics[width=\linewidth]{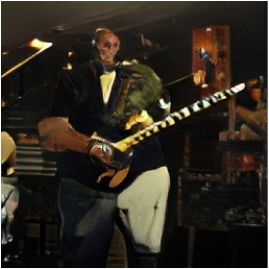}
    \parbox{\linewidth}{\centering \small \mbox{BigGAN}}
\end{subfigure}\hfill
\begin{subfigure}[t]{0.32\linewidth}
    \includegraphics[width=\linewidth]{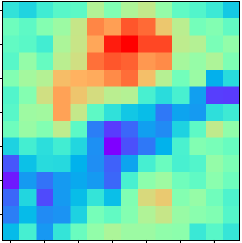}
    \parbox{\linewidth}{\centering \small \mbox{R-SaD}}
\end{subfigure}
\caption{
    \small Spatial evaluation on ImageNet. Each row shows a real example (left), a generated sample (middle), and the corresponding R-SaD heatmap (right). Heatmaps visualize regional divergence normalized from $0$ (blue, higher occurrence in generated images) to $1$ (red, higher occurrence in training data).
    }
    \label{fig:spatial_eval_3x3}
\end{minipage}
\hfill
\begin{minipage}[t]{0.48\linewidth}
\vspace{0pt}
\centering
\includegraphics[width=0.48\linewidth]{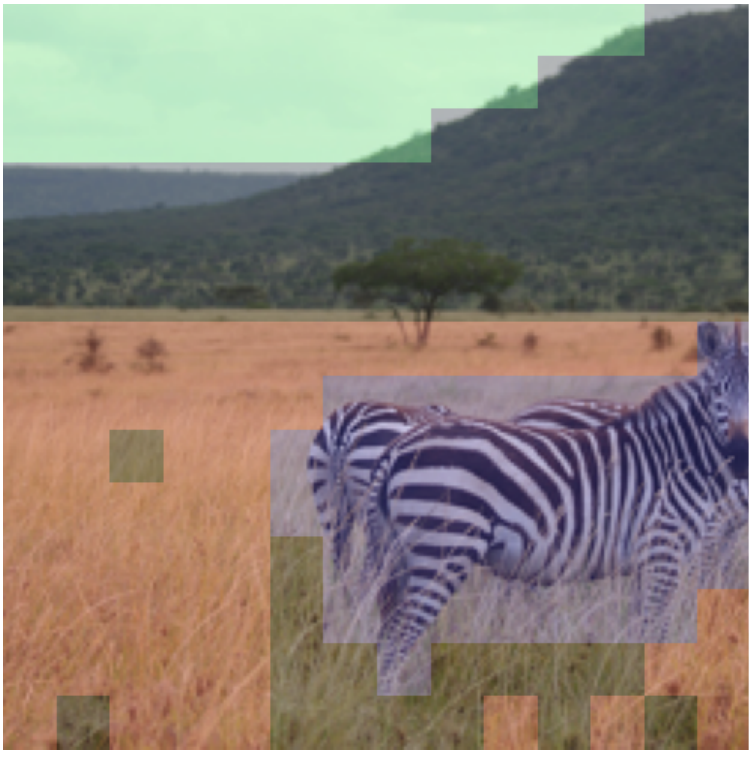}\hfill
\includegraphics[width=0.48\linewidth]{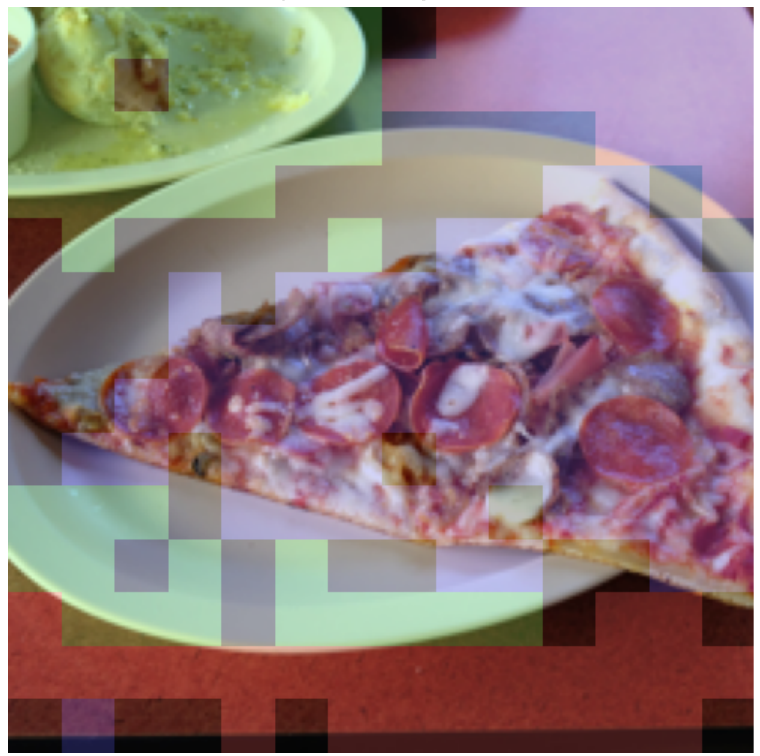}
\captionof{figure}{\small Qualitative visualizations on single images. Left: zebra (blue), sky (green), grass (orange). Right: pizza (blue), plate (green), table (orange).}
\label{fig:patch_scores_qual}
\vspace{1em}

\footnotesize
\setlength{\tabcolsep}{2pt}
\renewcommand{\arraystretch}{1.15}

\resizebox{\linewidth}{!}{%
\begin{tabular}{@{}lcl@{}}
\toprule
\textbf{Model}
& \textbf{R-SaD (Mean $\pm$ Std)} $\downarrow$
& \textbf{Top-3 Classes} \\
\midrule
StyleGAN-XL
& $\mathbf{0.0136 \pm 0.0140}$
& \makecell[l]{rotary dial telephone\\carousel\\iPod} \\
\midrule
BigGAN
& $0.0232 \pm 0.0234$
& \makecell[l]{balloon\\electric guitar\\microphone} \\
\midrule
LDM (250)
& $0.0146 \pm 0.0147$
& \makecell[l]{burrito\\chain-link fence\\carton} \\
\bottomrule
\end{tabular}%
}

\captionof{table}{\small
Spatial evaluation on ImageNet using R-SaD.
Values are reported as the mean $\pm$ standard deviation.
The top three classes for each model are ranked by their
mean absolute R-SaD across spatial patches.
}
\label{tab:spatial_eval}
\end{minipage}
\end{figure}

We report the mean and standard deviation of absolute R-SaD across all $1{,}000$ ImageNet classes in Tab.~\ref{tab:spatial_eval}, along with the three classes exhibiting the largest spatial discrepancies. These classes are ranked by the mean absolute R-SaD across spatial patches, defined as
$\operatorname*{arg\,max}_{t_i}
\frac{1}{|\mathcal{J}|}
\sum_{j \in \mathcal{J}}
|\mathrm{R\text{-}SaD}_j(t_i)|$.
We further visualize image-level local scores computed using Eq.~\eqref{eq:regional_score} in Fig.~\ref{fig:patch_scores_qual}, and present dataset-level spatial discrepancies through normalized R-SaD heatmaps in Fig.~\ref{fig:spatial_eval_3x3}. Values are scaled to $[0,1]$, where red indicates higher occurrence in training data and blue indicates higher occurrence in generated samples. For clarity, we display one representative example per category for real and generated datasets, additional randomly selected samples are provided in Supp.\ref{subsec:sup-mat_spatial_eval}, where similar trends are observed. The results reveal distinct spatial bias patterns across models. StyleGAN-XL~\cite{sauer2022stylegan} generates rotary dial phones at more varied locations, whereas real images are mostly centered, resulting in prominent red regions at the center. BigGAN~\cite{brock2018large} consistently produces certain objects, such as electric guitars, at an approximately $45^\circ$ diagonal orientation, while real ImageNet samples exhibit more diverse poses. For LDM~\cite{Rombach2022ldm}, categories such as burrito show a spatial shift toward the upper image region in generated samples.

\vspace{-2mm}
\subsection{Alignment with Human Perception}
\label{subsec:userstudy}
\vspace{-2mm}
We additionally conduct a user study to evaluate the alignment between R-SaD and human perception. A total of $29$ participants judged whether sets of $9$ images generated by model A or model B more closely resembled $9$ reference images from the training data. Each participant evaluated $20$ attributes, resulting in $580$ trials in total. Following Sec.\ref{subsec:spatial_eval}, we compare StyleGAN-XL~\cite{sauer2022stylegan}, BigGAN~\cite{brock2018large}, and LDM~\cite{Rombach2022ldm}. We report Pearson correlation coefficient ($r$) and observe that R-SaD achieves strong agreement with human preferences, demonstrating consistency with human perception. Additional details are provided in Supp.~\ref{subsec:sup-mat_userstudy}.

%% file: text/conclusions.tex
\vspace{-3mm}
\section{Conclusions and Discussion}
\label{sec:conclusions}
\vspace{-3mm}
Conventional metrics~\cite{heusel2017gans, sutherland2018demystifying} for image generative models provide only scalar scores, offering limited insights and diagnostic capability. To address this limitation, we propose RA-CLIPScore, a CLIP-based metric for attribute-wise and spatial evaluation. RA-CLIPScore mitigates two key limitations inherited from CLIP’s contrastive pretraining paradigm through dual prompts and local patch tokens. Unlike prior approaches, our aggregation scheme jointly integrates global and local information in a single forward pass without additional training or computational overhead. By incorporating localized features, RA-CLIPScore extends CLIP-based evaluation to the spatial domain for the first time, enabling the detection of systematic biases, such as preferences for specific object orientations or spatial locations, that remain invisible to existing diversity benchmarks. Extensive experiments and human studies demonstrate that our metrics achieve improved attribute disentanglement, robustness, and interpretability, while aligning closely with human perception.

\noindent\textbf{Limitations and Future Work.} 
Motivated by minimalism and comparability, we adopt CLIP as central feature extractor, as it increasingly replaces Inception-V3~\cite{szegedy2016rethinking} in traditional metrics~\cite{jayasumana2024rethinking,kynkaanniemi2022role,betzalel2022study,morozov2021self,parmar2022aliased} and serves as a backbone in many modern generative models~\cite{dhariwal2021diffusion, lorenz2022high,ramesh2022hierarchical}. This design maintains compatibility with prior evaluation metrics while avoiding additional sources of bias. However, CLIP itself has known limitations~\cite{yuksekgonul2023and}, particularly in compositional reasoning, counting, and accurately representing negation. Exploring alternative encoders for attribute-level and spatial evaluation would therefore be valuable. Future work could investigate other spatial feature extractors, such as SigLIP~\cite{zhai2023sigmoid}, SAM~\cite{kirillov2023segment}, or DINO~\cite{caron2021emerging}. Their stronger representation capabilities could provide complementary insights. We observe in Sec.~\ref{subsec:ablation_study} that global and local tokens are more effective for dominant and fine-grained attributes, respectively. Since practical scenarios typically involve both attribute types, we currently integrate them in our metric through a probability-weighted summation. An interesting direction for future work is attribute-adaptive aggregation of global and local features within RA-CLIPScore, such as confidence- or variance-weighted gating mechanisms between global and local representations. Our current evaluation follows the distribution-matching paradigm of conventional metrics such as FID and KID, using the training dataset as the reference distribution. Consequently, biases present in the reference data may not be penalized. However, our framework could be extended to user-defined spatial priors by replacing empirical reference statistics with target spatial distributions, such as uniform or task-specific priors in Eq.\eqref{eq:regional_sad} This would enable evaluation against desired spatial properties beyond the training distribution. Although our focus is on image generation, our design could be extended to other modalities, e.g. using CLAP~\cite{elizalde2023clap} for audio evaluation in future work.

%% file: text/acknowledgments.tex
\section*{Acknowledgments}
We gratefully acknowledge access to the Berzelius computing resource at the National Supercomputer Centre, provided by the Knut and Alice Wallenberg Foundation. This work was supported by the Wallenberg AI, Autonomous Systems and Software Program (WASP).

%% file: text/suppl.tex
\clearpage
\setcounter{page}{1}
\appendix
\section*{Supplementary Material}

\input{text/sup-mat/suppl_rethink}
\input{text/sup-mat/suppl_mllm}

\input{text/sup-mat/suppl_method}

\input{text/sup-mat/suppl_exp}

%% file: text/sup-mat/suppl_rethink.tex
\section{Rethinking HCS}
\label{sec:sup-mat_rethinking}

\citet{kim2023attribute} introduced the HCS to enable attribute-wise evaluation of generative models. HCS computes attribute scores by subtracting the global mean centers of image and text embeddings (Eq.\eqref{eq:hcs}), with the aim to reduce ambiguity between opposing attributes~\cite{kim2023attribute}. While conceptually appealing, this formulation inherently couples the score with both the training distribution and the attribute set, making it sensitive to dataset bias and attribute selection. Its effectiveness relies on strong assumptions: the training and generated datasets must be well aligned, and the selected textual attributes must be semantically relevant to the images. When these assumptions are violated, HCS becomes theoretically problematic and lacks robustness in realistic evaluation scenarios. We demonstrate this instability through both empirical and theoretical analyses.


\subsection{Empirical Observations}
\begin{figure*}[h]
  \centering
  \begin{subfigure}[t]{0.2\textwidth}
  \vspace{-34mm}
    \includegraphics[width=\linewidth]{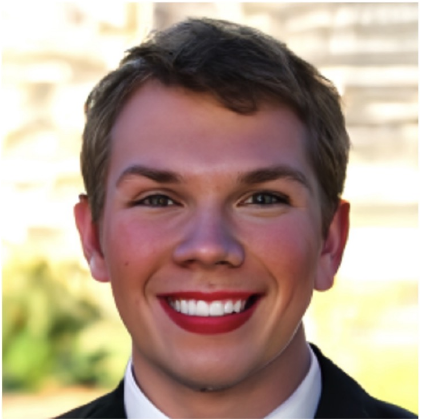} 
    \vspace{2mm}
    \caption{Image}
    \label{fig:three:a}
  \end{subfigure}\hfill
  \begin{subfigure}[t]{0.35\textwidth}
    \includegraphics[width=\linewidth]{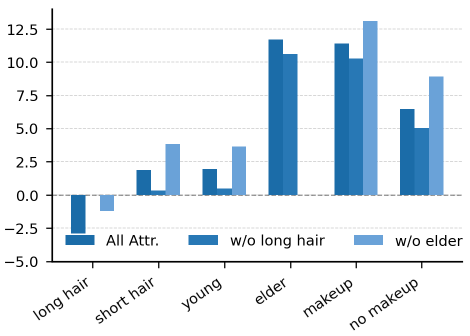}
    \caption{HCS}
    \label{fig:three:b}
  \end{subfigure}\hfill
  \begin{subfigure}[t]{0.35\textwidth}
    \includegraphics[width=\linewidth]{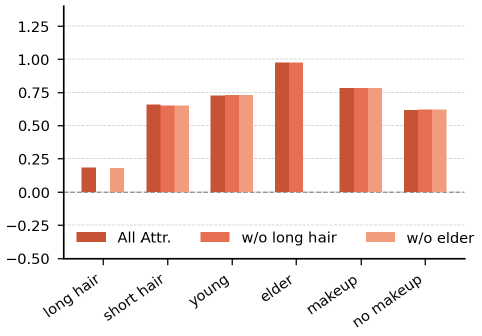}
    \caption{RA-CLIPScore}
    \label{fig:three:c}
  \end{subfigure}

  \caption{\small{
  Sensitivity to attribute removal. Removing a single attribute (\textit{long hair} or \textit{elder}) causes noticeable shifts in HCS, while RA-CLIPScore remains consistent under the same perturbations. Horizontal axes enumerate evaluated attributes. Three settings are compared: (1) using all $6$ attributes (2) removing \textit{long hair} (3) removing \textit{elder}.
  }}
  \label{fig:removing_attributes}
\end{figure*}

As illustrated in Fig. \ref{fig:removing_attributes}, HCS scores are highly sensitive to the choice of attributes. Removing a single attribute from the evaluation list can drastically alter the results, especially for semantically related or opposing attributes. For instance, excluding \textit{long hair} reduces the score for \textit{short hair} to nearly zero, while removing \textit{elder} increases the score for \textit{young} to almost twice its original value.  In contrast, our proposed metric remains stable under such perturbations, demonstrating robustness to variations in attribute composition.

\subsection{Analytical Case: Two-Attribute Scenario}
\label{subsec:sup-mat_theoretical_proof}
To further illustrate the limitation, consider a simplified case with two attributes, $t_p$ and $t_q$, with corresponding text embeddings $\mathbf{E}_{t}(t_p)$ and $\mathbf{E}_{t}(t_q) \in \mathbb{R}^{N_E}$, where $N_E$ denotes the CLIP text embedding dimension.  
According to Eq.\eqref{eq:hcs}, the global text mean is given by:
\vspace{-2mm}
\begin{equation}
    \mathbf{C}_{\text{text}} = \frac{\mathbf{E}_{v}(t_p) + \mathbf{E}_{v}(t_q)}{2}
\end{equation}
Subtracting this center yields the following:
\begin{equation}
\begin{split}
\mathbf{E}_{v}(t_p) - \mathbf{C}_{\text{text}}
&= \frac{\mathbf{E}_{v}(t_p) - \mathbf{E}_{v}(t_q)}{2} \\
&= -\frac{\mathbf{E}_{v}(t_q) - \mathbf{E}_{v}(t_p)}{2} = -\big(\mathbf{E}_{v}(t_q) - \mathbf{C}_{\text{text}}\big)
\end{split}
\end{equation}
This derivation shows that the two attributes become exact opposites in the embedding space, regardless of their true semantic relationship. Even synonymous attributes would thus be treated as opposing directions, distorting their relative semantics. Therefore, subtracting a global mean center does not guarantee attribute disentanglement.

\subsection{Hypothesis of Limitations}
Both HCS~\cite{kim2023attribute} and CLIPScore~\cite{hessel2021clipscore} rely on CLIP~\cite{radford2021learning} for feature extraction and therefore inherit its structural constraints for attribute-level and region-wise evaluation. We revisit CLIP’s embedding design and identify two key issues underlying these limitations~\cite{sun2022dualcoop,lin2024tagclip}:  
(1) its softmax-based training objective introduces competition among textual categories, amplifying dominant attributes while suppressing co-occurring or overlapping ones, thereby hindering the accurate representation of multi-attribute concepts at the same time; and (2) each image is encoded into a single global embedding via the \texttt{[CLS]} token, limiting its ability to capture localized or region-specific cues~\cite{dosovitskiy2020image}.

We argue that these factors fundamentally restrict the reliability of HCS and CLIPScore for attribute-wise and spatial evaluation. Mean-centering the embeddings, as done in HCS (Eq.\eqref{eq:hcs}), can offer correction under idealized conditions but fails to address these deeper architectural issues rooted in CLIP’s contrastive pretraining paradigm.




%% file: text/sup-mat/suppl_mllm.tex
\section{Comparison with MLLM/VQA-based Evaluation Methods}
\label{sec:sup-mat_mllm}

We compare RA-CLIPScore with representative MLLM/VQA-based evaluation methods in terms of computational cost and evaluation capabilities. As shown in Tab.~\ref{tab:sup-mat_mllm_cost}, RA-CLIPScore requires substantially fewer parameters and FLOPs per forward pass. Tab.~\ref{tab:sup-mat_mllm_comparison} further highlights their complementary strengths: RA-CLIPScore enables deterministic, training-free evaluation with native support for patch-level discrepancy maps and dataset-level distribution comparison, whereas MLLM/VQA-based judges offer stronger compositional reasoning and open-ended explanations. Overall, RA-CLIPScore provides an efficient and reproducible approach to large-scale, spatially interpretable evaluation, complementing the semantic reasoning capabilities of MLLM/VQA-based methods.

\begin{table*}[t]
\centering
\small

\begin{minipage}[t]{0.48\textwidth}
    \centering
    \setlength{\tabcolsep}{3pt}
    \begin{tabular*}{\linewidth}{@{\extracolsep{\fill}}lrr@{}}
        \toprule
        \textbf{Method} & \textbf{Params (M)} & \textbf{FLOPs (G)} \\
        \midrule
        \textbf{RA-CLIPScore} (ours) & \textbf{149.62} & \textbf{19.78} \\
        CLIP-FlanT5~\cite{li2024genai} & 3225.24 & 1059.99 \\
        BLIP-VQA~\cite{huang2023t2i} & 361.80 & 106.53 \\
        MiniGPT4~\cite{huang2023t2i} & 7832.63 & 863.74 \\
        ShareGPT4~\cite{huang2023t2i} & 7062.90 & 4427.06 \\
        mPLUG~\cite{hu2023tifa} & 428.19 & 529.06 \\
        \bottomrule
    \end{tabular*}
    \caption{Computational cost of RA-CLIPScore and representative MLLM/VQA-based methods, measured by model parameters and FLOPs per forward pass.}
    \label{tab:sup-mat_mllm_cost}
\end{minipage}
\hfill
\begin{minipage}[t]{0.48\textwidth}
    \centering
    \setlength{\tabcolsep}{3pt}
    \begin{tabular*}{\linewidth}{@{\extracolsep{\fill}}lll@{}}
        \toprule
        \textbf{Aspect} & \textbf{Ours} & \textbf{MLLM/VQA} \\
        \midrule
        Training / Fine-tuning & None & Usually none \\
        Deterministic & Yes & Model-dependent \\
        50k-image evaluation & Efficient & High cost \\
        Patch-level map & Native & Not native \\
        Dataset comparison & Native & Custom aggregation \\
        Compositional reasoning & Limited & Stronger \\
        Open-ended explanation & No & Yes \\
        Reproducibility & High & Model-dependent \\
        \bottomrule
    \end{tabular*}
    \caption{Qualitative comparison of evaluation capabilities between RA-CLIPScore and MLLM/VQA-based judges.}
    \label{tab:sup-mat_mllm_comparison}
\end{minipage}
\end{table*}

%% file: text/sup-mat/suppl_method.tex
\section{Experimental Validation of RA-CLIPScore}
\label{sec:sup-mat_method}

\subsection{Experimental Validation of Dual Prompts}
\label{subsec:sup-mat_dualprompt}

Although CLIP has known limitations in understanding negation~\cite{park2025know,nie2024out,sun2022dualcoop}, our dual-prompt design does not require perfect logical negation, but rather a meaningful relative semantic direction between positive and negative prompts. To empirically validate this, we define the relative semantic direction as $t = \mathbf{E}_t(\text{``young''}) - \mathbf{E}_t(\text{``not young''})$ and compute its cosine similarity with image embeddings, $\cos(\mathbf{E}_v(x), t)$, across different age groups, where $\mathbf{E}_v$ and $\mathbf{E}_t$ denote the CLIP visual and text encoders, respectively. As shown in Tab.~\ref{tab:sup-mat_dual_prompt_age}, the mean similarity decreases consistently with age, suggesting that dual prompts capture meaningful age-related variations despite CLIP's imperfect negation understanding. Our ablation study further supports the effectiveness of dual prompts, as removing them reduces the separation between attribute distributions (Sec.~\ref{subsec:ablation_study} and Supp.~\ref{subsec:sup-mat_ablation}).

\begin{table}[t]
    \centering
    \small
    \setlength{\tabcolsep}{4pt}
    \begin{tabular}{@{}lcccc@{}}
        \toprule
        \textbf{Age Group} & Baby & Children & Adults & Elderly \\
        \midrule
        $\cos(E_v(x), t)$
        & $0.037 \pm 0.013$
        & $0.016 \pm 0.004$
        & $0.011 \pm 0.030$
        & $-0.027 \pm 0.015$ \\
        \bottomrule
    \end{tabular}
    \caption{Cosine similarity between CLIP image embeddings and the relative semantic direction across age groups (mean $\pm$ std).}
    \label{tab:sup-mat_dual_prompt_age}
\end{table}

\subsection{Experimental Validation of Local Feature Extraction}
\label{subsec:sup-mat_local_features}

To validate our local feature extraction pipeline, we evaluate the localization quality of our adapted patch tokens against the original CLIP patch tokens using ground-truth segmentation masks from COCO. Specifically, we measure patch-level attribute presence using average precision (AP) and localization accuracy using intersection over union (IoU). As shown in Tab.~\ref{tab:local_feature_validation}, our method substantially outperforms the original CLIP tokens on both metrics. Since CLIP is trained with global image-text alignment, its final attention layer aggregates information into the global [CLS] token, potentially reducing the spatial specificity of local patch representations. In contrast, our attention masking bypasses this final aggregation for local tokens while retaining the global representation, thereby better preserving spatial information. These results provide empirical evidence that our adapted local features align more closely with ground-truth object regions, supporting the effectiveness of our feature extraction design.


\begin{table*}[b]
\centering

\begin{minipage}[t]{0.43\textwidth}
    \centering
    \small
    \setlength{\tabcolsep}{6pt}
    \begin{tabular}{lcc}
        \toprule
        \textbf{Method} & \textbf{AP} $\uparrow$ & \textbf{IoU} $\uparrow$ \\
        \midrule
        \textbf{Ours} & \textbf{0.626} & \textbf{0.457} \\
        Original CLIP & 0.179 & 0.093 \\
        \bottomrule
    \end{tabular}

    \captionof{table}{
        Localization performance of adapted and original
        CLIP patch tokens on COCO segmentation masks.
        Higher is better for both metrics.
    }
    \label{tab:local_feature_validation}
\end{minipage}
\hfill
\begin{minipage}[t]{0.55\textwidth}
    \centering
    \small
    \setlength{\tabcolsep}{4pt}
    \begin{tabular}{lccc}
        \toprule
        \textbf{Method}
        & \textbf{Acc.} $\uparrow$
        & \textbf{F1 (Macro)} $\uparrow$
        & \textbf{F1 (Micro)} $\uparrow$ \\
        \midrule
        HCS & 81.91 & 64.73 & 64.59 \\
        \textbf{RA-CLIPScore}
        & \textbf{84.57} & \textbf{69.91} & \textbf{69.72} \\
        CLIPScore & 82.68 & 66.24 & 65.56 \\
        \bottomrule
    \end{tabular}

    \captionof{table}{
        Action classification performance on the
        Stanford 40 Actions dataset across ten action
        categories. Higher is better for all metrics.
    }
    \label{tab:action_classification}
\end{minipage}

\end{table*}

%% file: text/sup-mat/suppl_exp.tex
\section{Details of Experiments}
\label{sec:sup-mat_exp}

\subsection{CelebA Attribute Classification}

\subsubsection{List of Used Attributes}
\label{subsec:sup-mat_celeba_attributes}
The CelebA dataset~\cite{liu2015faceattributes} provides $40$ facial attributes annotations. Some of them are subjective and ambiguous (e.g. \textit{attractive}, \textit{blurry}), which may lead to unfair evaluation. To avoid subjective bias and ensure a fair comparison, we follow the experimental setup of~\cite{kim2023attribute} and retain only the attributes with relatively objective semantic definitions. In total, we use $28$ CelebA attributes and $12$ human-face-irrelevant attributes for classification experiments to test robustness (Sec.~\ref{subsubsec:toy_celeba_classification}). The full list of all $40$ attributes, along with the selected $28$ and the $12$ irrelevant attributes used in our experiments, is provided in Tab.~\ref{tab:sup-mat_celeba_attributes}.

\begin{table*}[t] 
\centering  
\begin{tabular}{@{}ll@{}} 
\toprule 
\textbf{Attribute type} & \textbf{Attributes} \\ 
\midrule 
\begin{tabular}[t]{@{}l@{}}
$20$ attributes from CelebA \\ Injection Experiment (Sec.\ref{subsubsec:toy_injection}) \\ Dataset Comparison (Sec.\ref{subsubsec:toy_dataset_compare})
\end{tabular}
& \begin{tabular}[t]{@{}l@{}}
Arched\_Eyebrows, Bald, Bangs, Black\_Hair, \\ Blond\_Hair, Double\_Chin, Eyeglasses, Gray\_Hair, \\
Heavy\_Makeup, Male, Mustache, Rosy\_Cheeks, \\ 
Smiling, Straight\_Hair, Wavy\_Hair, \\
Wearing\_Earrings, Wearing\_Hat, \\
Wearing\_Lipstick, Wearing\_Necklace, Young 
\end{tabular} \\[4pt]
\midrule 
\begin{tabular}[t]{@{}l@{}}
$28$ attributes from CelebA
\\ Attribute Classification (Sec.\ref{subsubsec:toy_celeba_classification})
\end{tabular}
& \begin{tabular}[t]{@{}l@{}} 
Arched\_Eyebrows, Bags\_Under\_Eyes, Bald, Bangs, \\
Big\_Nose, Black\_Hair, Blond\_Hair, Brown\_Hair, \\
Chubby, Double\_Chin, Eyeglasses, Goatee, \\
Gray\_Hair, Heavy\_Makeup, Male, \\
Mouth\_Slightly\_Open, Mustache, No\_Beard, \\
Sideburns, Smiling, Straight\_Hair, Wavy\_Hair, \\
Wearing\_Earrings, Wearing\_Hat, Wearing\_Lipstick, \\
Wearing\_Necklace, Wearing\_Necktie, Young 
\end{tabular} \\[4pt] 
\midrule 
all $40$ attributes from CelebA & 
\begin{tabular}[t]{@{}l@{}} 
5\_o\_Clock\_Shadow, Arched\_Eyebrows, Attractive, \\
Bags\_Under\_Eyes, Bald, Bangs, Big\_Lips, \\
Big\_Nose, Black\_Hair, Blond\_Hair, Blurry, \\ 
Brown\_Hair, Chubby, Double\_Chin, Eyeglasses, \\
Goatee, Gray\_Hair, Heavy\_Makeup, \\
High\_Cheekbones, Male, Mouth\_Slightly\_Open, \\
Mustache, Narrow\_Eyes, No\_Beard, Oval\_Face, \\
Pale\_Skin, Pointy\_Nose, Receding\_Hairline, \\ 
Rosy\_Cheeks, Sideburns, Smiling, Straight\_Hair, \\
Wavy\_Hair, Wearing\_Earrings, Wearing\_Hat, \\
Wearing\_Lipstick, Wearing\_Necklace, \\
Wearing\_Necktie, Young 
\end{tabular} \\
\midrule
\begin{tabular}[t]{@{}l@{}}
human face irrelevant attributes
\\Attribute Classification (Sec.\ref{subsubsec:toy_celeba_classification})
\end{tabular}
&\begin{tabular}[t]{@{}l@{}}
Giraffe, Zebra, Penguin, Sofa, Chair, Nightstand, \\
Carpet, Refrigerator, Washing Machine, Pillow\\
Microwave Oven, Curtain
\end{tabular}\\
\bottomrule 
\end{tabular} 
\caption{\small{Attributes used for experiments.}}
\label{tab:sup-mat_celeba_attributes} 
\end{table*}

\subsubsection{Action Classification}
\label{subsec:sup-mat_action}

To assess the effectiveness of RA-CLIPScore beyond object-related attributes, we conduct an action classification experiment on the Stanford 40 Actions dataset~\cite{yao2011human}, following the setup in Sec.~\ref{subsec:toy_exp}. Specifically, we evaluate classification performance on ten action categories using predicted attribute scores. As shown in Tab.~\ref{tab:action_classification}, RA-CLIPScore consistently outperforms HCS and CLIPScore across accuracy, macro-F1, and micro-F1, demonstrating its effectiveness in capturing action-related concepts. Nevertheless, evaluating highly subjective or abstract concepts, such as artistic styles, remains challenging due to limitations in CLIP's semantic representations and ambiguity in ground-truth annotations. Such evaluations should therefore be interpreted with caution.


\subsubsection{Evaluation with Learnable Prompts}
\label{subsec:sup-mat_coop}

To investigate whether improved prompt representations can further enhance attribute evaluation, we adapt RA-CLIPScore using learnable prompt contexts from CoOp~\cite{zhou2022learning}. Specifically, we optimize the prompt contexts using $1,000$ labeled CelebA images and evaluate the learned prompts on the attribute classification task described in Sec.~\ref{subsubsec:toy_celeba_classification}. As shown in Tab.~\ref{tab:coop_classification}, prompt optimization substantially improves classification performance, increasing accuracy from $79.75$ to $87.46$ and macro-F1 from $45.03$ to $59.68$. However, this improvement requires supervised training and is evaluated on attributes seen during prompt optimization. In contrast, the original RA-CLIPScore remains training-free and supports evaluation of user-specified attributes without task-specific adaptation. In practical generative-model evaluation, this trade-off between accuracy, training cost, and open-vocabulary generalization should be considered.

\begin{table}[t]
\centering
\small
\setlength{\tabcolsep}{3pt}
\renewcommand{\arraystretch}{1.0}
\begin{tabular}{lccc}
\toprule
\textbf{Method} & \textbf{Acc.} & \textbf{F1 (Macro)} & \textbf{F1 (Micro)} \\
\midrule
HCS~\cite{kim2023attribute} & 79.30 & 42.12 & 54.20 \\
RA-CLIPScore & 79.75 & 45.03 & 55.20 \\
CLIPScore~\cite{hessel2021clipscore} & 79.44 & 44.04 & 54.52 \\
\midrule
RA-CLIPScore w/ CoOp~\cite{zhou2022learning}
& \textbf{87.46} & \textbf{59.68} & \textbf{73.65} \\
\bottomrule
\end{tabular}
\caption{CelebA attribute classification performance. Best results are highlighted in bold.}
\label{tab:coop_classification}
\end{table}

\subsection{Injection Experiment}
\label{subsec:sup-mat_injec_exp}

\subsubsection{List of Used Attributes}
For the injection experiments (Sec.~\ref{subsubsec:toy_injection}), we strictly follow the same setup as~\cite{kim2023attribute} and use a subset of $20$ attributes from CelebA~\cite{liu2015faceattributes}. A complete list of $20$ selected attributes is provided in Tab.~\ref{tab:sup-mat_celeba_attributes}.

\subsubsection{COCO Experiment}
\begin{figure}[t]
    \centering
    \begin{minipage}[t]{\linewidth}
        \centering
        \includegraphics[width=0.35\linewidth]{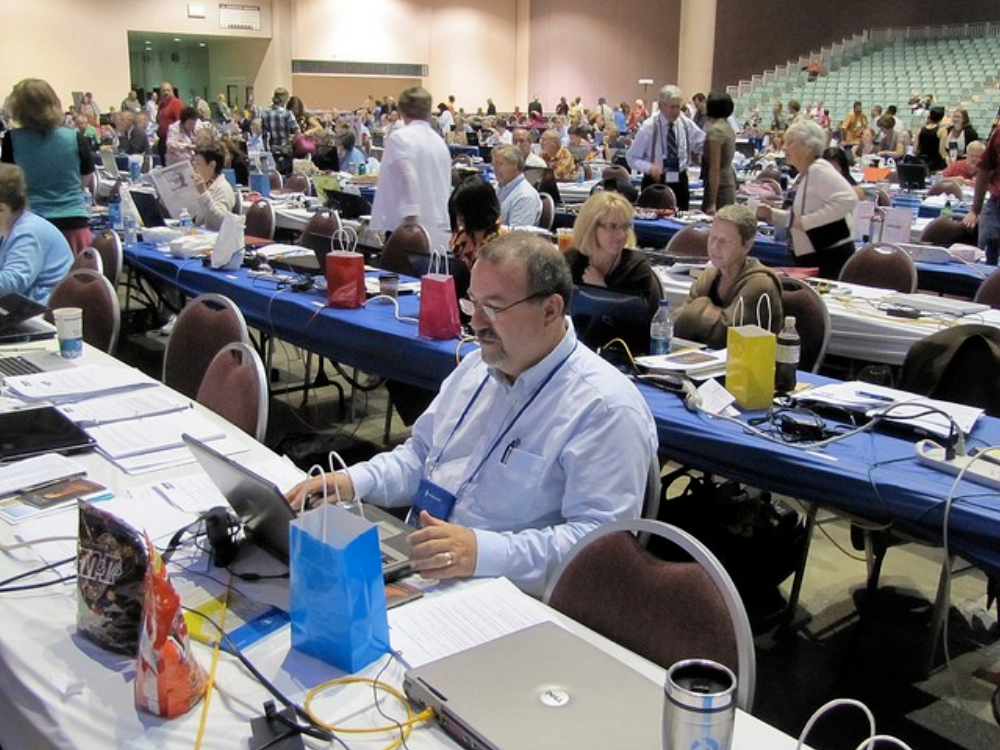}
        \quad\includegraphics[width=0.35\linewidth]{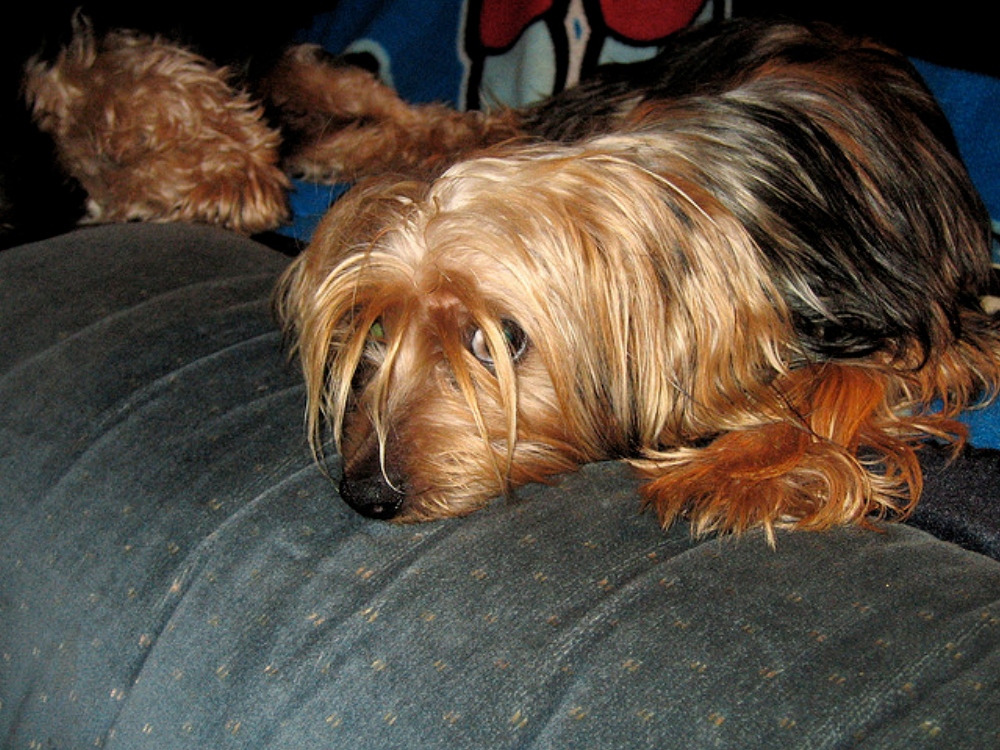}
        
        \hspace{-4mm}\includegraphics[width=0.4\linewidth]{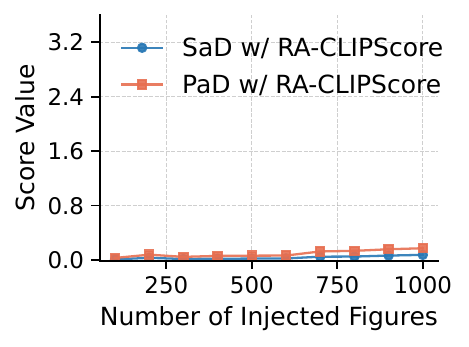}
        \hspace{-2mm}\includegraphics[width=0.4\linewidth]{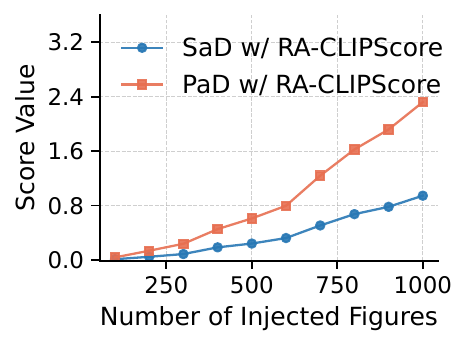}
    \end{minipage}
    \caption{\small{Top: example COCO images containing \textit{chair}, typically in complex scenes. Bottom: SaD and PaD responses as a function of the fraction of injected images. Bottom-left: injecting original COCO images (control group). Bottom-right: injecting \textit{chair} images from COCO.}}
    \label{fig:coco_injection}
\end{figure}

To validate efficacy beyond CelebA-style facial attributes, we conduct experiments analogous to Sec.\ref{subsubsec:toy_injection} by injecting different attributes into a $50$k-image COCO base dataset, using original COCO images as the control group. We use the $80$ COCO class names as attributes for evaluation. An example with injected \textit{chair} images is shown in Fig. \ref{fig:coco_injection}. The results closely mirror those observed on CelebA in Fig. \ref{fig:result_injection}. As the proportion of injected \textit{chair} images increases, both SaD and PaD increase monotonically. In contrast, when randomly sampled COCO images are injected, which preserve the original distribution, SaD and PaD remain nearly constant. Compared with the CelebA results, the curves are less smooth due to the greater complexity of COCO images, which often contain multiple objects (examples in Fig. \ref{fig:coco_injection}). Nevertheless, our metrics remain effective under these more complex conditions.


\subsection{Dataset Comparison}
\label{subsec:sup-mat_dataset_compare}

\begin{figure}[t]
\centering
\begin{subfigure}[t]{0.20\linewidth}
\centering
\includegraphics[width=\linewidth]{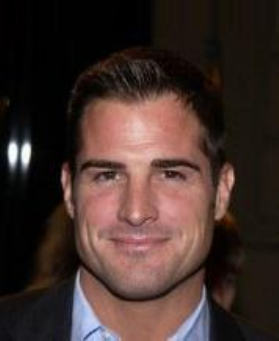}
\end{subfigure}
\begin{subfigure}[t]{0.20\linewidth}
\centering
\includegraphics[width=\linewidth]{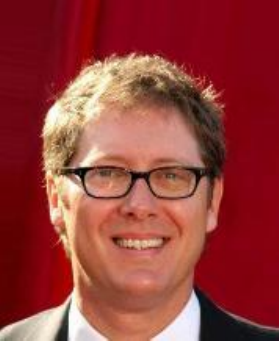}
\end{subfigure}
\begin{subfigure}[t]{0.20\linewidth}
\centering
\includegraphics[width=\linewidth]{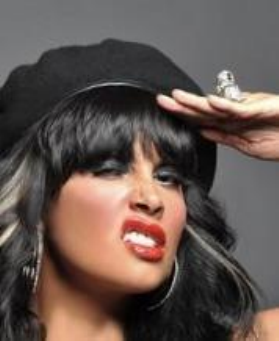}
\end{subfigure}
\begin{subfigure}[t]{0.20\linewidth}
\centering
\includegraphics[width=\linewidth]{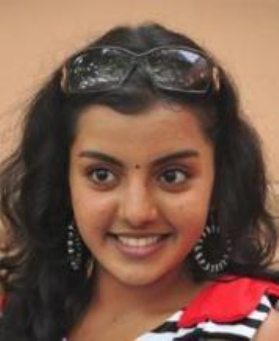}
\end{subfigure}
\caption{\small Example images from CelebA (left to right): man without eyeglasses, man with eyeglasses, woman without eyeglasses, woman with eyeglasses.}
\label{fig:sup-mat_dataset_compare}
\end{figure}

In this section, we provide details for dataset comparison experiments from Sec.\ref{subsubsec:toy_dataset_compare}. Example images used in the experiments are provided in Fig. \ref{fig:sup-mat_dataset_compare}. The list of used attributes is provided in Tab.\ref{tab:sup-mat_celeba_attributes}.

\begin{figure}[t]
    \centering
    \begin{subfigure}[t]{0.8\linewidth}
        \centering
        \includegraphics[width=\linewidth]{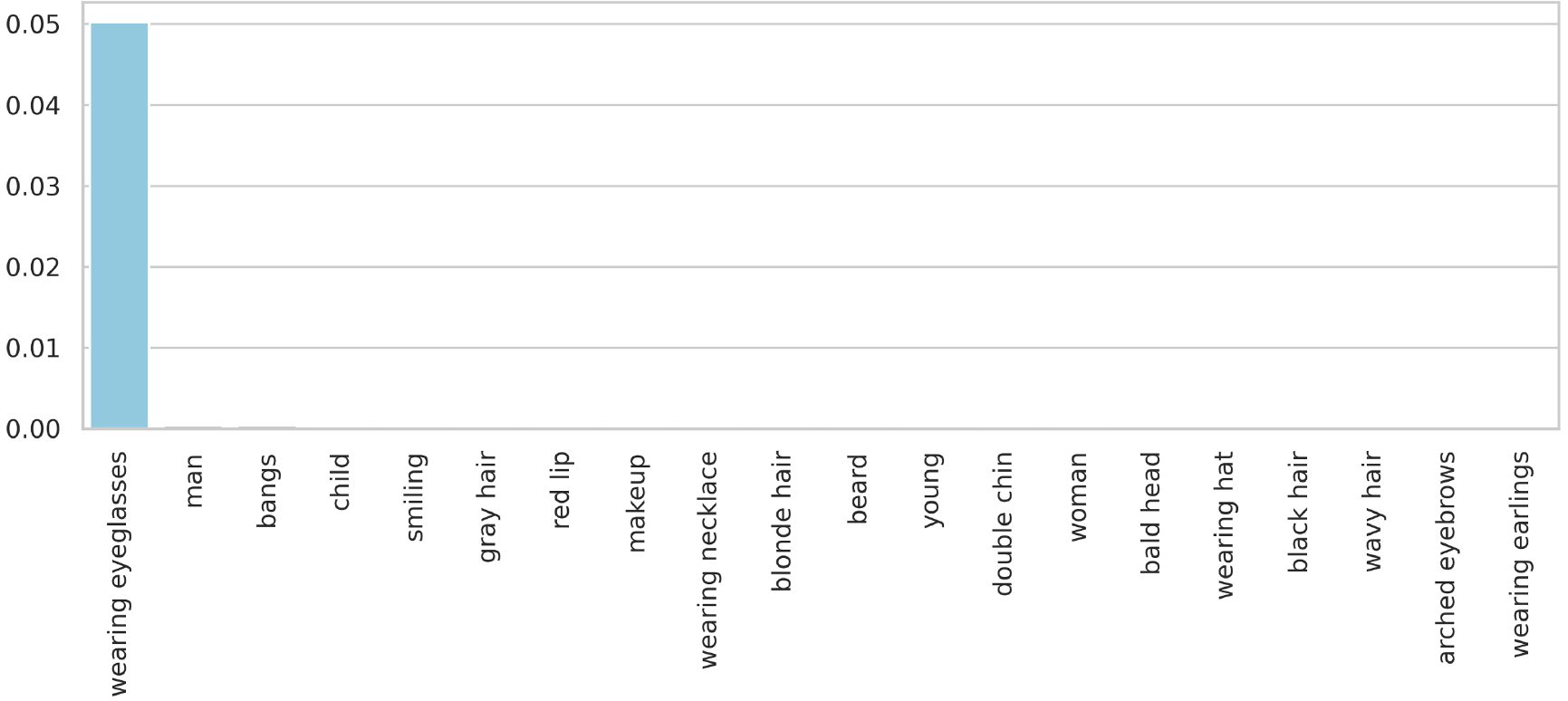}
    \end{subfigure}
    \begin{subfigure}[t]{0.8\linewidth}
        \centering
        \includegraphics[width=\linewidth]{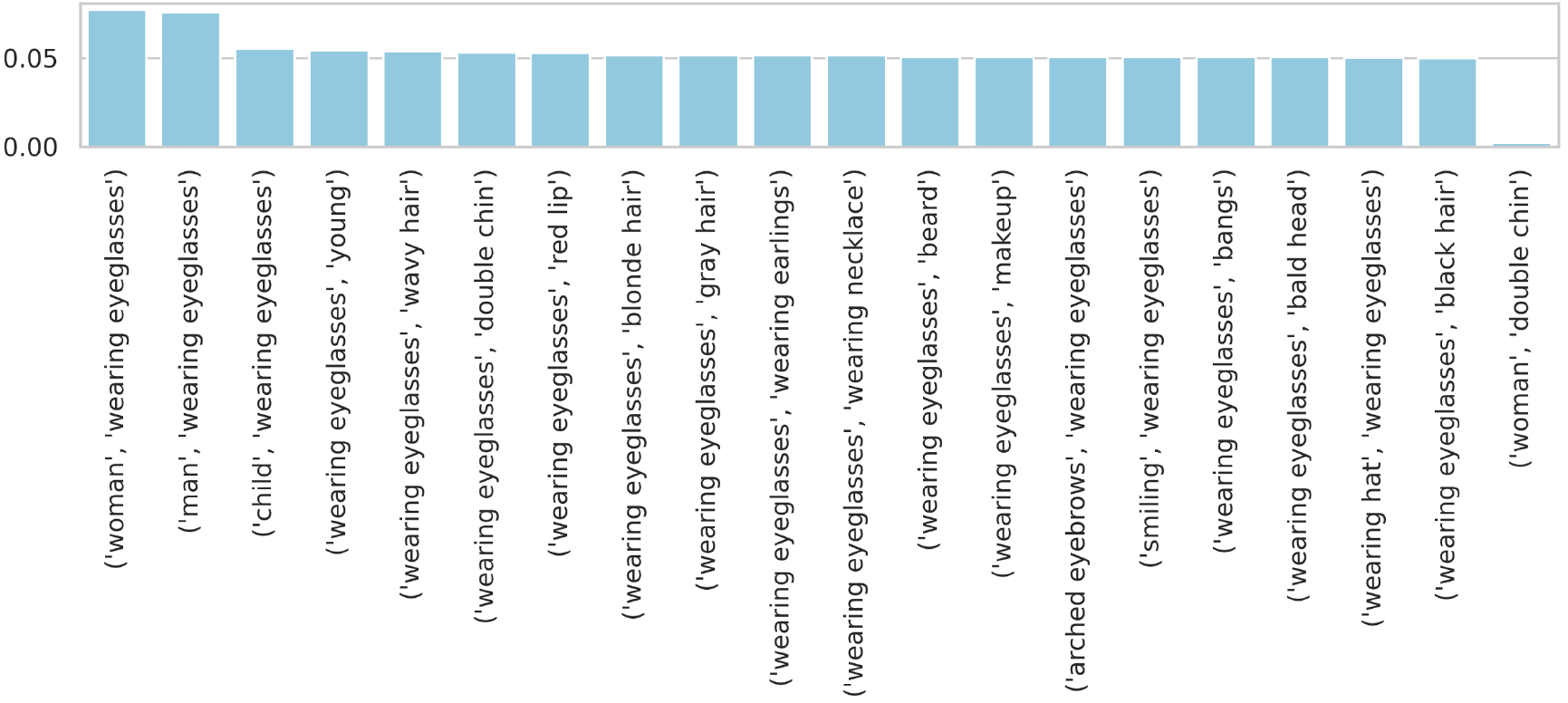}
    \end{subfigure}
    \caption{\small Visualization of the top attributes and attribute pairs with the largest SaD and PaD scores. \textit{Eyeglasses} and \textit{gender} consistently rank among the top two attributes or attribute pairs, demonstrating successful detection of reversed attribute combinations.}
    \label{fig:sup-mat_dataset_compare_bar}
\end{figure}

In addition to the numerical values reported in Tab.\ref{tab:inverse_attr_coupling}, Fig. \ref{fig:sup-mat_dataset_compare_bar} highlights which attributes (or attribute pairs) contribute most to the divergence between datasets. For SaD, the attribute \textit{wearing eyeglasses} appears at the top with a significantly larger score than all other attributes. For PaD, the highest contributions come from the pair \textit{(gender, wearing eyeglasses)}, indicating successful detection of the reverse attribute combination.

\subsubsection{Robustness to Prompt Templates}
\label{subsec:sup-mat_prompt_robustness}

To assess the sensitivity of RA-CLIPScore to prompt wording, we evaluate four positive--negative prompt pairs:
(1) ``This is a photo of $t_i$.'' / ``This is a photo without $t_i$.''
(2) ``There is $t_i$ in the scene.'' / ``There is no $t_i$ in the scene.''
(3) ``The image contains $t_i$.'' / ``The image does not contain $t_i$.''
(4) ``It is $t_i$ in the picture.'' / ``It is not $t_i$ in the picture.''

Following the inverse attribute-coupling experiment in Sec.~\ref{subsubsec:toy_dataset_compare}, we compare Datasets A and C using SaD and PaD under each prompt template. We assess ranking consistency using Kendall's coefficient of concordance ($W$) and Rank-Biased Overlap (RBO). SaD achieves $W=0.454$, indicating moderate agreement, while PaD exhibits stronger consistency with $W=0.615$.

The pairwise RBO matrices across the four prompt templates are:

\begin{equation}
\mathrm{RBO}_{\mathrm{SaD}} =
\begin{bmatrix}
1 & 0.67 & 0.65 & 0.70 \\
0.67 & 1 & 0.68 & 0.72 \\
0.65 & 0.68 & 1 & 0.78 \\
0.70 & 0.72 & 0.78 & 1
\end{bmatrix},
\quad\mathrm{RBO}_{\mathrm{PaD}} =
\begin{bmatrix}
1 & 0.73 & 0.52 & 0.75 \\
0.73 & 1 & 0.59 & 0.90 \\
0.52 & 0.59 & 1 & 0.64 \\
0.75 & 0.90 & 0.64 & 1
\end{bmatrix}.
\label{eq:rbo}
\end{equation}

RBO emphasizes agreement among highly ranked entries. The results show substantial overlap across prompt templates, with pairwise RBO values ranging from $0.65$ to $0.78$ for SaD and from $0.52$ to $0.90$ for PaD. Notably, the top-ranked attribute under SaD is identical across all four templates. For PaD, the highest-ranked attribute pairs also remain consistent, with RBO reaching $0.90$ and top-3 overlap reaching $100\%$ for certain template pairs. Overall, these results suggest that the most prominent attribute-level and pairwise discrepancies are relatively stable across prompt formulations, despite variations in the complete rankings.

\subsection{Ablation Study}
\label{subsec:sup-mat_ablation}

In Sec.~\ref{subsec:ablation_study}, we evaluate the contributions of individual components of RA-CLIPScore across 20 CelebA attributes, grouped into three categories according to their semantic characteristics. Here, we provide a more detailed analysis of three representative attributes: \textit{Male} (global), \textit{Eyeglasses} (local), and \textit{Smiling} (other). We first introduce Cohen's $d$, the separation metric used in the main paper, and further validate our findings using two complementary measures: Hellinger distance (HD) and one minus the overlapping coefficient ($1-\mathrm{OVL}$). As shown in Tab.~\ref{tab:sup-mat_ablation_study}, RA-CLIPScore consistently outperforms CLIPScore and HCS across all three attributes and evaluation criteria. The ablation results further reveal attribute-dependent contributions of different components, highlighting the complementary roles of global and local representations. The corresponding score distributions are visualized in Figs.~\ref{fig:sup-mat_ablation_male}--\ref{fig:sup-mat_ablation_smiling}.

\begin{table}[t]
\centering
\setlength{\tabcolsep}{2.8pt}
\renewcommand{\arraystretch}{0.94}
\begin{tabular}{@{}lccccccc@{}}
\toprule
\textbf{Criteria} & \textbf{Attr.} & \makecell{\textbf{CLIP} \\\ \textbf{Score}} & \textbf{HCS} & \makecell{\textbf{Ours} \\\ \textbf{(Full)}} & \makecell{\textbf{w/o} \\\ \textbf{DP}} & \makecell{\textbf{w/o} \\\ \textbf{LT}}& \makecell{\textbf{w/o} \\\ \textbf{GE}} \\
\midrule
C's d   & \multirow{3}{*}{Male}
  & 1.22 & 1.55 & 3.16 & 1.11 &  \textbf{3.35} & 0.85 \\
HD  
  &  & 0.41 & 0.51 & 0.80 & 0.40 & \textbf{0.87} & 0.30 \\
$1-\text{OVL}$  
  &  & 0.47 & 0.57 & 0.88 & 0.42 & \textbf{0.90} & 0.34 \\
\midrule
C's d   & \multirow{3}{*}{\makecell{Eye- \\ glasses}}
  & 2.05 & 2.01 &  3.03 & 1.12 & 2.07 & \textbf{4.41} \\
HD  
  &  & 0.63 & 0.63 &  0.78 & 0.38 & 0.67 & \textbf{0.82} \\
$1-\text{OVL}$ 
  &  & 0.73 & 0.71 & 0.85 & 0.41 & 0.74 & \textbf{0.91} \\
\midrule
C's d  & \multirow{3}{*}{Smiling}
  & 1.67 & 1.70 & \textbf{2.45} & 1.42 & 2.18 & 2.44 \\
HD  
  &  & 0.53 & 0.54 & \textbf{0.68} & 0.46 & 0.64 & 0.66 \\
$1-\text{OVL}$  
  &  & 0.60 & 0.60 & \textbf{0.74} & 0.52 & 0.71 & 0.72 \\
\bottomrule
\end{tabular}
\caption{\small{Ablation study under different settings and criteria: Cohen's d (C's d), Hellinger distance (HD) and 1-OVL . Higher values indicate better separation between the score distributions of positive and negative images. Our RA-CLIPScore consistently outperforms HCS and CLIPScore.}
}
\label{tab:sup-mat_ablation_study}
\end{table}

\subsubsection{Distribution Separation Metrics}
\label{subsubsec:sup-mat_ablation_metrics}

We report results using three distribution separation criteria, namely Cohen's d (C's d), Hellinger distance (HD), and $1$-OVL, to quantify how well the score distributions are separated for attribute presence and absence. We detail the computation of these criteria here.

We define positive images as those where the attribute is present and negative images as those where the attribute is absent. For each attribute $t$, we compute scores for all positive and negative images using Eq.\eqref{eq:clipscore} for CLIPScore, Eqs.\eqref{eq:hcs} for HCS, and Eq.\eqref{eq:raclipscore} for our RA-CLIPScore, as well as the corresponding variants with components removed according to Tab.\ref{tab:sup-mat_ablation_study_settings}. This yields one set of scores for positive images and another for negative images. We then estimate the empirical score distributions using Kernel Density Estimation (KDE) with Gaussian kernels, obtaining two continuous distributions over the score $r$ ($r$ denotes CLIPScore, HCS, or RA-CLIPScore): $p(r)$ for positive scores and $q(r)$ for negative scores. The separation metrics are subsequently computed based on these distributions.

Cohen's d (C's d) is defined as follows:
\begin{equation}
\text{Cohen's } d = \frac{|\mu_p - \mu_n|}{\sqrt{\tfrac{1}{2}(\sigma_p^2 + \sigma_n^2)}}
\label{eq:supp-mat_cohens_d}
\end{equation}
\noindent
where $\mu_p$ and $\mu_n$ denote the means, and $\sigma_p$ and $\sigma_n$ the standard deviations, of the positive and negative score distributions, respectively. Cohen's d measures how far apart the two distributions are relative to their pooled variance and larger values indicate stronger separation.

Hellinger Distance measures how much two probability distributions differ in shape. It quantifies the amount of overlap between them: a value of 0 means the distributions are identical, while values closer to 1 indicate less overlap and stronger separation. It's defined below:
\begin{equation}
\text{Hellinger Distance} = \sqrt{1 - \int_r \sqrt{p(r)\, q(r)}\,dr}
\label{eq:supp-mat_hellinger_distance}
\end{equation}

$1 -$OVL measures how well two distributions can be separated by their non-overlapping area.
The overlap coefficient is defined as: 
\begin{equation}
    \text{OVL} =  \int_r \min(p(r), q(r))\,dr,
\end{equation}
which quantifies the shared probability mass between two distributions. Thus, $1 -$OVL reports the non-overlapping region, where larger values indicate stronger separation.

\subsubsection{Distribution of scores}
\label{subsubsec:sup-mat_ablation_figures}

\begin{figure*}[t]
    \centering
    \begin{subfigure}[t]{0.32\textwidth}
        \centering
        \includegraphics[width=\linewidth]{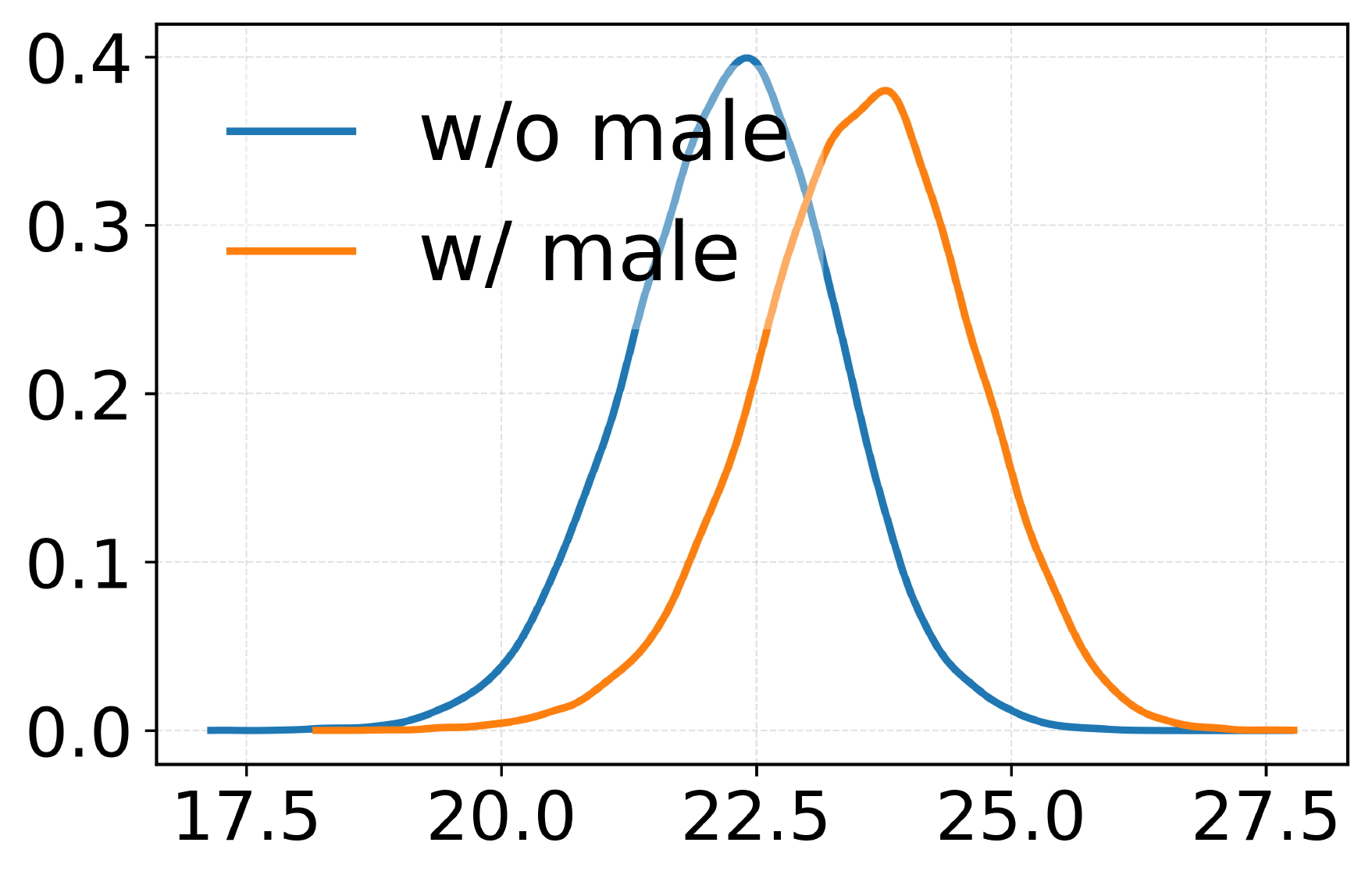}
        \caption{CLIPScore}
        \label{fig:ablation_male_clipscore}
    \end{subfigure}
    \begin{subfigure}[t]{0.32\textwidth}
        \centering
        \includegraphics[width=\linewidth]{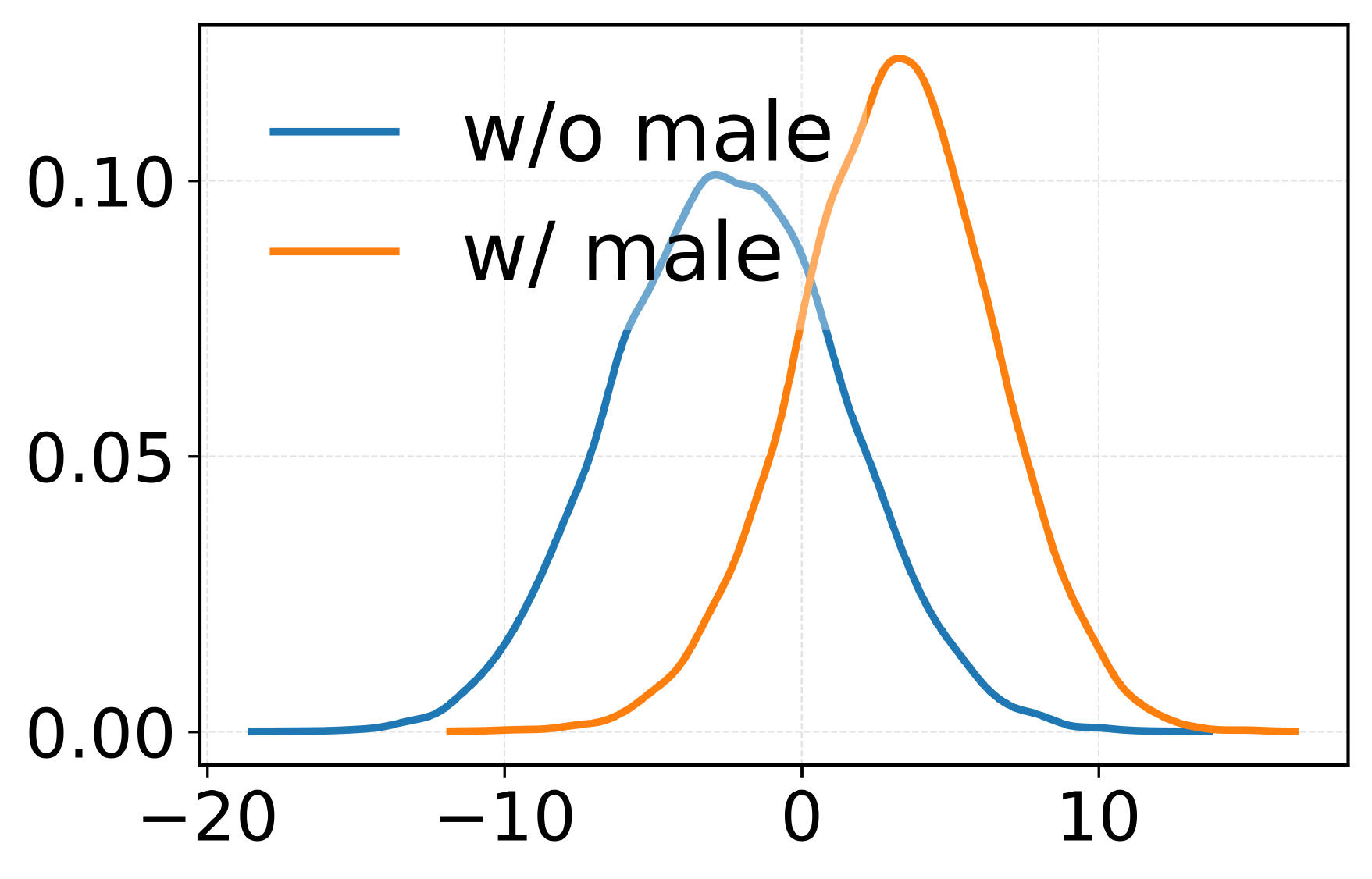}
        \caption{HCS}
        \label{fig:ablation_male_hcs}
    \end{subfigure}
    \begin{subfigure}[t]{0.32\textwidth}
        \centering
        \includegraphics[width=\linewidth]{figures/Male_EvalCS_local_rn_full.pdf}
        \caption{Ours}
        \label{fig:ablation_male_ours}
    \end{subfigure}

    \begin{subfigure}[t]{0.32\textwidth}
        \centering
        \includegraphics[width=\linewidth]{figures/Male_EvalCS_local_rn_without_dual_prompts.pdf}
        \caption{w/o Dual Prompts}
        \label{fig:ablation_male_wo_dual}
    \end{subfigure}
    \begin{subfigure}[t]{0.32\textwidth}
        \centering
        \includegraphics[width=\linewidth]{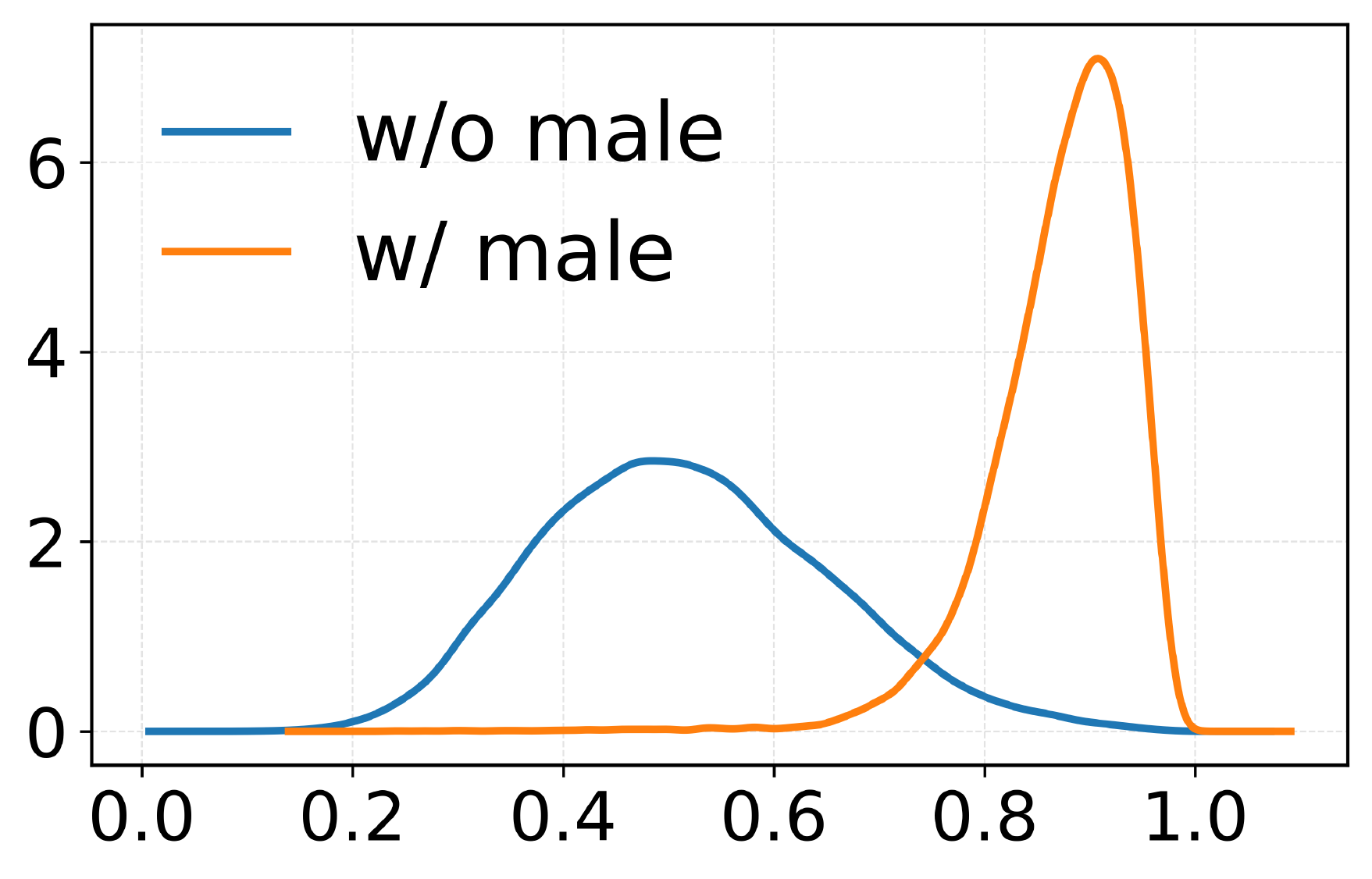}
        \caption{w/o Local Tokens}
        \label{fig:ablation_male_wo_local}
    \end{subfigure}
    \begin{subfigure}[t]{0.32\textwidth}
        \centering
        \includegraphics[width=\linewidth]{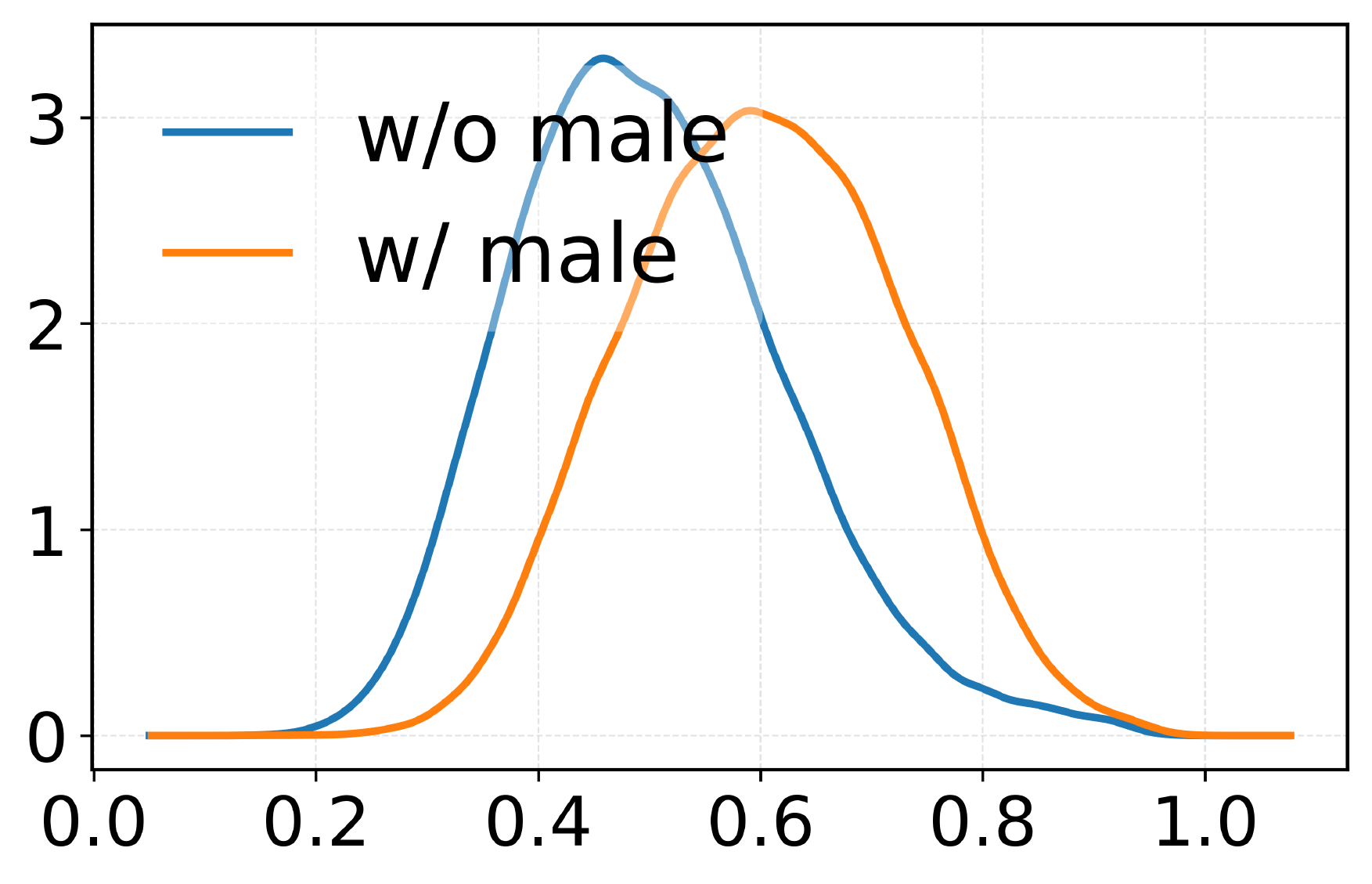}
        \caption{w/o Global Embeddings}
        \label{fig:ablation_male_wo_global}
    \end{subfigure}

    \caption{
    Ablation study on the \textit{male} attribute. 
    }
    \label{fig:sup-mat_ablation_male}
\end{figure*}

\begin{figure*}[t]
    \centering
    \begin{subfigure}[t]{0.32\textwidth}
        \centering
        \includegraphics[width=\linewidth]{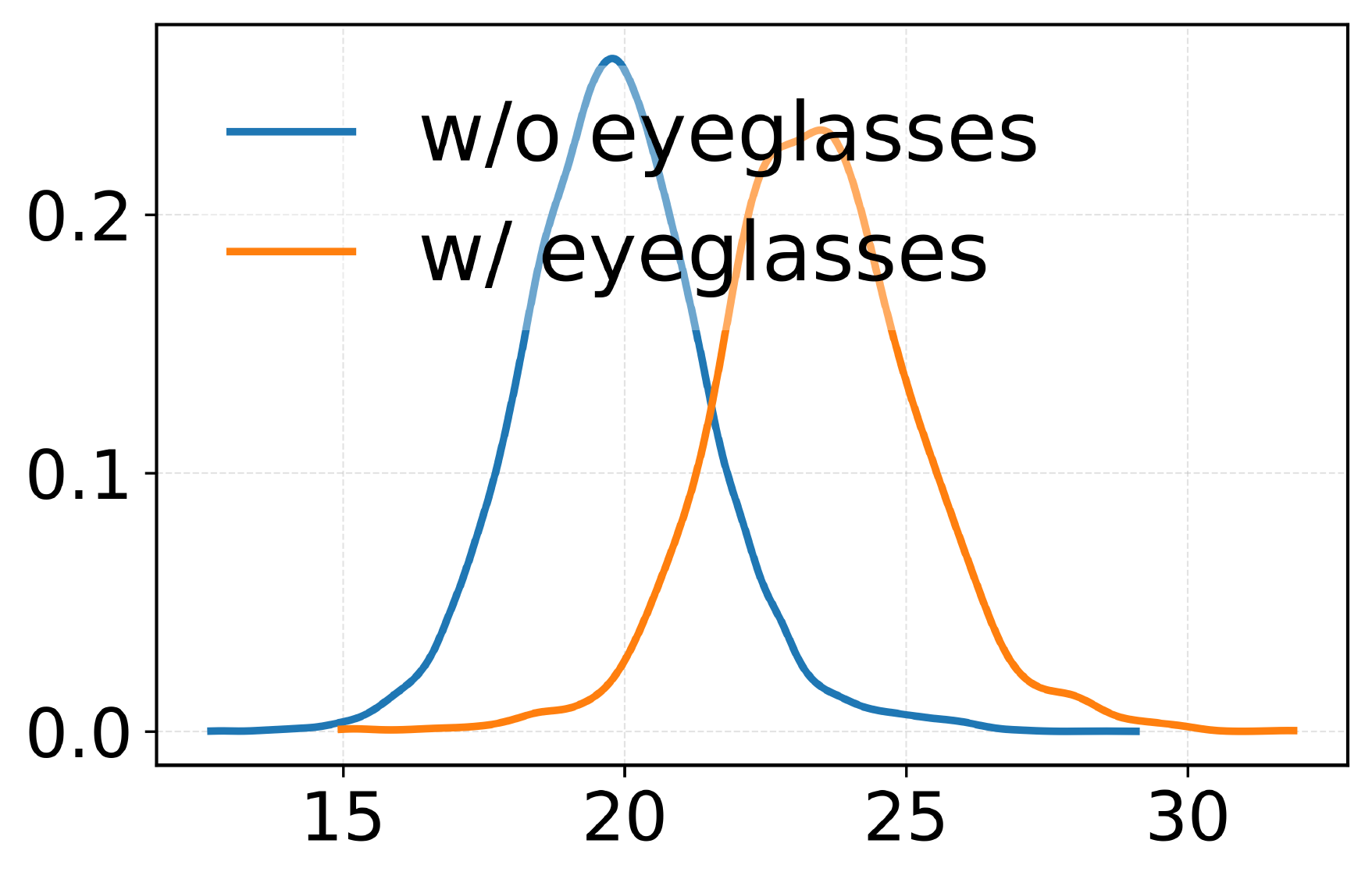}
        \caption{CLIPScore}
    \end{subfigure}
    \begin{subfigure}[t]{0.32\textwidth}
        \centering
        \includegraphics[width=\linewidth]{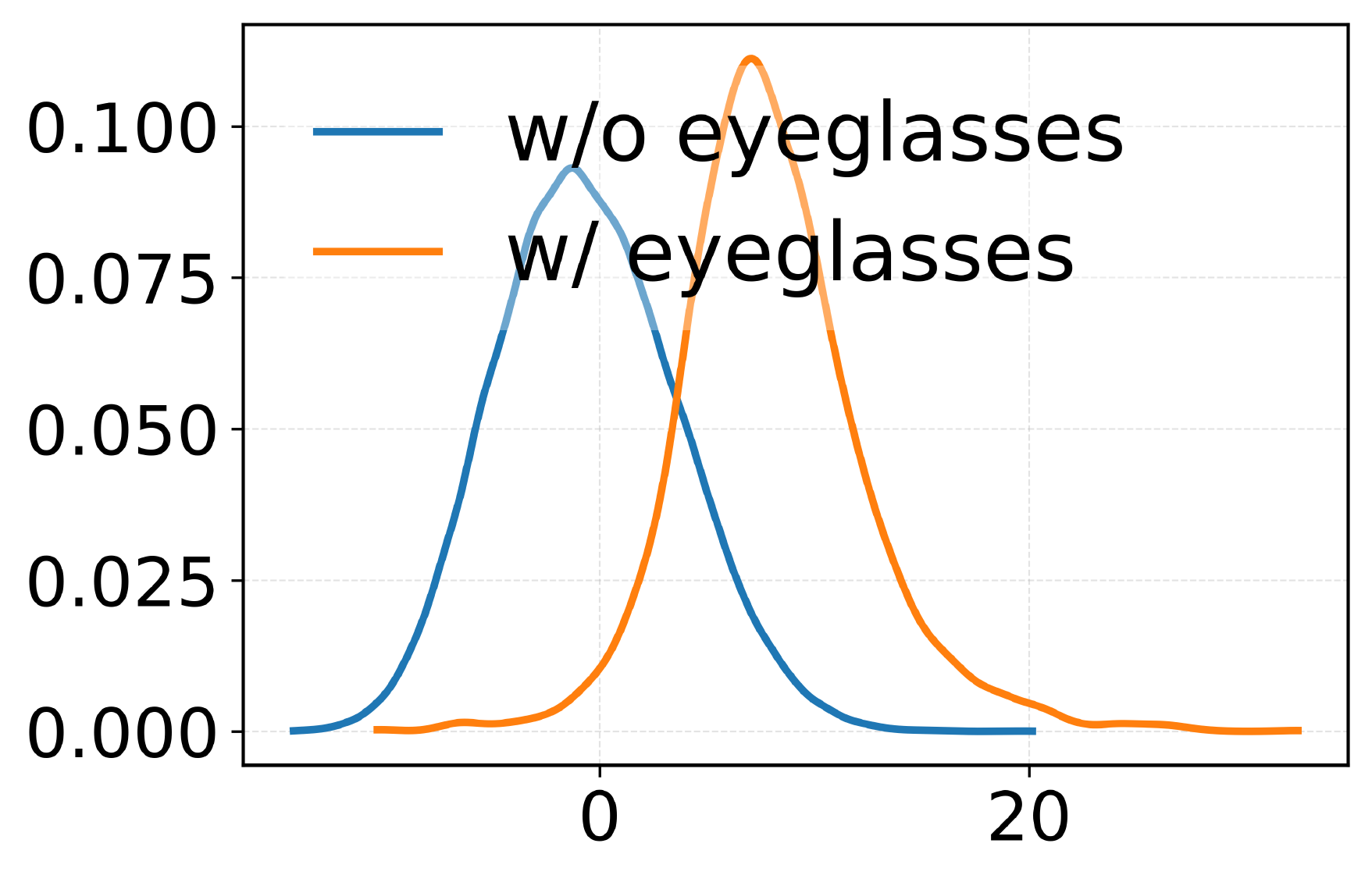}
        \caption{HCS}
    \end{subfigure}
    \begin{subfigure}[t]{0.32\textwidth}
        \centering
        \includegraphics[width=\linewidth]{figures/Eyeglasses_EvalCS_local_rn_full.pdf}
        \caption{Ours}
    \end{subfigure}

    \begin{subfigure}[t]{0.32\textwidth}
        \centering
        \includegraphics[width=\linewidth]{figures/Eyeglasses_EvalCS_local_rn_without_dual_prompts.pdf}
        \caption{w/o Dual Prompts}
    \end{subfigure}
    \begin{subfigure}[t]{0.32\textwidth}
        \centering
        \includegraphics[width=\linewidth]{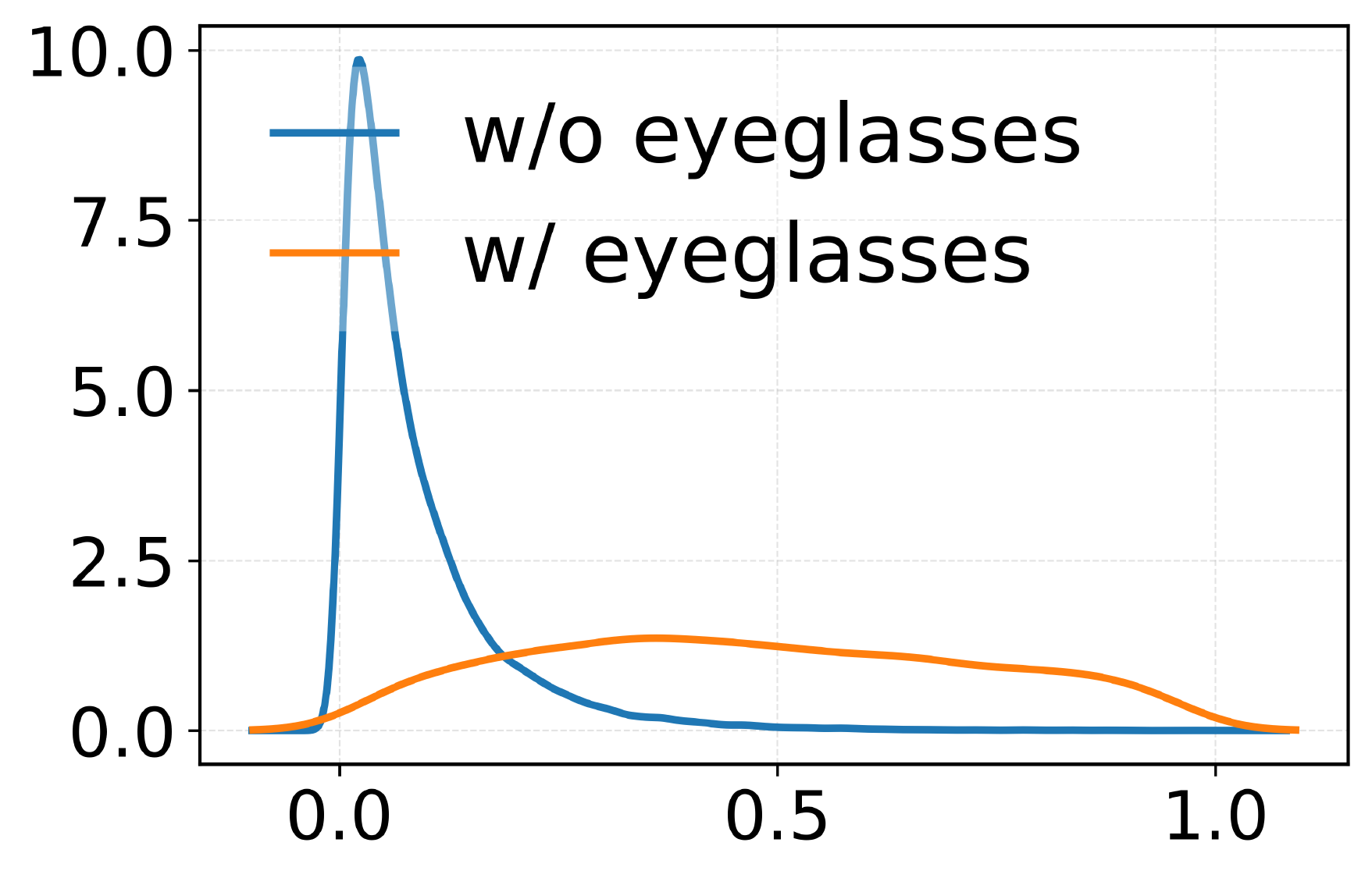}
        \caption{w/o Local Tokens}
    \end{subfigure}
    \begin{subfigure}[t]{0.32\textwidth}
        \centering
        \includegraphics[width=\linewidth]{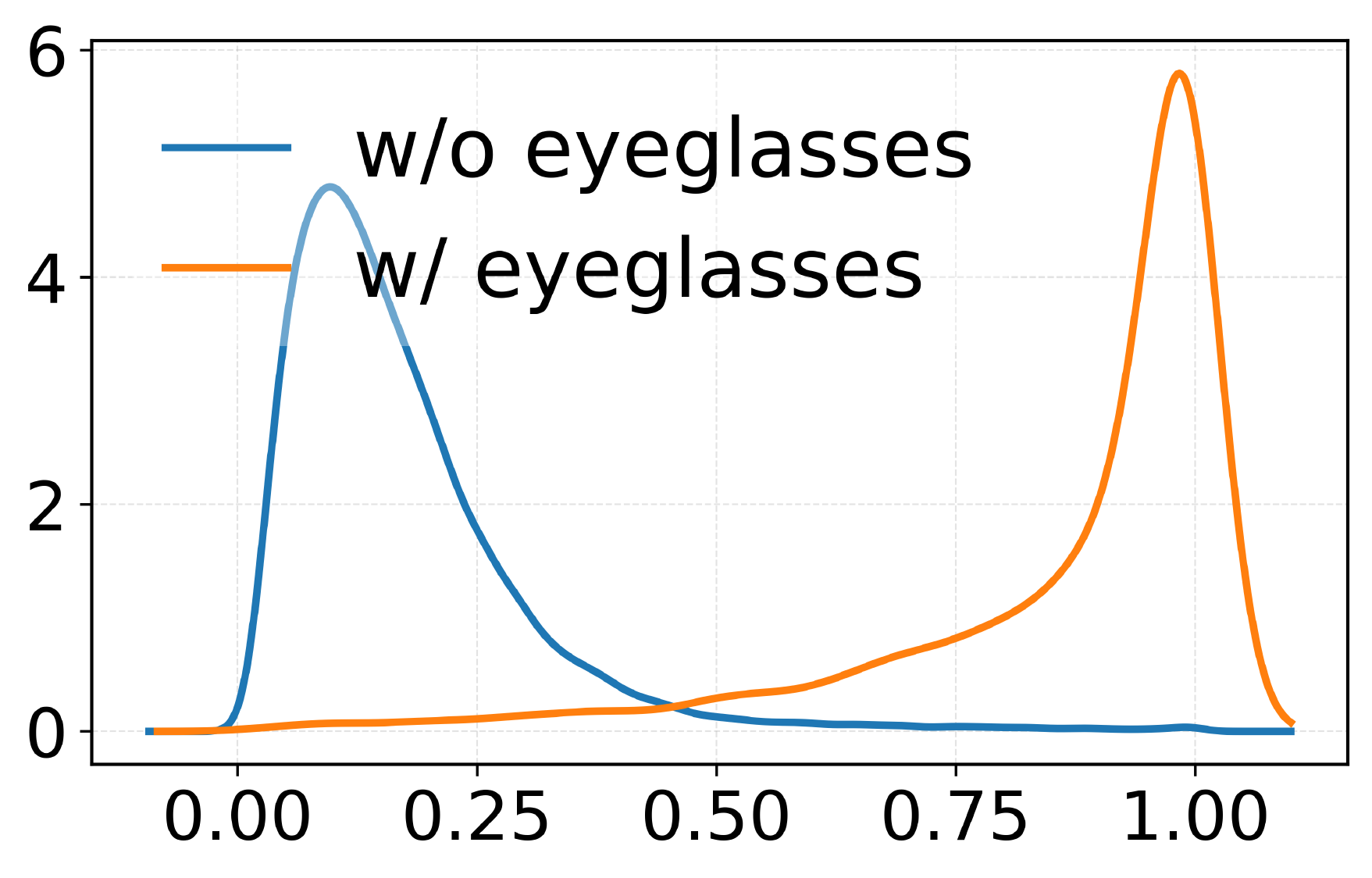}
        \caption{w/o Global Embeddings}
    \end{subfigure}

    \caption{
    Ablation study on the \textit{eyeglasses} attribute. 
    }
    \label{fig:sup-mat_ablation_eyeglasses}
\end{figure*}

\begin{figure*}[t]
    \centering
    \begin{subfigure}[t]{0.32\textwidth}
        \centering
        \includegraphics[width=\linewidth]{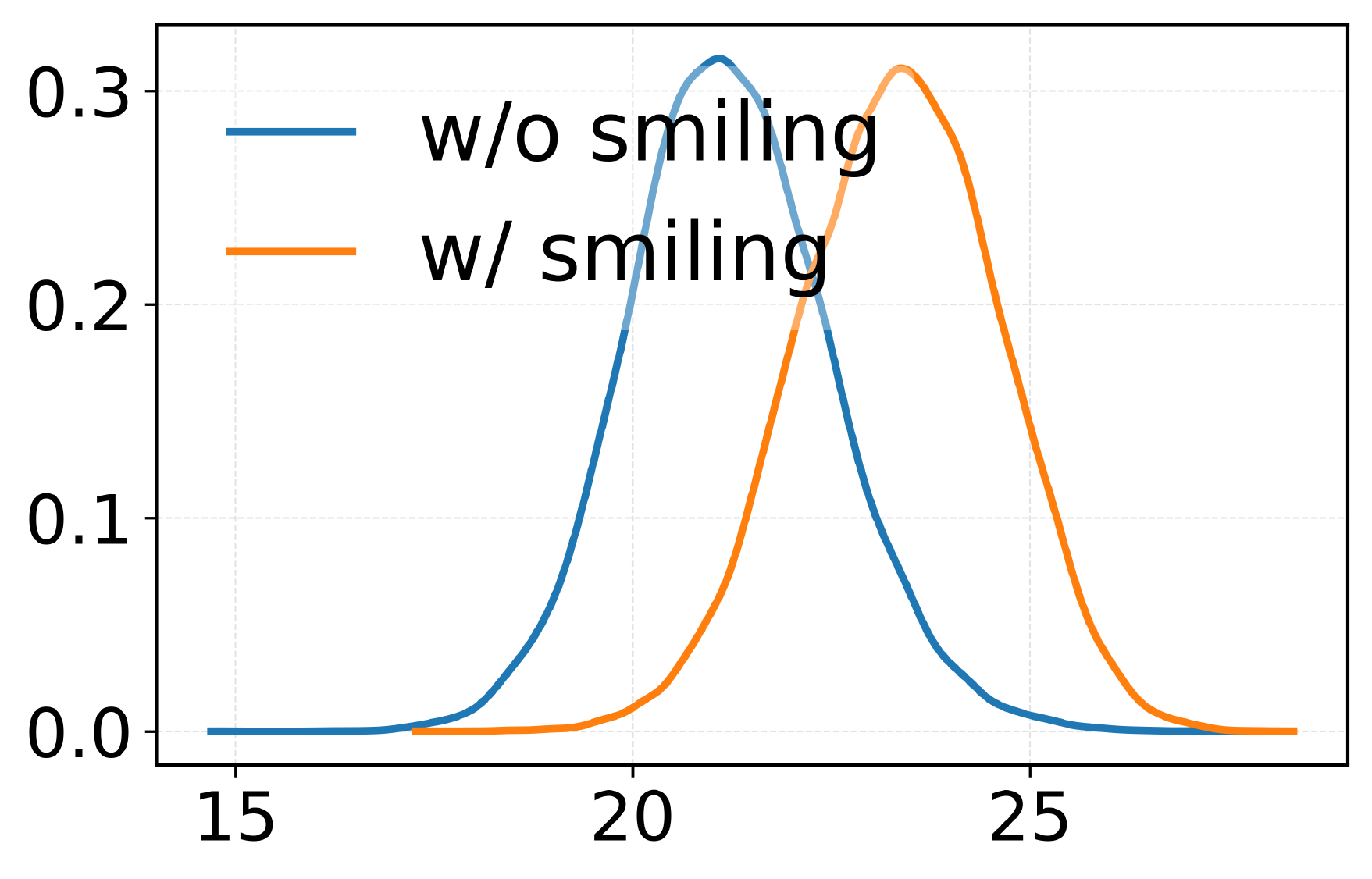}
        \caption{CLIPScore}
    \end{subfigure}
    \begin{subfigure}[t]{0.32\textwidth}
        \centering
        \includegraphics[width=\linewidth]{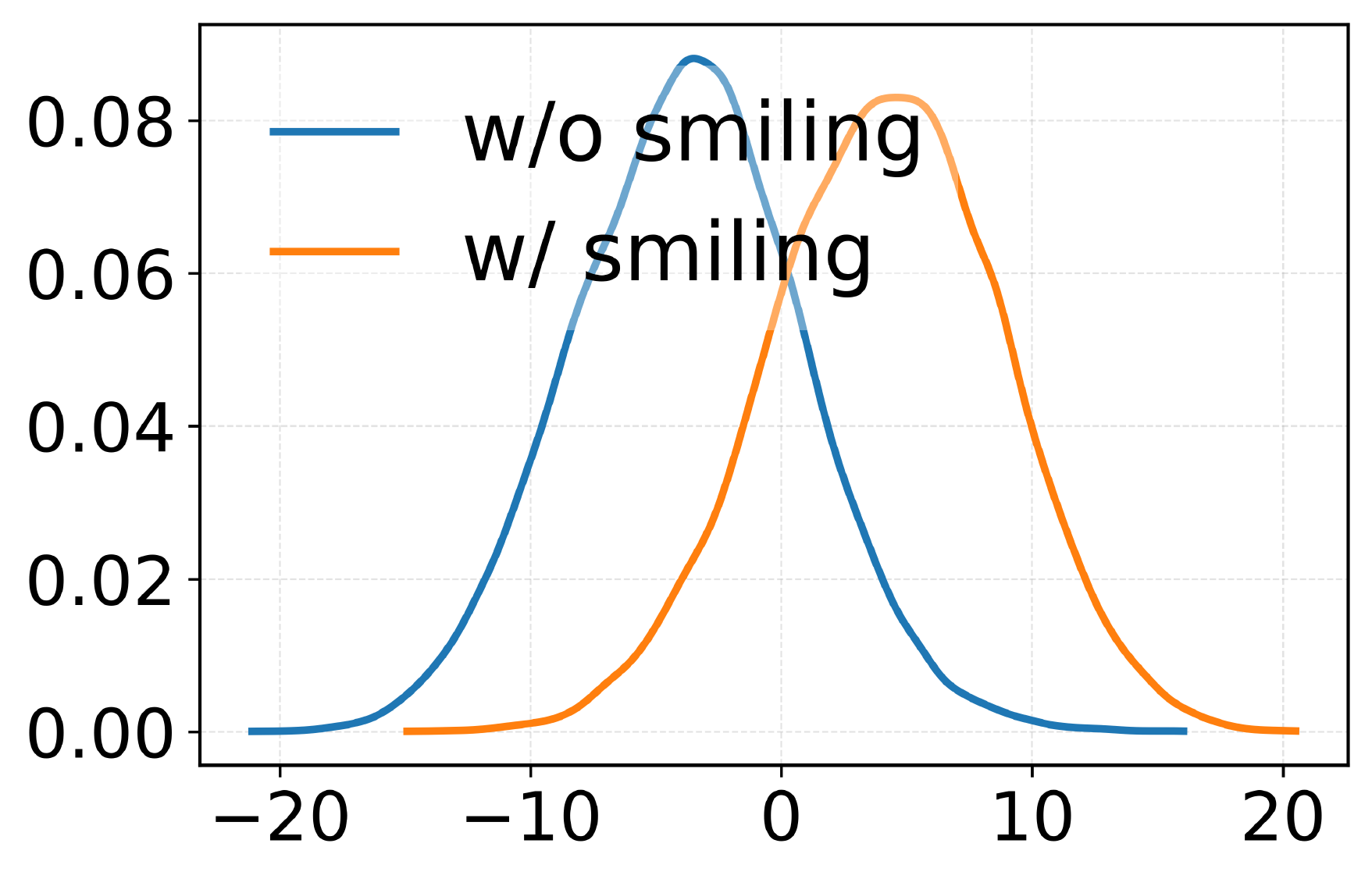}
        \caption{HCS}
    \end{subfigure}
    \begin{subfigure}[t]{0.32\textwidth}
        \centering
        \includegraphics[width=\linewidth]{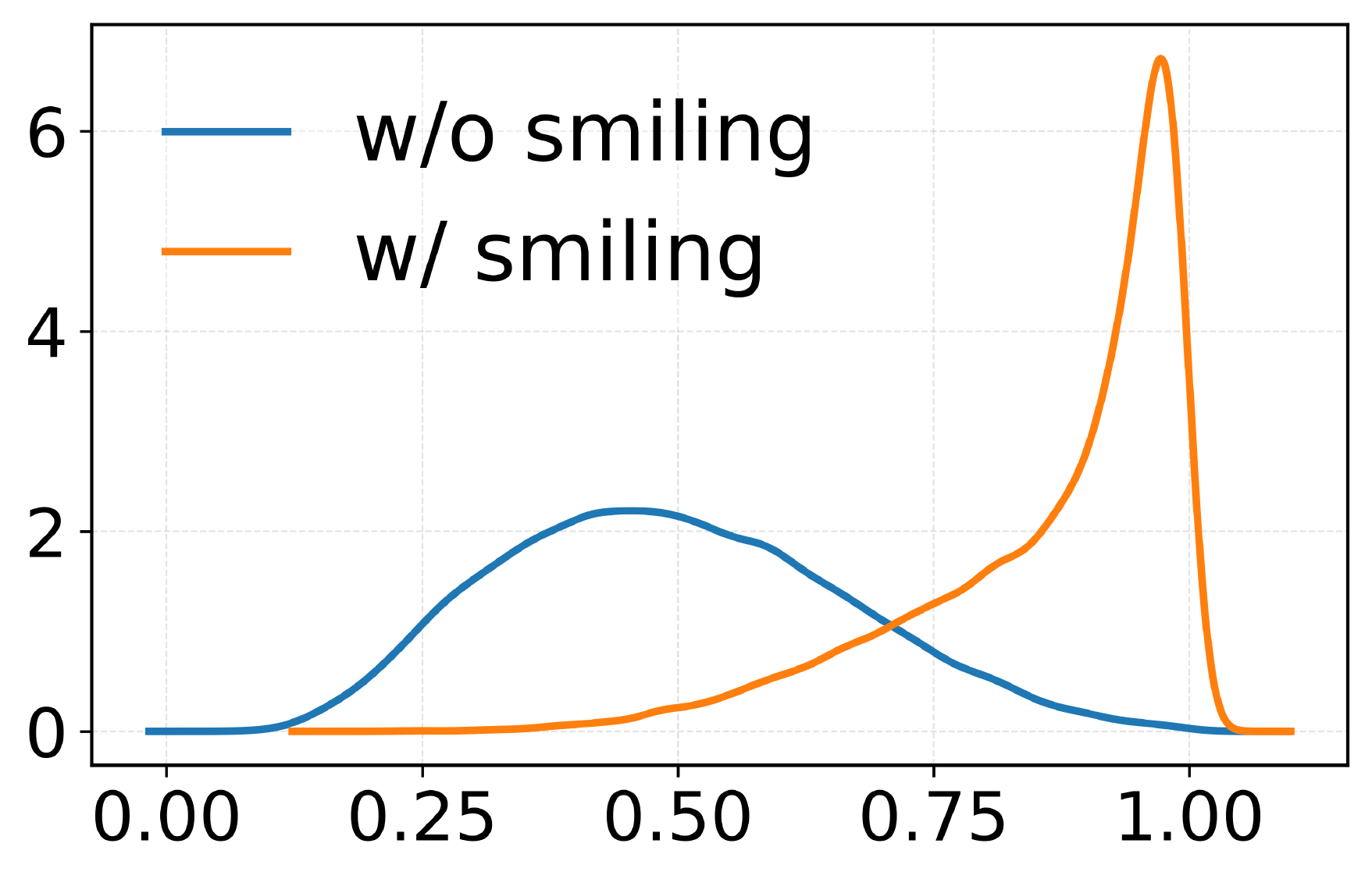}
        \caption{Ours}
    \end{subfigure}

    \begin{subfigure}[t]{0.32\textwidth}
        \centering
        \includegraphics[width=\linewidth]{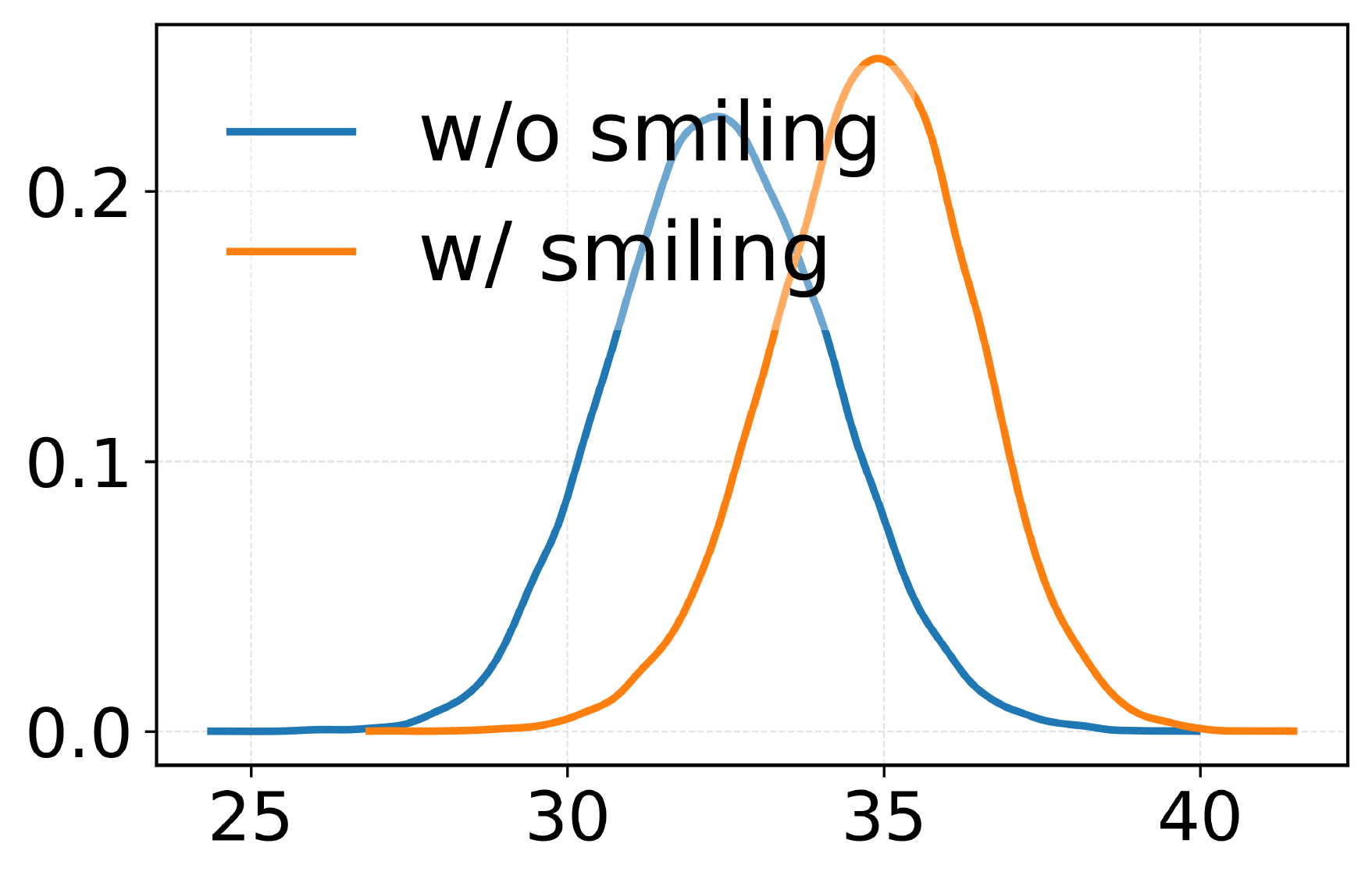}
        \caption{w/o Dual Prompts}
    \end{subfigure}
    \begin{subfigure}[t]{0.32\textwidth}
        \centering
        \includegraphics[width=\linewidth]{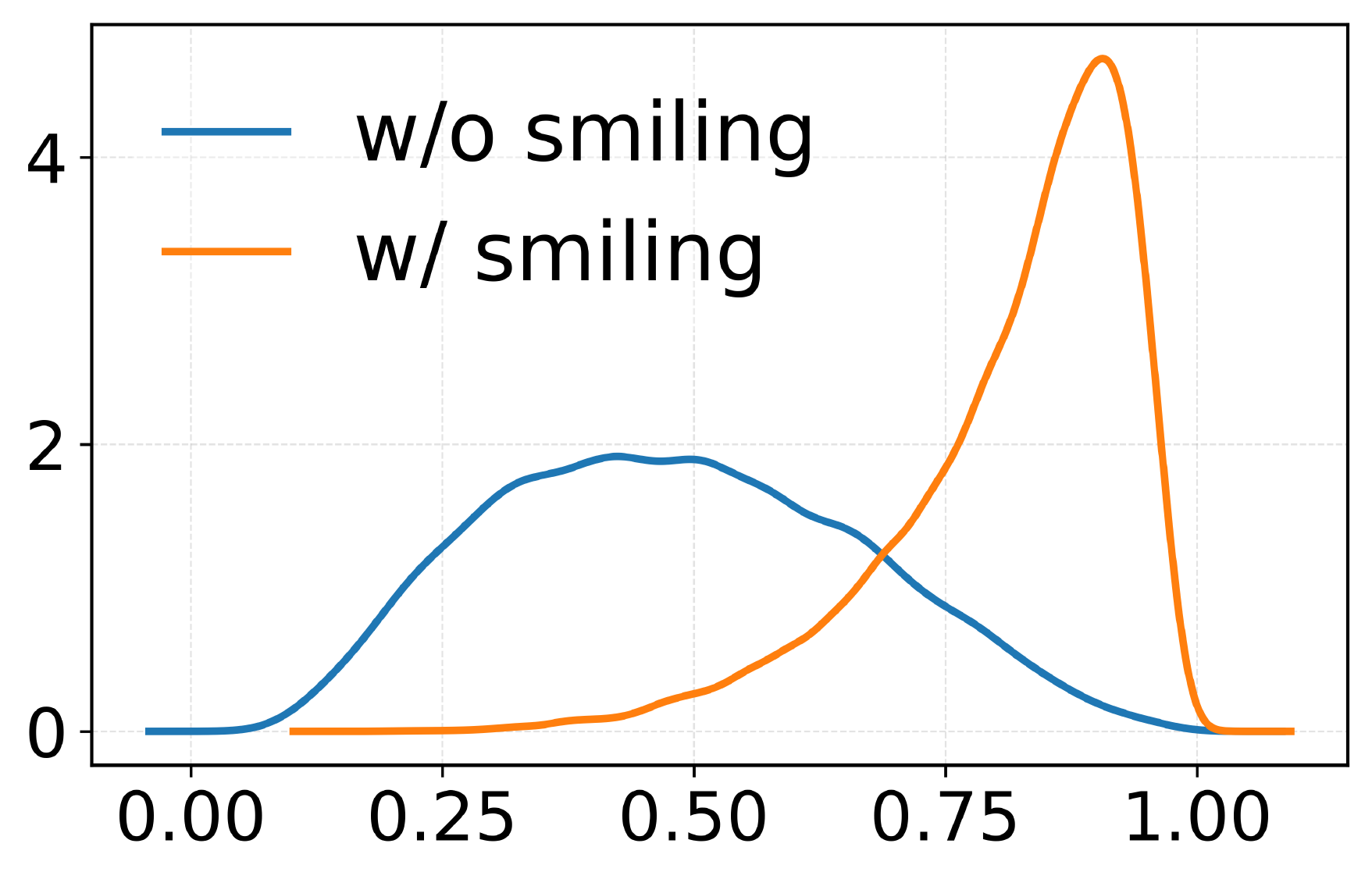}
        \caption{w/o Local Tokens}
    \end{subfigure}
    \begin{subfigure}[t]{0.32\textwidth}
        \centering
        \includegraphics[width=\linewidth]{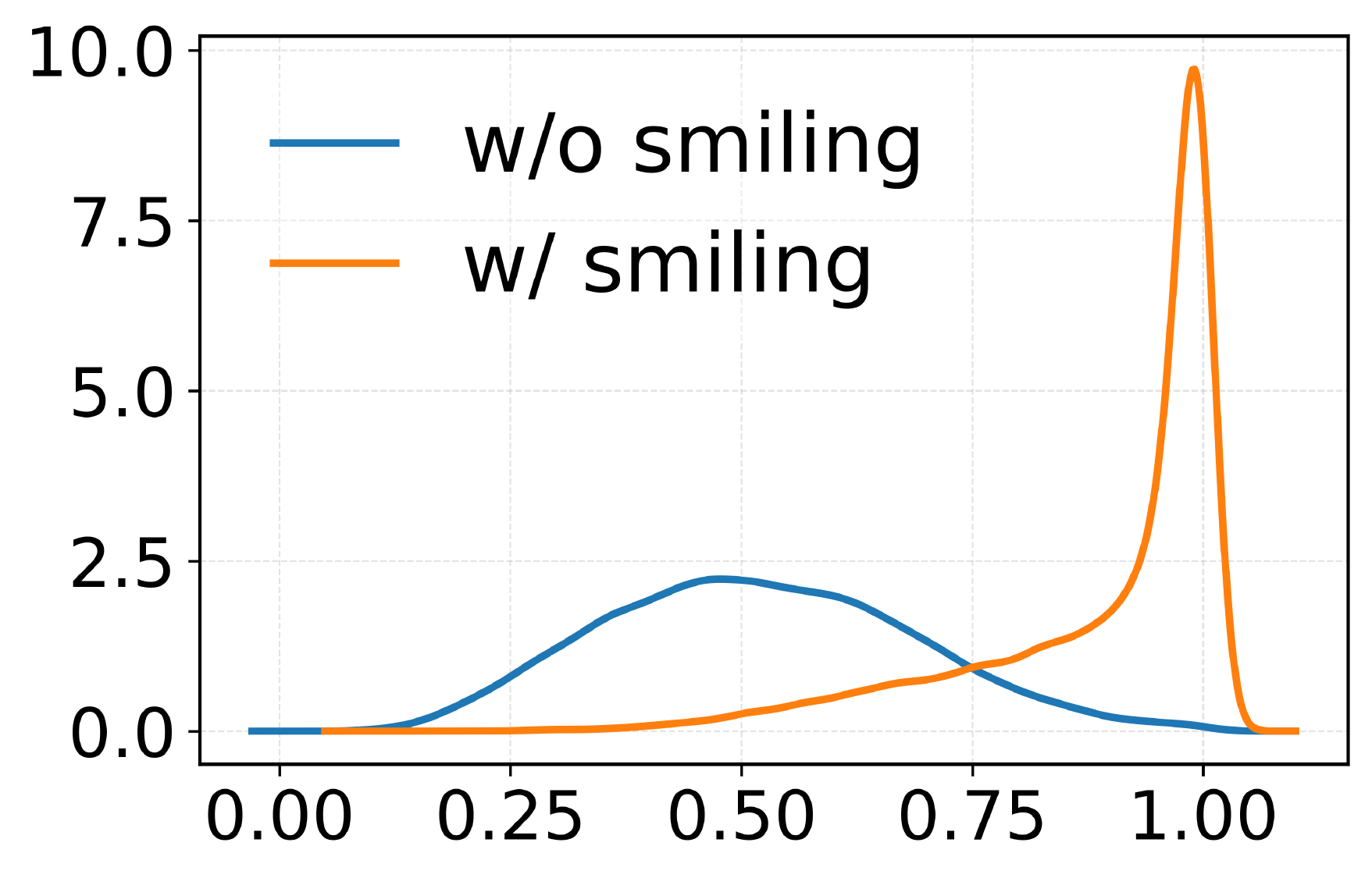}
        \caption{w/o Global Embeddings}
    \end{subfigure}

    \caption{
    Ablation study on the \textit{smiling} attribute. 
    }
    \label{fig:sup-mat_ablation_smiling}
\end{figure*}

We visualize the distributions of positive and negative scores in Fig. \ref{fig:sup-mat_ablation_male}–\ref{fig:sup-mat_ablation_smiling}, allowing us to observe the influence of each component directly. For dominant attributes such as \textit{male}, a reliable decision requires information from multiple regions, and thus the global embedding plays a decisive role. Removing global embedding makes the score distributions overlap (Fig. \ref{fig:sup-mat_ablation_male} (f)). In contrast, for localized attributes such as \textit{wearing eyeglasses}, local features have an overriding effect. Removing local tokens flattens the negative score distribution (Fig. \ref{fig:sup-mat_ablation_eyeglasses} (e)). In addition, dual prompts enlarge the separation between positive and negative score distributions, making the distinction more pronounced.

\subsection{Evaluation of Generative Models}
\label{subsec:sup-mat_eval}

\begin{figure*}[t]
    \centering
    \begin{subfigure}[t]{0.8\linewidth}
        \centering
        \includegraphics[width=\linewidth]{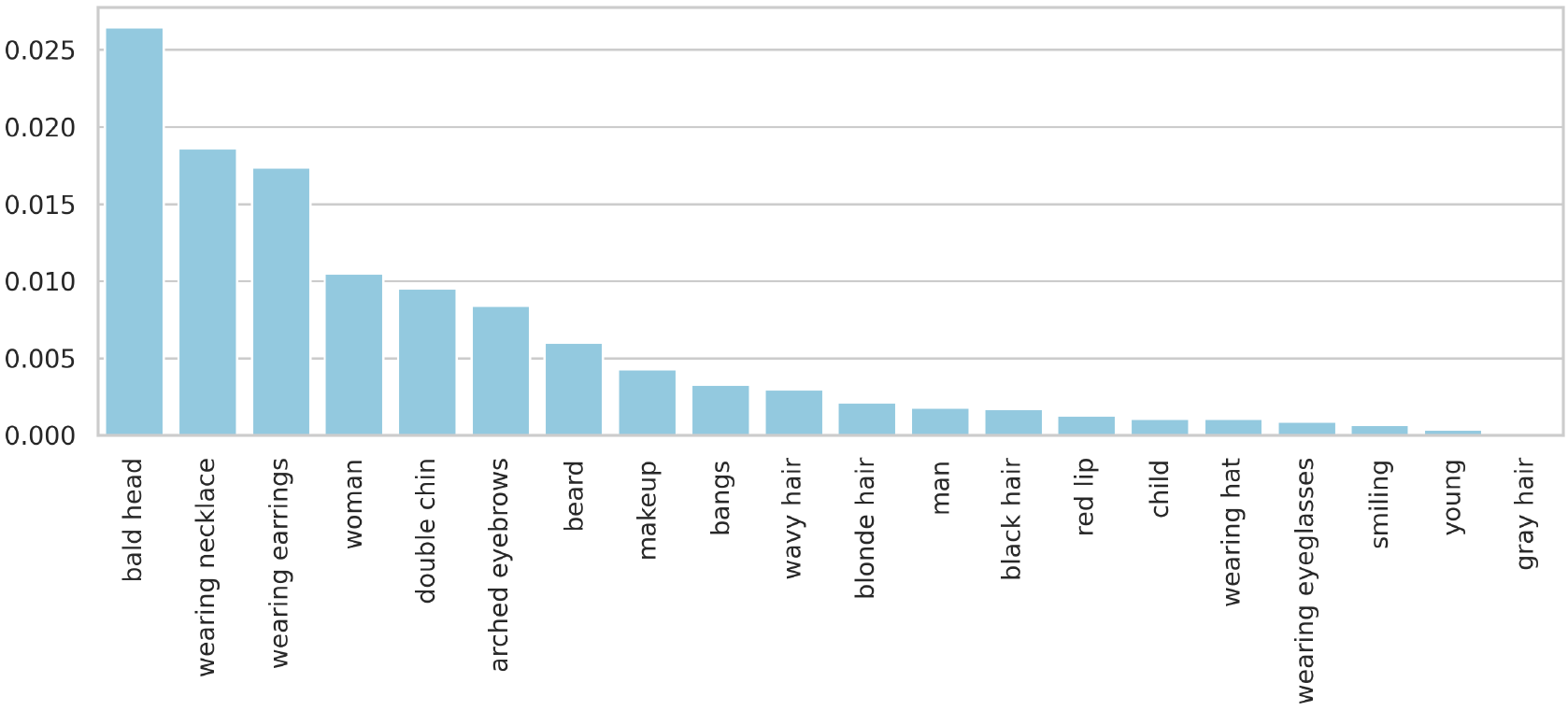}
    \end{subfigure}
    \hfill
    \begin{subfigure}[t]{0.8\linewidth}
        \centering
        \includegraphics[width=\linewidth]{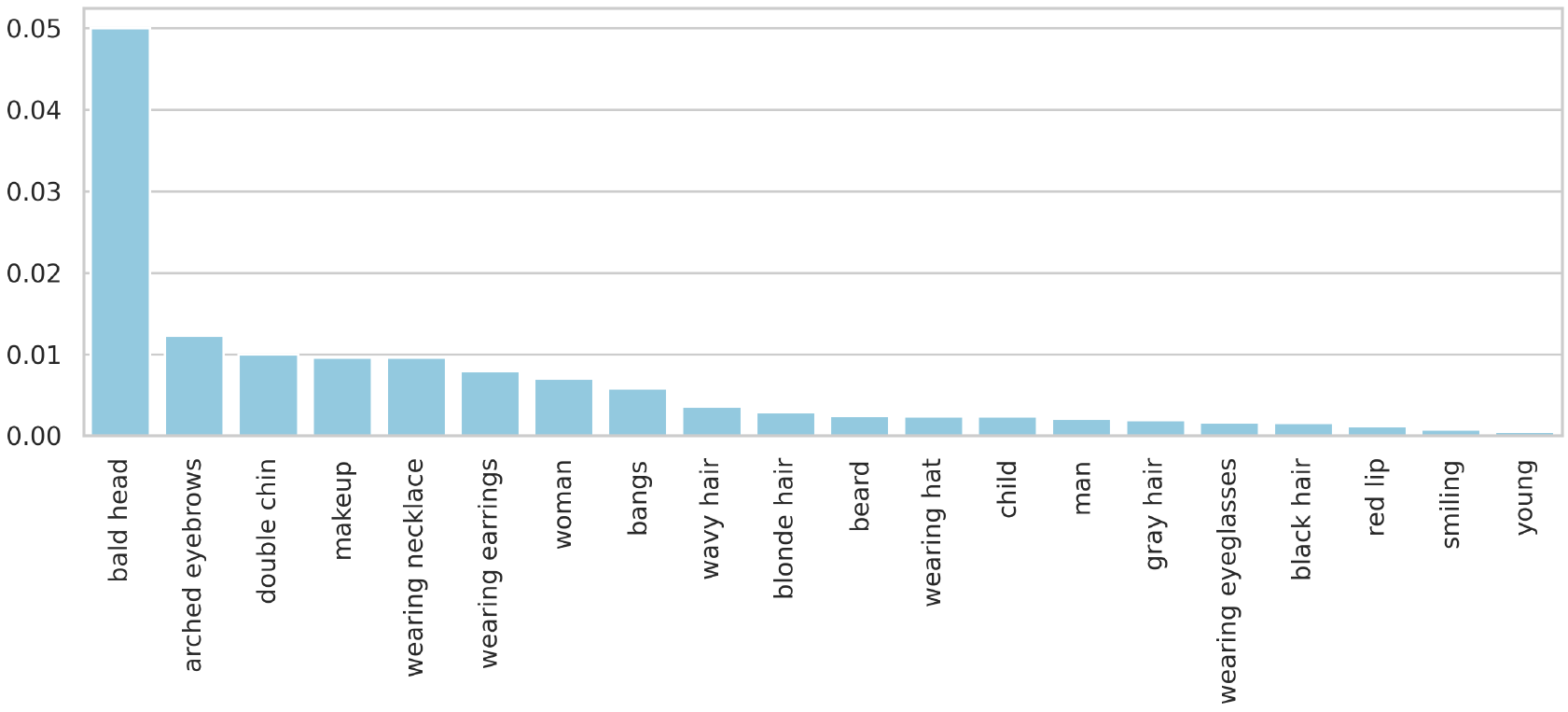}
    \end{subfigure}

    \caption{
        \small Visualization of the top attributes with the largest SaD scores. Top: LDM~\cite{Rombach2022ldm} with $50$ denoising steps; Bottom: LDM~\cite{Rombach2022ldm} with $200$ steps. \textit{Bald head} consistently ranks as the top attribute, with larger denoising steps further increasing the divergence. Additionally, \textit{wearing necklace} and \textit{wearing earrings} appear among the top attributes in both settings.
    }
    \label{fig:sup-mat_generative_eval}
\end{figure*}

In Sec.\ref{subsec:evaluation_sota}, we evaluate several generative models on the FFHQ dataset and report results in Tab.\ref{tab:sota_evaluation}. Diffusion models~\cite{Rombach2022ldm} perform worse than StyleGAN variants~\cite{karras2019style,karras2020analyzing} in the face domain, and increasing the number of denoising steps further widens this gap. To illustrate the attribute-level differences from real FFHQ data, we present bar charts in Figure \ref{fig:sup-mat_generative_eval}. The attribute \textit{bald head} contributes most to the observed divergence, with larger denoising steps amplifying this effect. Meanwhile, diffusion models tend to generate more fine details, such as \textit{earrings} and \textit{necklaces}.

\subsubsection{ImageNet Evaluation}
\label{subsec:sup-mat_imagenet_eval}

We further evaluate SaD computed with RA-CLIPScore on ImageNet and compare its results with conventional generative-model evaluation metrics, including FID, KID, Precision, Recall, Density, and Coverage. As shown in Tab.~\ref{tab:sup-mat_imagenet_sad_eval}, StyleGAN-XL~\cite{sauer2022stylegan} achieves the lowest SaD, consistent with its leading performance in FID, KID, and Recall. This agreement suggests that SaD captures aspects of distributional fidelity reflected by conventional metrics while providing complementary attribute-level information. Notably, the large standard deviations of SaD across ImageNet classes reveal substantial class-wise variation in attribute-level discrepancies. Such variations are not directly exposed by aggregate metrics such as FID and KID, highlighting the potential of RA-CLIPScore for more fine-grained, class-wise evaluation.

\begin{table*}[t]
\centering
\small
\setlength{\tabcolsep}{3pt}
\renewcommand{\arraystretch}{1.1}

\resizebox{\textwidth}{!}{%
\begin{tabular}{@{}lccccccc@{}}
\toprule
\textbf{Model}
& \textbf{SaD} $\downarrow$
& \textbf{FID} $\downarrow$
& \shortstack{\textbf{KID} $\downarrow$\\$(\times 10^{-3})$}
& \textbf{Prec.} $\uparrow$
& \textbf{Recall} $\uparrow$
& \textbf{Density} $\uparrow$
& \textbf{Coverage} $\uparrow$ \\
\midrule
StyleGAN-XL~\cite{sauer2022stylegan}
& $\mathbf{0.77 \pm 8.67}$
& \textbf{2.91}
& \textbf{0.25}
& 0.89
& \textbf{0.76}
& 1.21
& 0.97 \\
LDM~\cite{Rombach2022ldm}
& $362.92 \pm 1114.33$
& 4.23
& 0.99
& \textbf{0.93}
& 0.70
& 1.39
& \textbf{0.98} \\
BigGAN~\cite{brock2018large}
& $430.07 \pm 1339.78$
& 7.94
& 1.31
& 0.92
& 0.58
& \textbf{1.48}
& 0.94 \\
\bottomrule
\end{tabular}%
}

\caption{Comparison of SaD and conventional generative-model evaluation metrics on ImageNet. SaD is reported as mean $\pm$ standard deviation across classes, and KID values are scaled by $10^3$. Arrows indicate the preferred direction for each metric, and best results are highlighted in bold.}
\label{tab:sup-mat_imagenet_sad_eval}
\end{table*}

\subsection{Spatial Evaluation}
\label{subsec:sup-mat_spatial_eval}

We extend the evaluation to the spatial domain without additional computational cost using Eq.\eqref{eq:regional_sad}, and apply it to evaluate state-of-the-art generative models in Sec.\ref{subsec:spatial_eval}. We rank class-wise divergences using mean absolute R-SaD in Tab.\ref{tab:spatial_eval}, which is defined as:

\begin{align}
\argmax_{t_i}\frac{1}{|\mathcal{J}|}\sum_{j\in \mathcal{J}}\big|\text{R-SaD}_j(t_i)\big|, \label{eq:sup-mat_mean_r_sad}
\end{align}
where $\text{R-SaD}_j(t_i)$ is computed from Eq.\eqref{eq:regional_sad}, $t_i$ denotes the $i$-th ImageNet class, and $j$ indexes local patches. For each class $t_i$, mean R-SaD measures the average magnitude of regional bias over all patches. We sort all classes by this metric and report the top three most biased classes per model.

\subsubsection{More Examples}
\label{subsubsec:sup-mat_spatial_examples}

\begin{figure*}[t]
    \centering
    \newlength{\colH}
    \setlength{\colH}{0.30\textwidth}  

    \begin{minipage}[c][\colH][t]{0.15\textwidth}
        \centering
        \includegraphics[width=\linewidth]{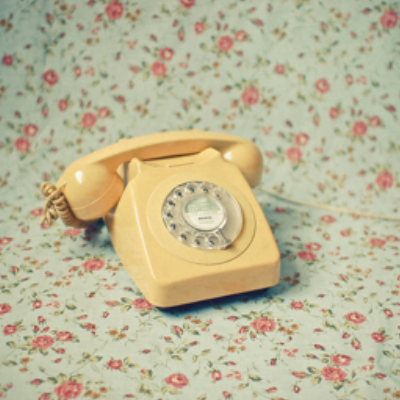}\\
        \includegraphics[width=\linewidth]{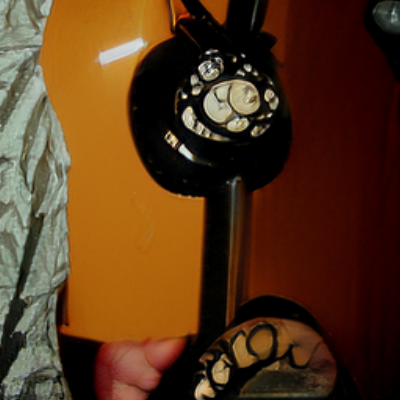}
    \end{minipage}
    \begin{minipage}[c][\colH][t]{0.15\textwidth}
        \centering
        \includegraphics[width=\linewidth]{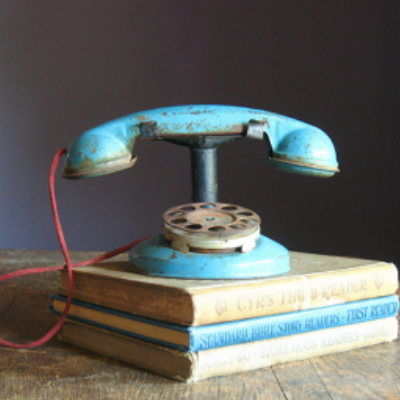}\\
        \includegraphics[width=\linewidth]{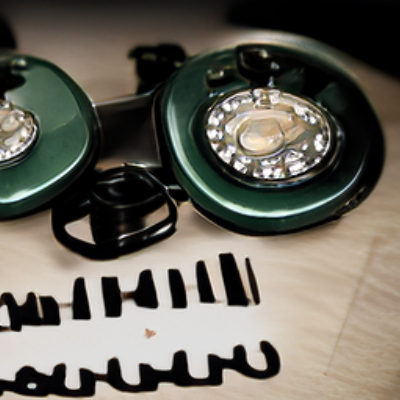}
    \end{minipage}
    \begin{minipage}[c][\colH][t]{0.15\textwidth}
        \centering
        \includegraphics[width=\linewidth]{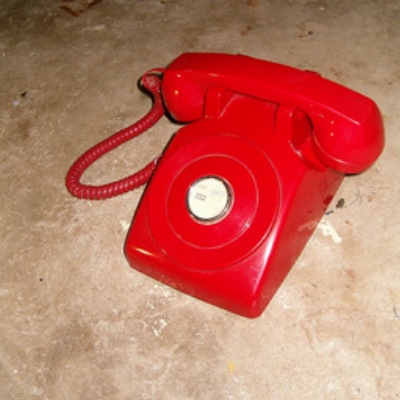}\\
        \includegraphics[width=\linewidth]{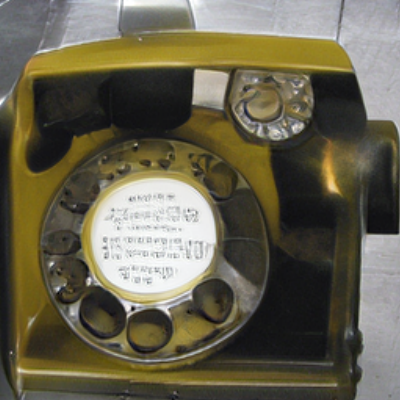}
    \end{minipage}
    \begin{minipage}[c][\colH][t]{0.15\textwidth}
        \centering
        \includegraphics[width=\linewidth]{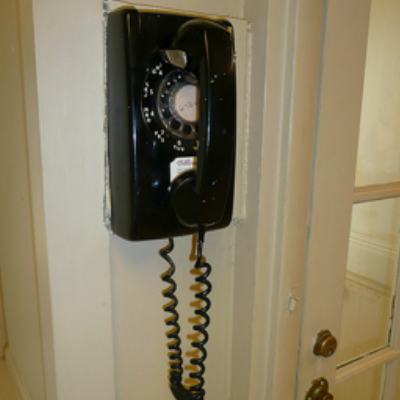}\\
        \includegraphics[width=\linewidth]{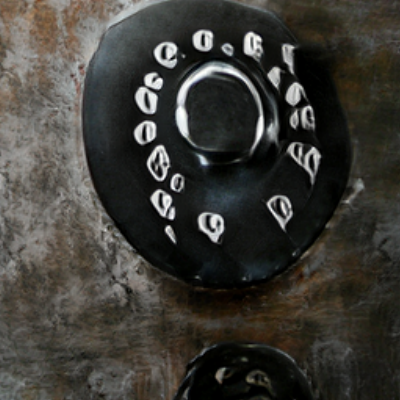}
    \end{minipage}
    \begin{minipage}[c][\colH][c]{0.15\textwidth}
        \centering
        \includegraphics[width=\linewidth]{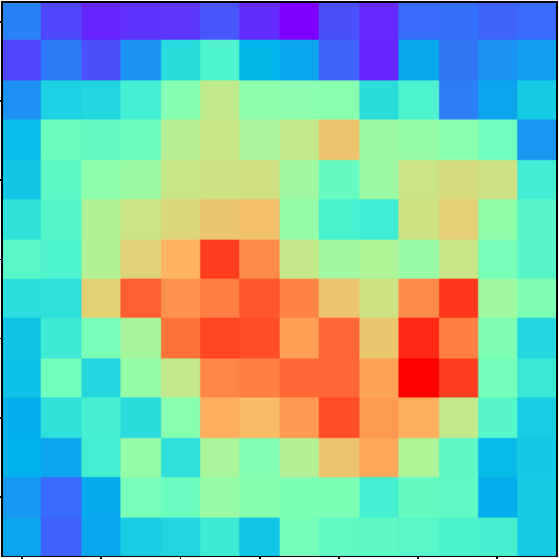}
    \end{minipage}
    
    \caption{\small\textit{Rotary dial telephone}. The first row shows samples from ImageNet, and the second row shows images generated by StyleGAN~\cite{sauer2022stylegan}.}

    \label{fig:sup-mat_spatial_eval_0}
\end{figure*}

\begin{figure*}[t]
    \centering
    \setlength{\colH}{0.30\textwidth}  

    \begin{minipage}[c][\colH][t]{0.15\textwidth}
        \centering
        \includegraphics[width=\linewidth]{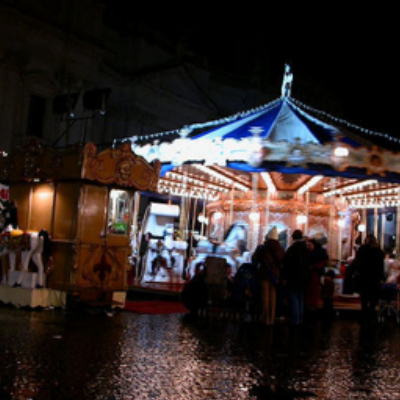}\\
        \includegraphics[width=\linewidth]{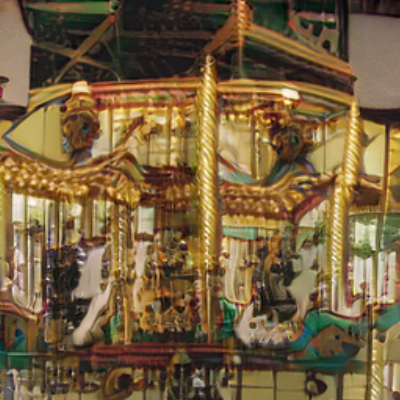}
    \end{minipage}
    \begin{minipage}[c][\colH][t]{0.15\textwidth}
        \centering
        \includegraphics[width=\linewidth]{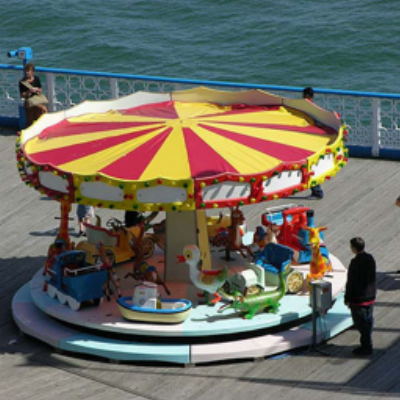}\\
        \includegraphics[width=\linewidth]{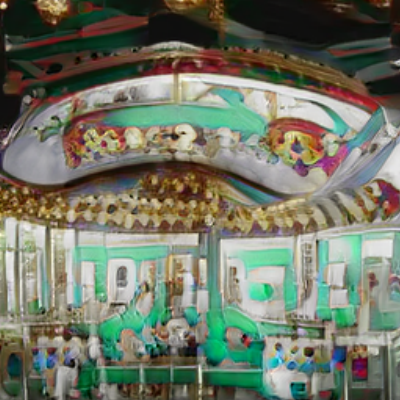}
    \end{minipage}
    \begin{minipage}[c][\colH][t]{0.15\textwidth}
        \centering
        \includegraphics[width=\linewidth]{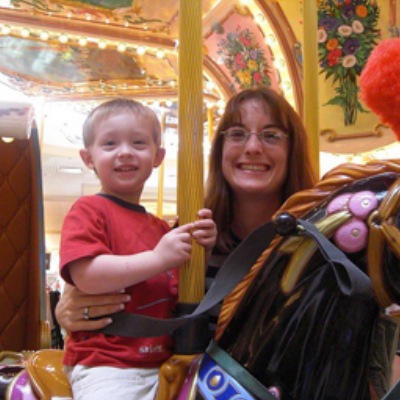}\\
        \includegraphics[width=\linewidth]{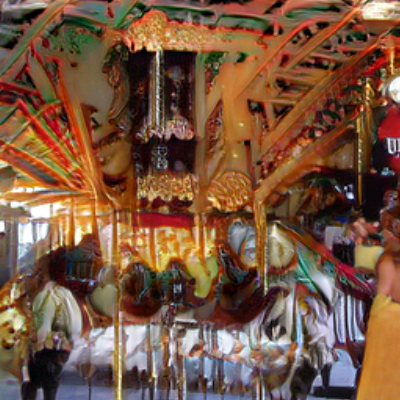}
    \end{minipage}
    \begin{minipage}[c][\colH][t]{0.15\textwidth}
        \centering
        \includegraphics[width=\linewidth]{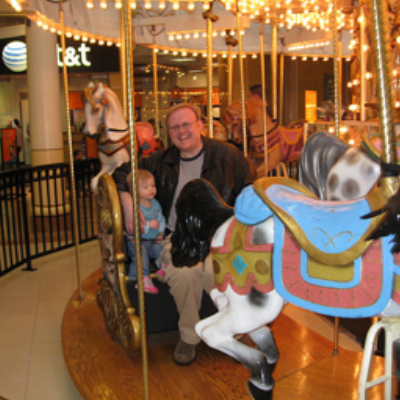}\\
        \includegraphics[width=\linewidth]{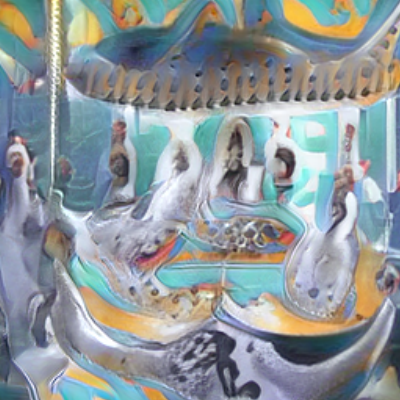}
    \end{minipage}
    \begin{minipage}[c][\colH][c]{0.15\textwidth}
        \centering
        \includegraphics[width=\linewidth]{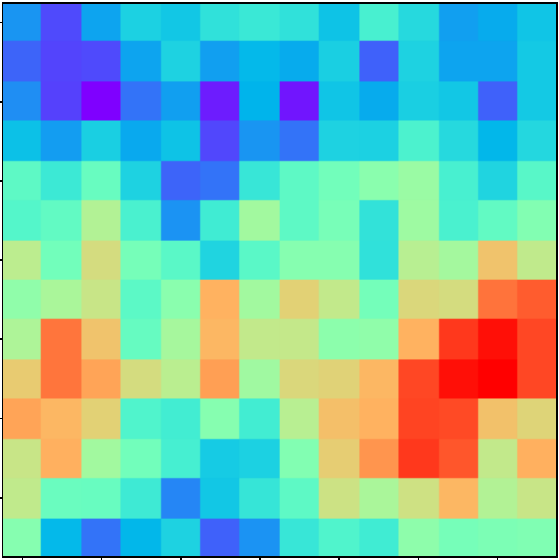}
    \end{minipage}

    \caption{\small\textit{Carousel}. The first row shows samples from ImageNet, and the second row shows images generated by StyleGAN~\cite{sauer2022stylegan}.}

    \label{fig:sup-mat_spatial_eval}
\end{figure*}

\begin{figure*}[t]
    \centering
    \setlength{\colH}{0.30\textwidth}  

    \begin{minipage}[c][\colH][t]{0.15\textwidth}
        \centering
        \includegraphics[width=\linewidth]{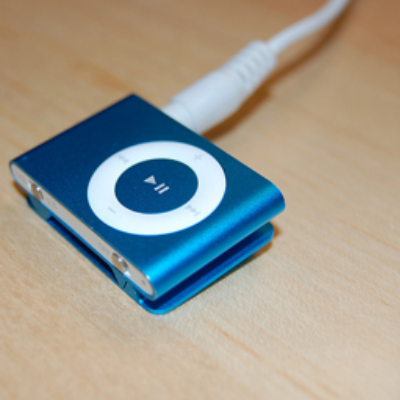}\\
        \includegraphics[width=\linewidth]{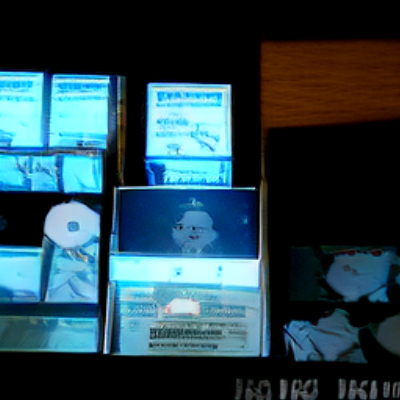}
    \end{minipage}
    \begin{minipage}[c][\colH][t]{0.15\textwidth}
        \centering
        \includegraphics[width=\linewidth]{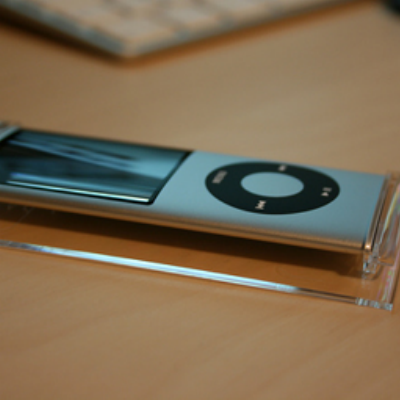}\\
        \includegraphics[width=\linewidth]{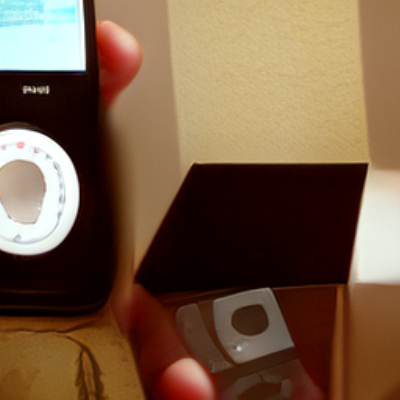}
    \end{minipage}
    \begin{minipage}[c][\colH][t]{0.15\textwidth}
        \centering
        \includegraphics[width=\linewidth]{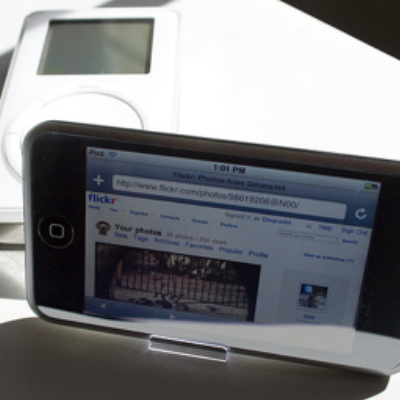}\\
        \includegraphics[width=\linewidth]{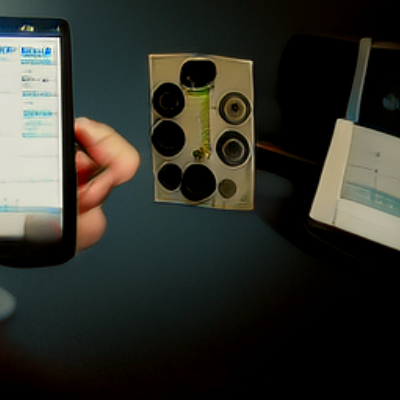}
    \end{minipage}
    \begin{minipage}[c][\colH][t]{0.15\textwidth}
        \centering
        \includegraphics[width=\linewidth]{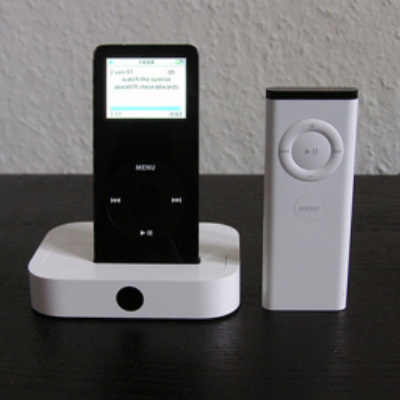}\\
        \includegraphics[width=\linewidth]{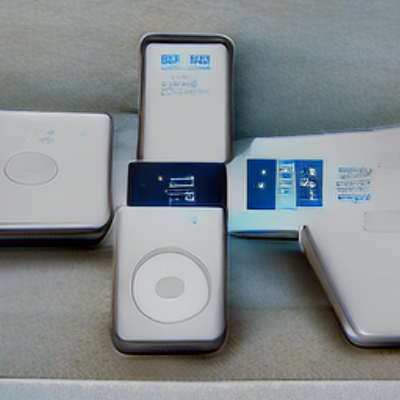}
    \end{minipage}
    \begin{minipage}[c][\colH][c]{0.15\textwidth}
        \centering
        \includegraphics[width=\linewidth]{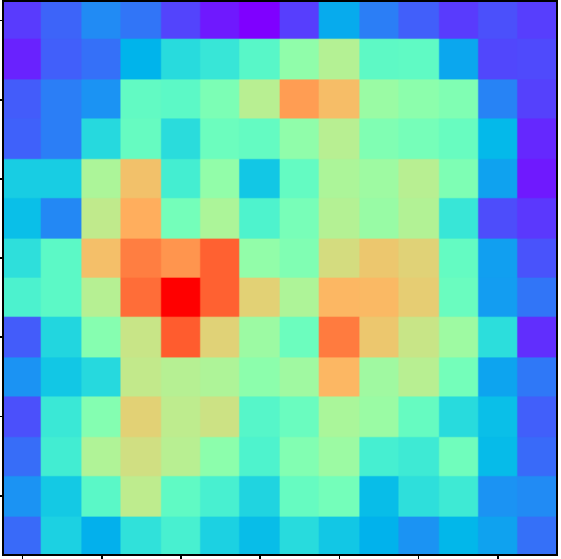}
    \end{minipage}

    \caption{\small\textit{iPod}. The first row shows samples from ImageNet, and the second row shows images generated by StyleGAN~\cite{sauer2022stylegan}.}

    \label{fig:sup-mat_spatial_eval_1}
\end{figure*}

\begin{figure*}[t]
    \centering
    \setlength{\colH}{0.30\textwidth}  

    \begin{minipage}[c][\colH][t]{0.15\textwidth}
        \centering
        \includegraphics[width=\linewidth]{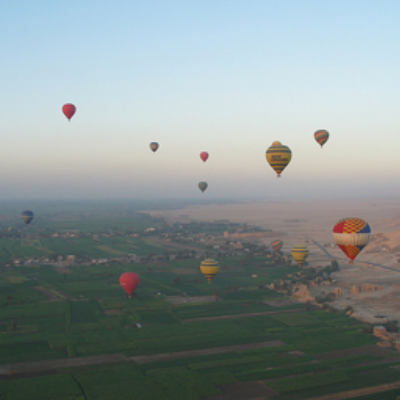}\\
        \includegraphics[width=\linewidth]{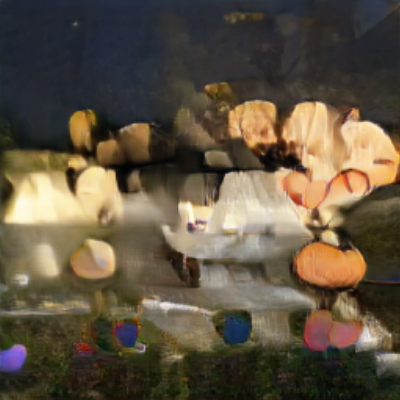}
    \end{minipage}
    \begin{minipage}[c][\colH][t]{0.15\textwidth}
        \centering
        \includegraphics[width=\linewidth]{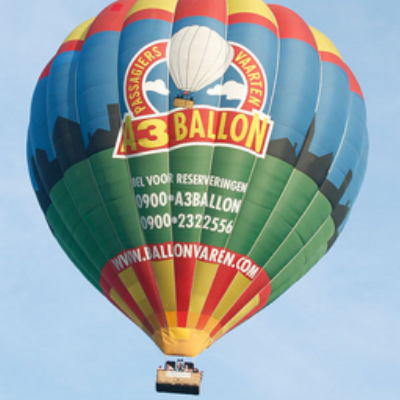}\\
        \includegraphics[width=\linewidth]{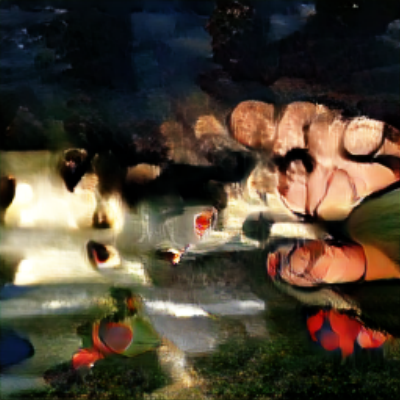}
    \end{minipage}
    \begin{minipage}[c][\colH][t]{0.15\textwidth}
        \centering
        \includegraphics[width=\linewidth]{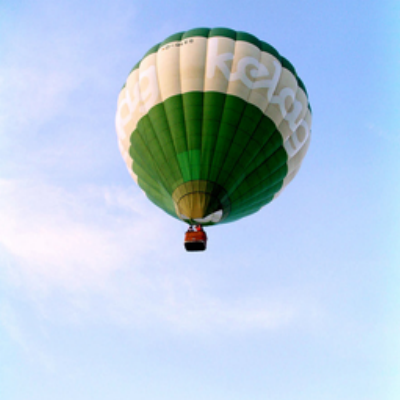}\\
        \includegraphics[width=\linewidth]{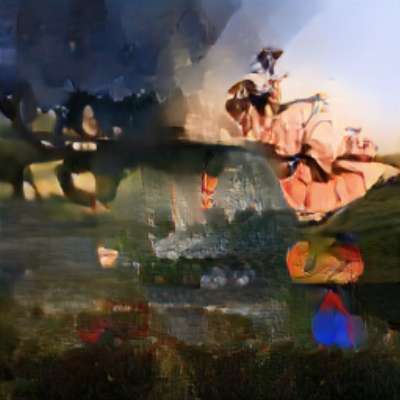}
    \end{minipage}
    \begin{minipage}[c][\colH][t]{0.15\textwidth}
        \centering
        \includegraphics[width=\linewidth]{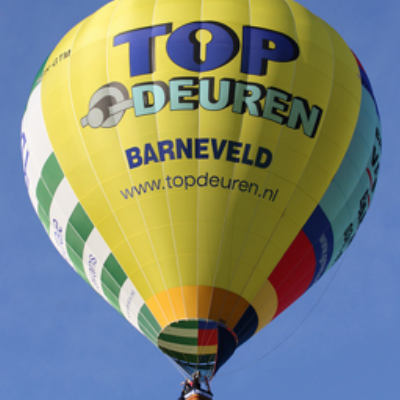}\\
        \includegraphics[width=\linewidth]{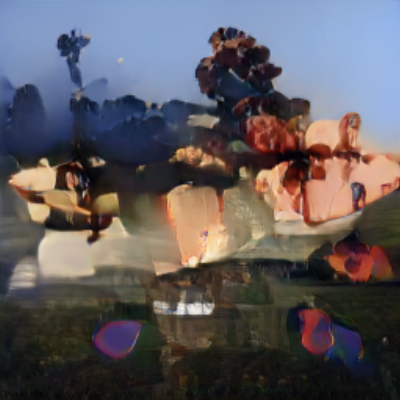}
    \end{minipage}
    \begin{minipage}[c][\colH][c]{0.15\textwidth}
        \centering
        \includegraphics[width=\linewidth]{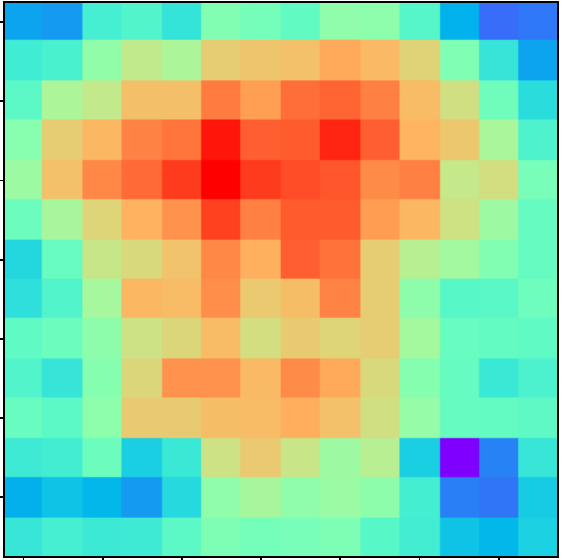}
    \end{minipage}

    \caption{\small\textit{Balloon}. The first row shows samples from ImageNet, and the second row shows images generated by BigGAN~\cite{brock2018large}.}
    \label{fig:sup-mat_spatial_eval_3}
\end{figure*}

\begin{figure*}[t]
    \centering
    \setlength{\colH}{0.30\textwidth}  

    \begin{minipage}[c][\colH][t]{0.15\textwidth}
        \centering
        \includegraphics[width=\linewidth]{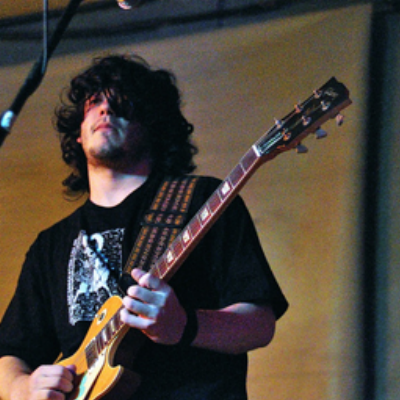}\\
        \includegraphics[width=\linewidth]{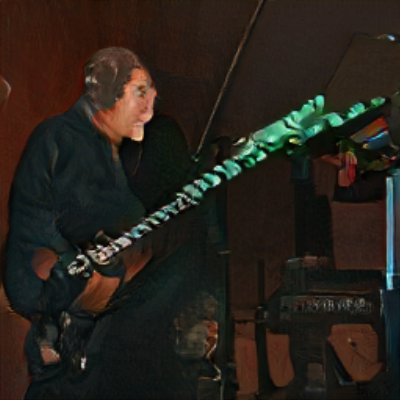}
    \end{minipage}
    \begin{minipage}[c][\colH][t]{0.15\textwidth}
        \centering
        \includegraphics[width=\linewidth]{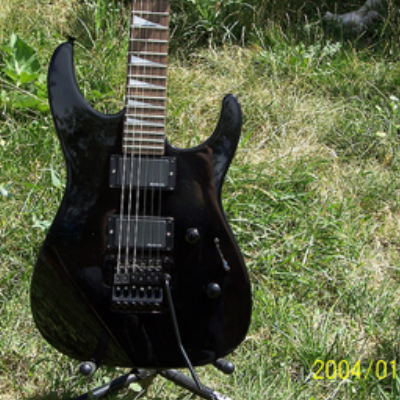}\\
        \includegraphics[width=\linewidth]{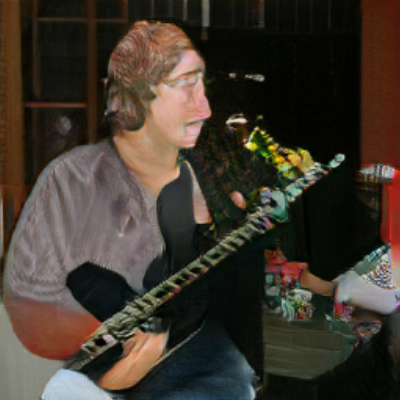}
    \end{minipage}
    \begin{minipage}[c][\colH][t]{0.15\textwidth}
        \centering
        \includegraphics[width=\linewidth]{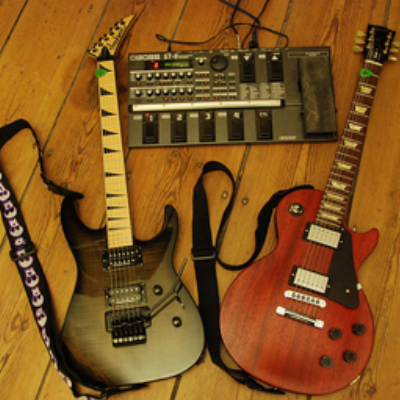}\\
        \includegraphics[width=\linewidth]{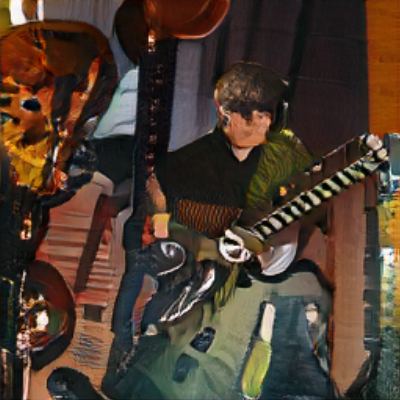}
    \end{minipage}
    \begin{minipage}[c][\colH][t]{0.15\textwidth}
        \centering
        \includegraphics[width=\linewidth]{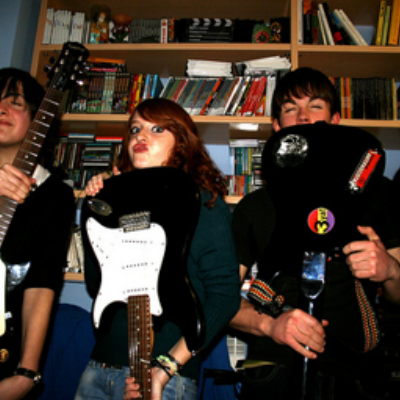}\\
        \includegraphics[width=\linewidth]{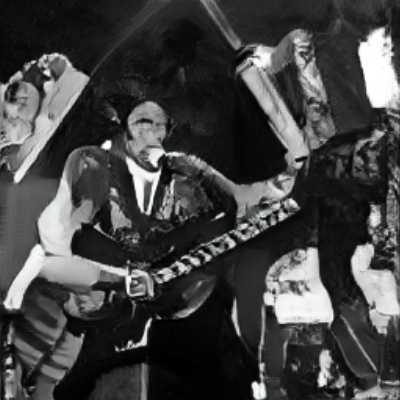}
    \end{minipage}
    \begin{minipage}[c][\colH][c]{0.15\textwidth}
        \centering
        \includegraphics[width=\linewidth]{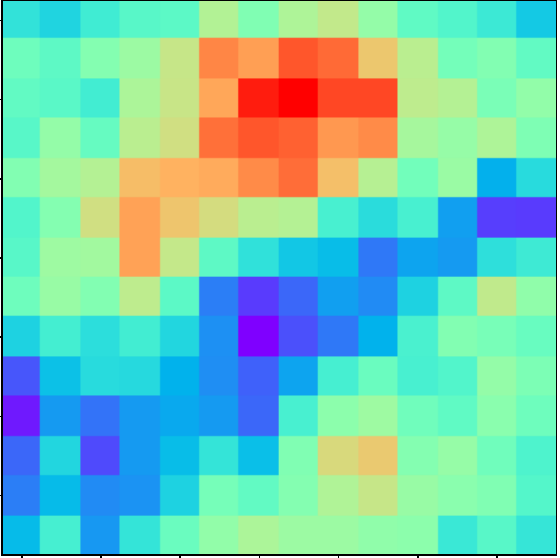}
    \end{minipage}

    \caption{\small\textit{Electric guitar}. The first row shows samples from ImageNet, and the second row shows images generated by BigGAN~\cite{brock2018large}.}
    \label{fig:sup-mat_spatial_eval_4}
\end{figure*}

\begin{figure*}[t]
    \centering
    \setlength{\colH}{0.30\textwidth}  

    \begin{minipage}[c][\colH][t]{0.15\textwidth}
        \centering
        \includegraphics[width=\linewidth]{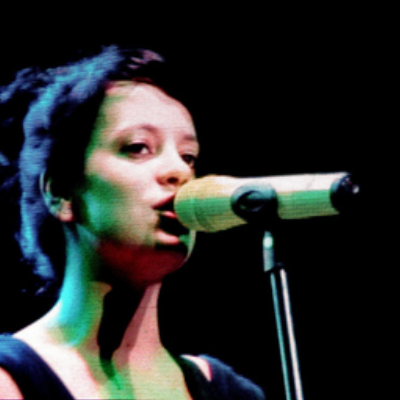}\\
        \includegraphics[width=\linewidth]{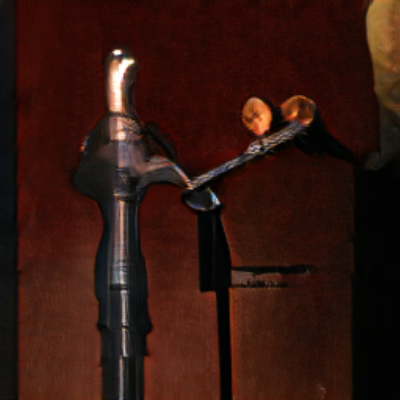}
    \end{minipage}
    \begin{minipage}[c][\colH][t]{0.15\textwidth}
        \centering
        \includegraphics[width=\linewidth]{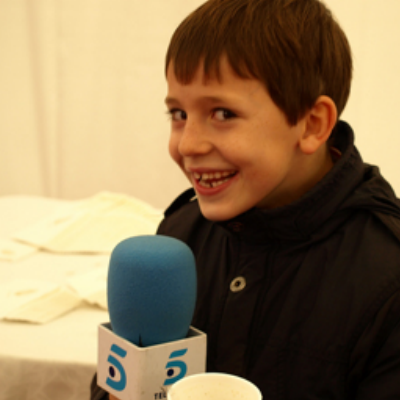}\\
        \includegraphics[width=\linewidth]{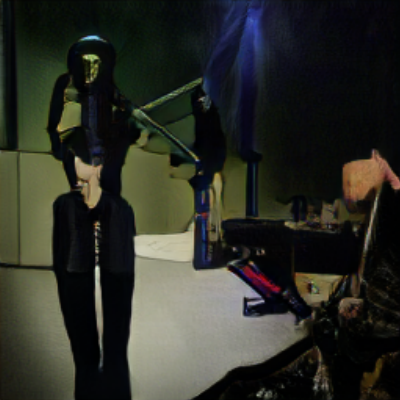}
    \end{minipage}
    \begin{minipage}[c][\colH][t]{0.15\textwidth}
        \centering
        \includegraphics[width=\linewidth]{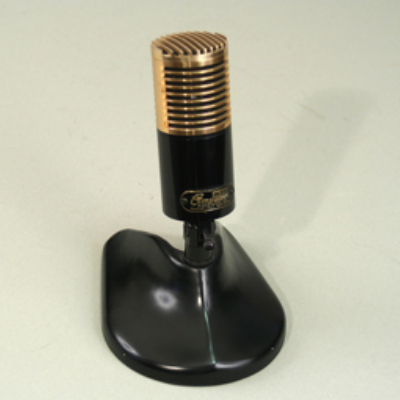}\\
        \includegraphics[width=\linewidth]{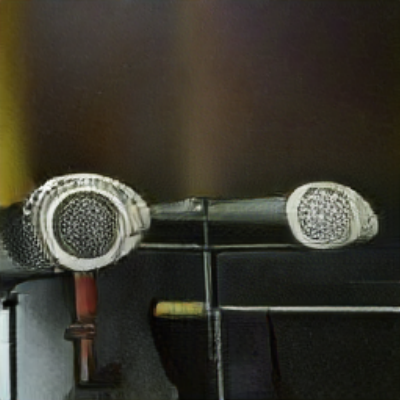}
    \end{minipage}
    \begin{minipage}[c][\colH][t]{0.15\textwidth}
        \centering
        \includegraphics[width=\linewidth]{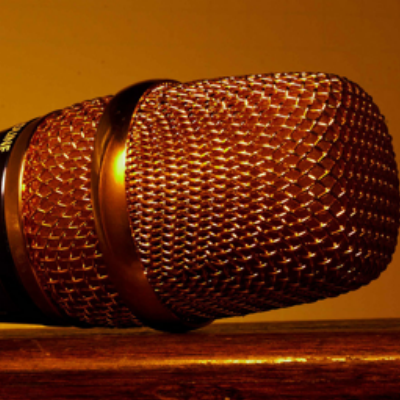}\\
        \includegraphics[width=\linewidth]{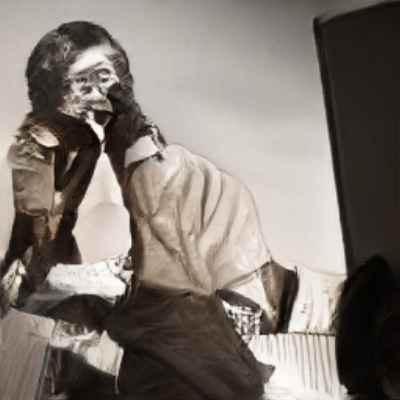}
    \end{minipage}
    \begin{minipage}[c][\colH][c]{0.15\textwidth}
        \centering
        \includegraphics[width=\linewidth]{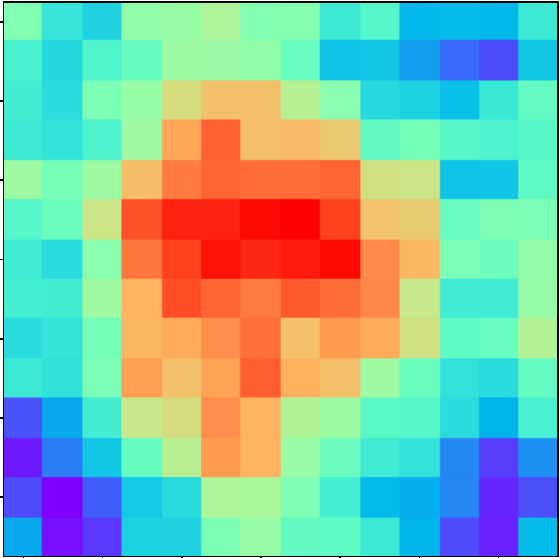}
    \end{minipage}

    \caption{\small\textit{Microphone}. The first row shows samples from ImageNet, and the second row shows images generated by BigGAN~\cite{brock2018large}.}
    \label{fig:sup-mat_spatial_eval_5}
\end{figure*}

\begin{figure*}[t]
    \centering
    \setlength{\colH}{0.30\textwidth}  

    \begin{minipage}[c][\colH][t]{0.15\textwidth}
        \centering
        \includegraphics[width=\linewidth]{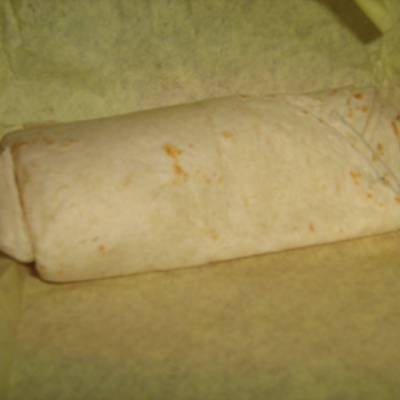}\\
        \includegraphics[width=\linewidth]{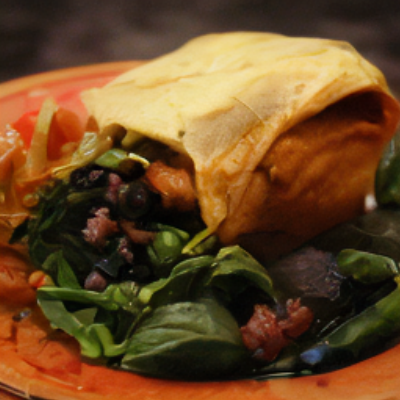}
    \end{minipage}
    \begin{minipage}[c][\colH][t]{0.15\textwidth}
        \centering
        \includegraphics[width=\linewidth]{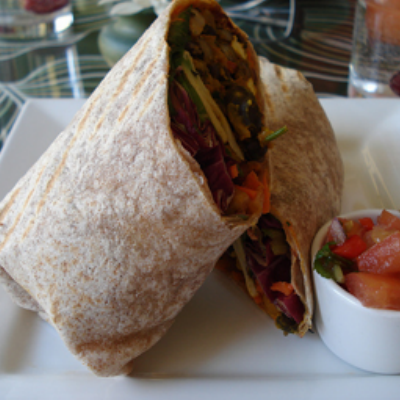}\\
        \includegraphics[width=\linewidth]{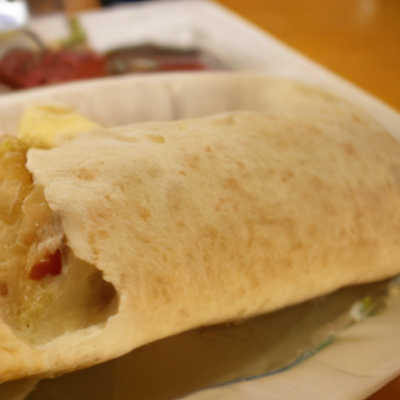}
    \end{minipage}
    \begin{minipage}[c][\colH][t]{0.15\textwidth}
        \centering
        \includegraphics[width=\linewidth]{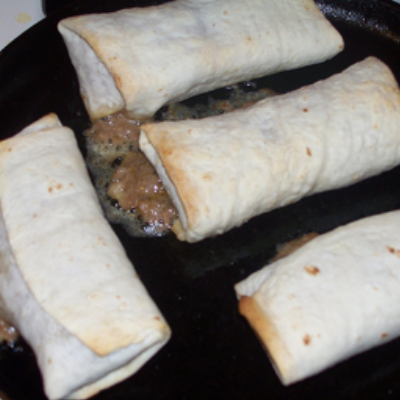}\\
        \includegraphics[width=\linewidth]{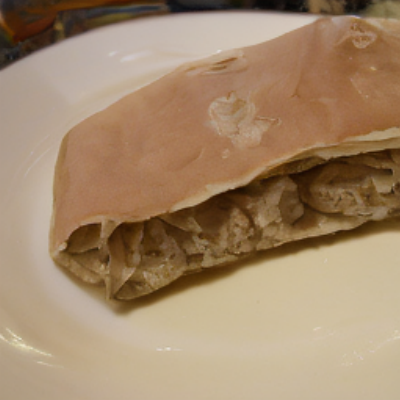}
    \end{minipage}
    \begin{minipage}[c][\colH][t]{0.15\textwidth}
        \centering
        \includegraphics[width=\linewidth]{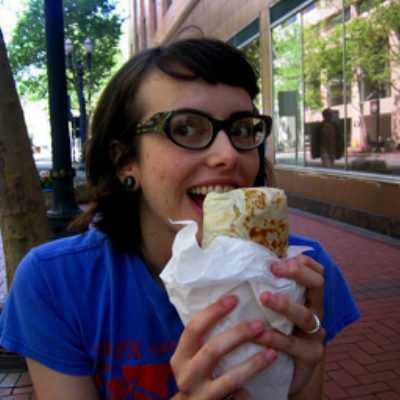}\\
        \includegraphics[width=\linewidth]{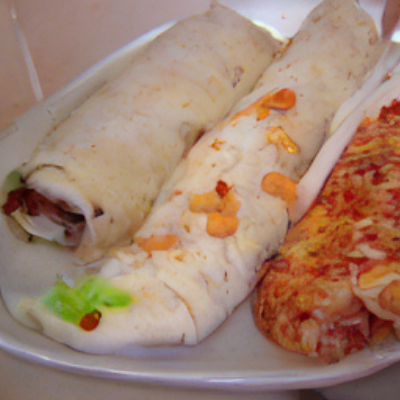}
    \end{minipage}
    \begin{minipage}[c][\colH][c]{0.15\textwidth}
        \centering
        \includegraphics[width=\linewidth]{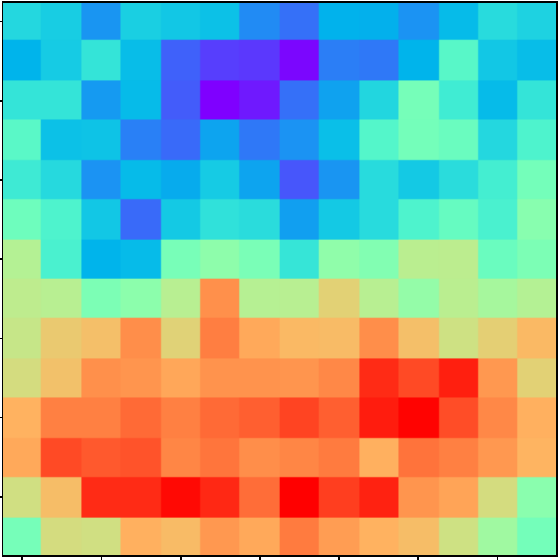}
    \end{minipage}

    \caption{\small\textit{Burrito}. The first row shows samples from ImageNet, and the second row shows images generated by LDM(250)~\cite{Rombach2022ldm}.}
    \label{fig:sup-mat_spatial_eval_7}
\end{figure*}

\begin{figure*}[t]
    \centering
    \setlength{\colH}{0.30\textwidth}  

    \begin{minipage}[c][\colH][t]{0.15\textwidth}
        \centering
        \includegraphics[width=\linewidth]{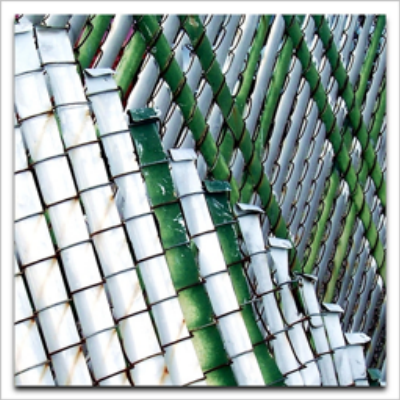}\\
        \includegraphics[width=\linewidth]{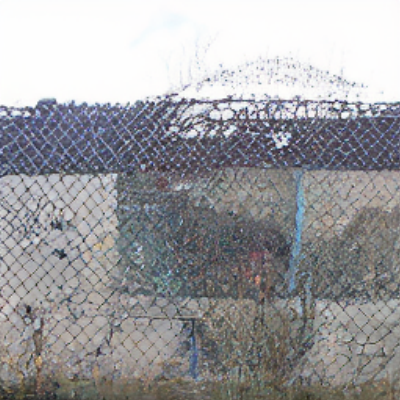}
    \end{minipage}
    \begin{minipage}[c][\colH][t]{0.15\textwidth}
        \centering
        \includegraphics[width=\linewidth]{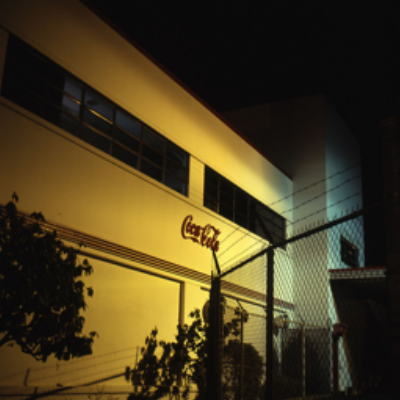}\\
        \includegraphics[width=\linewidth]{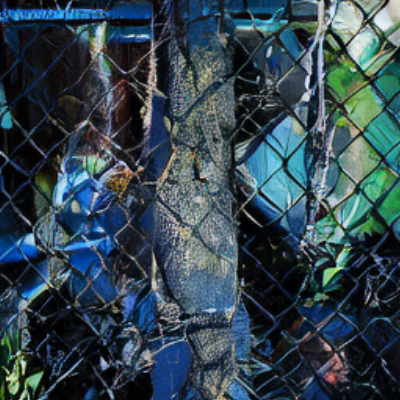}
    \end{minipage}
    \begin{minipage}[c][\colH][t]{0.15\textwidth}
        \centering
        \includegraphics[width=\linewidth]{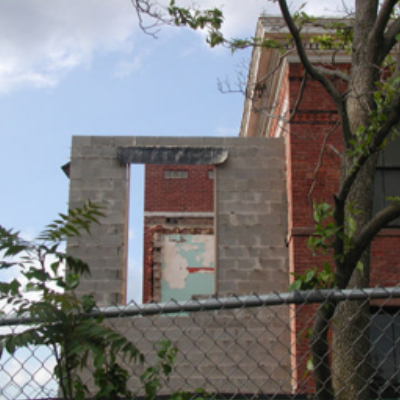}\\
        \includegraphics[width=\linewidth]{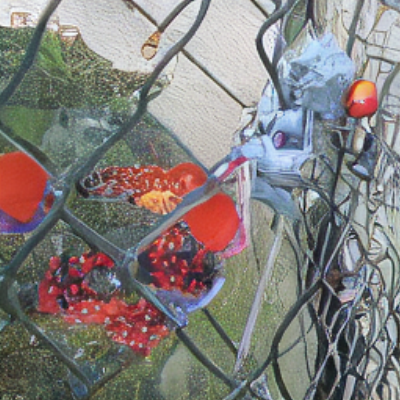}
    \end{minipage}
    \begin{minipage}[c][\colH][t]{0.15\textwidth}
        \centering
        \includegraphics[width=\linewidth]{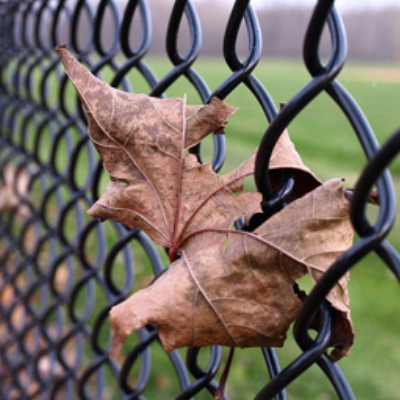}\\
        \includegraphics[width=\linewidth]{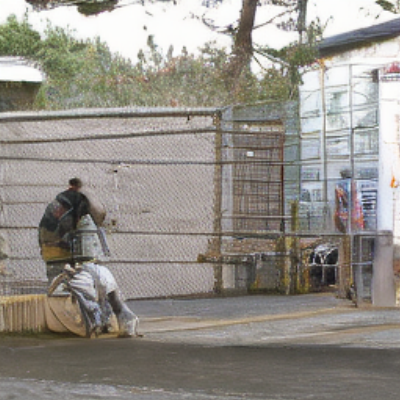}
    \end{minipage}
    \begin{minipage}[c][\colH][c]{0.15\textwidth}
        \centering
        \includegraphics[width=\linewidth]{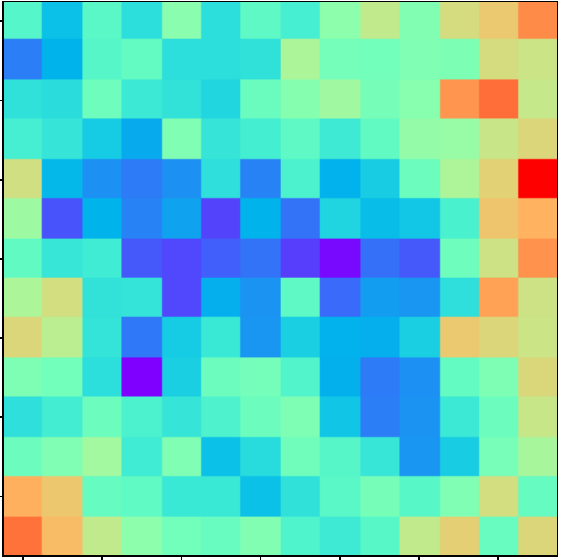}
    \end{minipage}

    \caption{\small\textit{Chain-link fence}. The first row shows samples from ImageNet, and the second row shows images generated by LDM(250)~\cite{Rombach2022ldm}.}
    \label{fig:sup-mat_spatial_eval_8}
\end{figure*}

\begin{figure*}[t]
    \centering
    \setlength{\colH}{0.30\textwidth}  

    \begin{minipage}[c][\colH][t]{0.15\textwidth}
        \centering
        \includegraphics[width=\linewidth]{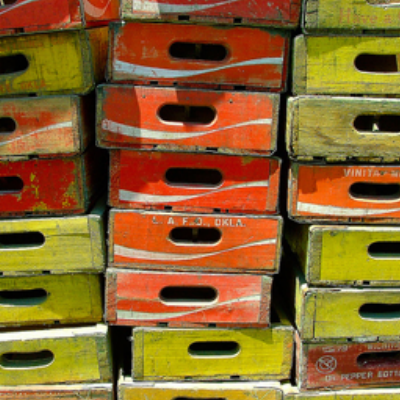}\\
        \includegraphics[width=\linewidth]{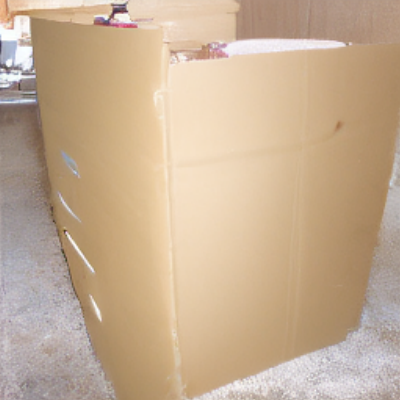}
    \end{minipage}
    \begin{minipage}[c][\colH][t]{0.15\textwidth}
        \centering
        \includegraphics[width=\linewidth]{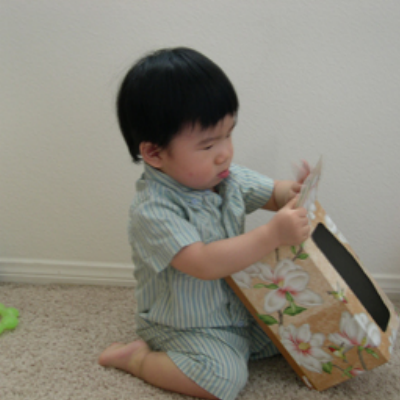}\\
        \includegraphics[width=\linewidth]{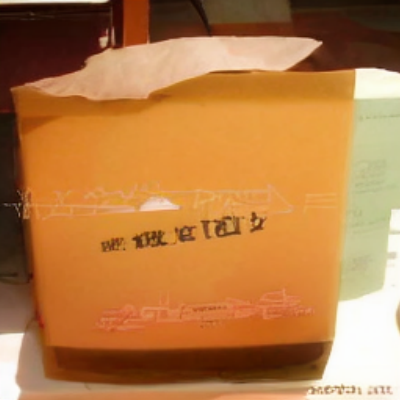}
    \end{minipage}
    \begin{minipage}[c][\colH][t]{0.15\textwidth}
        \centering
        \includegraphics[width=\linewidth]{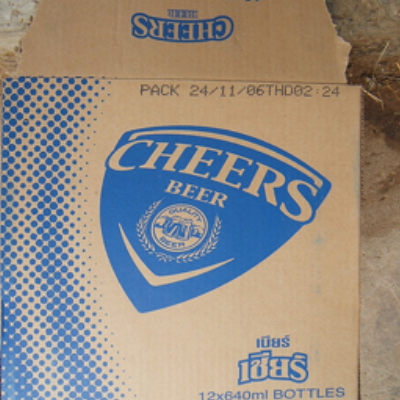}\\
        \includegraphics[width=\linewidth]{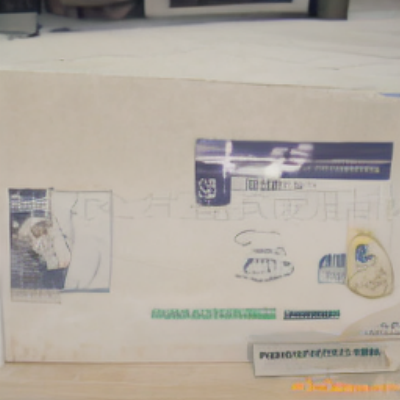}
    \end{minipage}
    \begin{minipage}[c][\colH][t]{0.15\textwidth}
        \centering
        \includegraphics[width=\linewidth]{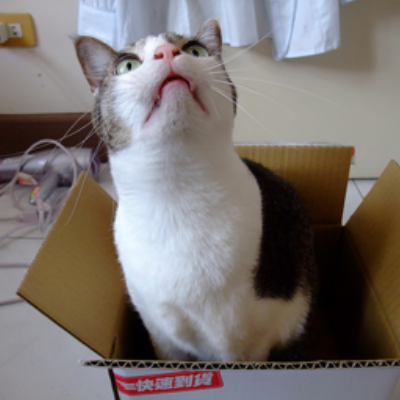}\\
        \includegraphics[width=\linewidth]{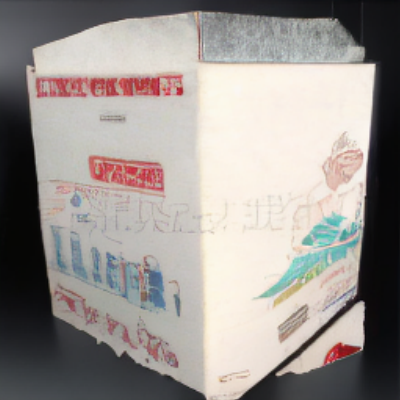}
    \end{minipage}
    \begin{minipage}[c][\colH][c]{0.15\textwidth}
        \centering
        \includegraphics[width=\linewidth]{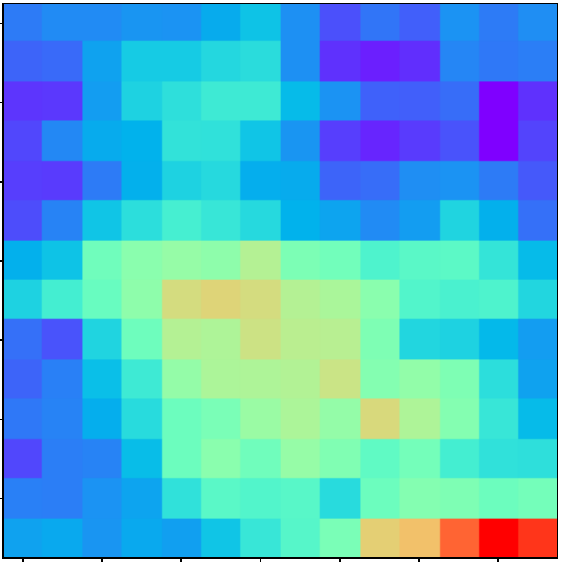}
    \end{minipage}

    \caption{\small\textit{Carton}. The first row shows samples from ImageNet, and the second row shows images generated by LDM(250)~\cite{Rombach2022ldm}.}
    \label{fig:sup-mat_spatial_eval_9}
\end{figure*}

We provide additional qualitative examples for different classes and generative models in Figs. \ref{fig:sup-mat_spatial_eval_0}–\ref{fig:sup-mat_spatial_eval_9}. In each set of figures, the first row shows real images from ImageNet~\cite{deng2009imagenet}, while the second row shows samples generated by the model specified in the figure caption. The image on the far right is the corresponding normalized R-SaD heatmap, where red indicates higher occurrence in training data and blue indicates higher occurrence in generated samples.

\subsection{User Study}
\label{subsec:sup-mat_userstudy}

\begin{figure}[t]
\centering

\begin{minipage}[t]{0.58\linewidth}
\vspace{0pt}
\centering
\includegraphics[width=\linewidth]{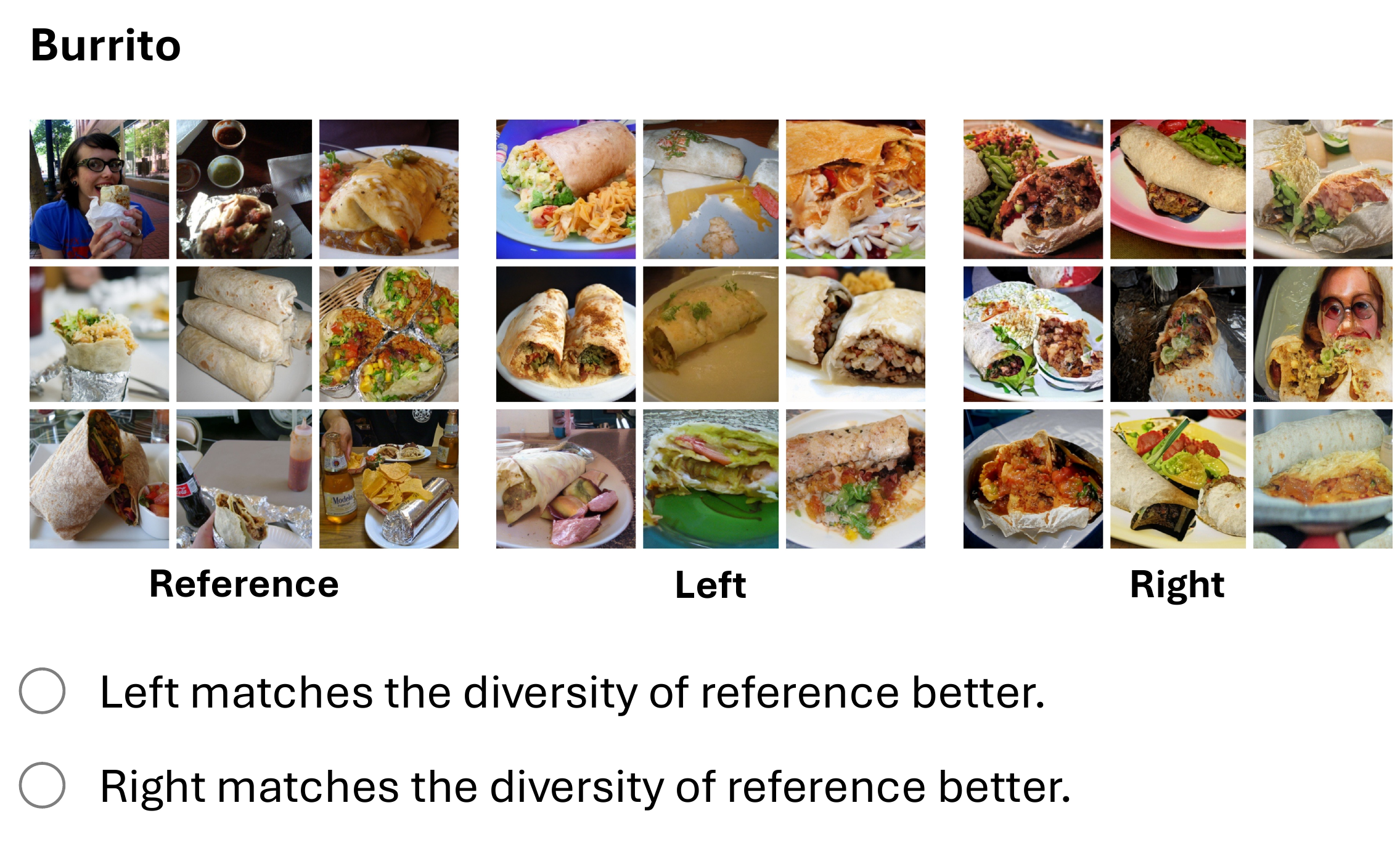}
\caption{\small Experimental setup for human user study}
\label{fig:expsetup}
\end{minipage}
\hfill
\begin{minipage}[t]{0.38\linewidth}
\vspace{0pt}
\centering
\setlength{\tabcolsep}{1.6pt}
\renewcommand{\arraystretch}{0.90}
\begin{tabular}{@{}lc@{}}
\toprule
\textbf{Metric} & \textbf{Correlation r} \\
\midrule
R-SaD     & 1.0 \\
Density~\cite{naeem2020reliable}   & 0.70 \\
RA-CLIPScore & 0.60 \\
KID~\cite{sutherland2018demystifying}       & 0.60 \\
LPIPS~\cite{zhang2018unreasonable}     & 0.53 \\
FID~\cite{heusel2017gans}       & 0.50 \\
CLIPScore~\cite{hessel2021clipscore} & 0.49 \\
Precision~\cite{kynkaanniemi2019improved,sajjadi2018assessing} & 0.39 \\
Recall~\cite{kynkaanniemi2019improved,sajjadi2018assessing}    & 0.29 \\
Coverage~\cite{naeem2020reliable}  & 0.29 \\
HCS~\cite{kim2023attribute} & 0.21 \\
\bottomrule
\end{tabular}
\captionof{table}{\small Pearson Correlation Coefficient between metric preference and human judgment.}
\label{tab:userstudy_results}
\end{minipage}
\end{figure}

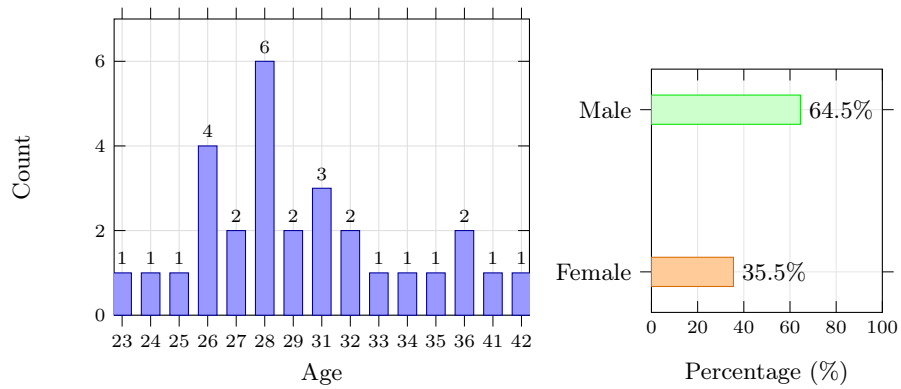
\begin{figure}[t]
\centering

\begin{minipage}[t]{0.58\linewidth}
\centering
\begin{tikzpicture}

\begin{axis}[
ybar,
width=\linewidth,
height=5.5cm,
bar width=7pt,
ymin=0,
ymax=7,
ylabel={Count},
xlabel={Age},
symbolic x coords={23,24,25,26,27,28,29,31,32,33,34,35,36,41,42},
xtick=data,
enlarge x limits=0.02,
grid=major,
grid style={gray!25},
axis line style={black},
tick style={black},
xticklabel style={font=\scriptsize},
yticklabel style={font=\scriptsize},
label style={font=\small},
nodes near coords,
every node near coord/.append style={
font=\scriptsize,
anchor=south,
},
]

\addplot[
fill=blue!40!white,
draw=blue!60!black
]
coordinates {
(23,1) (24,1) (25,1) (26,4)
(27,2) (28,6) (29,2) 
(31,3) (32,2) (33,1) (34,1)
(35,1) (36,2) (41,1) (42,1)
};
\end{axis}
\end{tikzpicture}
\end{minipage}
\begin{minipage}[t]{0.38\linewidth}
\centering
\begin{tikzpicture}
\begin{axis}[
    xbar,
    width=\linewidth,
    height=4.8cm,
    xmin=0,
    xmax=100,
    bar width=11pt,
    xlabel={Percentage (\%)},
    symbolic y coords={Male,Female},
    ytick=data,
    y dir=reverse,
    grid=major,
    grid style={gray!20},
    axis line style={black},
    tick style={black},
    xticklabel style={font=\scriptsize},
    yticklabel style={font=\small},
    label style={font=\small},
    enlarge y limits=0.25
]

\addplot[
    draw=green!90!black,
    fill=green!20!white,
    bar shift=0pt
] coordinates {(64.5,Male)(0,Female)};

\addplot[
    draw=orange!85!black,
    fill=orange!40,
    bar shift=0pt
] coordinates {(35.5,Female)};

\node[font=\small, anchor=west] at (axis cs:64.5,Male) {64.5\%};
\node[font=\small, anchor=west] at (axis cs:35.5,Female) {35.5\%};
\end{axis}
\end{tikzpicture}
\end{minipage}
\caption{\small Age and gender distributions of participants in the user study.}
\label{fig:sup-mat_age_distribution}
\end{figure}

We conduct a user study to assess the alignment with human perception. Images are generated using StyleGAN-XL~\cite{sauer2022stylegan}, BigGAN~\cite{brock2018large}, and LDM~\cite{Rombach2022ldm} trained on ImageNet. For each model, we select classes exhibiting clear spatial biases (Tab.~\ref{tab:spatial_eval}). A total of $29$ participants were asked to choose whether a set of $9$ images generated by model A or model B more closely resembled $9$ reference training images in terms of diversity (see Fig. \ref{fig:expsetup}). One of the models (A or B) was chosen to be the model with high R-SaD score for a given attribute, while the other was chosen from the remaining two models with lower R-SaD scores. The order was randomized. Each participant evaluated $20$ attributes. The $29$ participants were aged between $23$ and $42$, with an average age of $30$ (age distribution shown in Fig. \ref{fig:sup-mat_age_distribution}); $65\%$ were male and $35\%$ female. Each participant answered questions on $20$ attributes listed in Tab.\ref{tab:sup-mat_userstudy_attributes}. No monetary compensation was provided for the participants. We computed R-SAD and RA-CLIPScore, along with $9$ widely used evaluation metrics, using generated images from each model and real ImageNet samples for the selected classes ($1,000$ samples per class). Unlike other metrics, LPIPS~\cite{zhang2018unreasonable} does not use the training dataset as a reference. Nonetheless, as it measures perceptual diversity, we include it for comparison. For each class, we determine whether model A or B is preferable according to each metric. Human preference is determined by majority voting with an overall agreement of 72.9 \% among raters. 

We report Pearson Correlation Coefficient ($r$) between metric preference and human judgment in Tab.~\ref{tab:userstudy_results}. R-SaD shows high correlation with human preferences ($r$=1) , demonstrating its effectiveness in capturing diversity, including spatial diversity. RA-CLIPScore exhibits a lower correlation with human perception ($r$ = 0.6). We hypothesize that in some cases the global features might dominate and override the spatial diversity signals captured by the local features, thus RA-CLIPScore is not as sensitive as R-SaD for spatial diversity.

\begin{table*}[h] 
\centering  
\begin{tabular}{@{}l@{}} 
\toprule 
\textbf{Attributes} \\
\midrule 
\begin{tabular}[t]{@{}l@{}}
Burrito, Carton, Shopping cart, School bus, Frying pan, Broccoli, \\
Church, iPod, Balloon, Rotary dial telephone, Leatherback sea turtle, \\
Computer mouse, Flute, Spaghetti squash, Electric guitar, Paddle wheel, \\
Clown fish, Computer keyboard, Remote control, Radio telescope
\end{tabular} \\
\bottomrule 
\end{tabular} 
\caption{\small{$20$ attributes used for user study.}}
\label{tab:sup-mat_userstudy_attributes} 
\end{table*}

%% file: checklist.tex
\section*{NeurIPS Paper Checklist}

\begin{enumerate}

\item {\bf Claims}
    \item[] Question: Do the main claims made in the abstract and introduction accurately reflect the paper's contributions and scope?
    \item[] Answer: \answerYes{} 
    \item[] Justification: We state our contributions in the abstract and summarize them in the final paragraph of the Introduction. 

\item {\bf Limitations}
    \item[] Question: Does the paper discuss the limitations of the work performed by the authors?
    \item[] Answer: \answerYes{} 
    \item[] Justification: We discuss the limitations and future directions in Sec.\ref{sec:conclusions}.

\item {\bf Theory assumptions and proofs}
    \item[] Question: For each theoretical result, does the paper provide the full set of assumptions and a complete (and correct) proof?
    \item[] Answer: \answerYes{} 
    \item[] Justification: We provide a theoretical proof in Sec.\ref{subsec:sup-mat_theoretical_proof}.

    \item {\bf Experimental result reproducibility}
    \item[] Question: Does the paper fully disclose all the information needed to reproduce the main experimental results of the paper to the extent that it affects the main claims and/or conclusions of the paper (regardless of whether the code and data are provided or not)?
    \item[] Answer: \answerYes{} 
    \item[] Justification: We provide detailed experimental settings in Sec.~\ref{sec:experiments} and Supp.~\ref{sec:sup-mat_exp}, and release our code through an anonymous github repository for review.

\item {\bf Open access to data and code}
    \item[] Question: Does the paper provide open access to the data and code, with sufficient instructions to faithfully reproduce the main experimental results, as described in supplemental material?
    \item[] Answer: \answerYes{} 
    \item[] Justification: We release our code through an anonymous github repository for review.

\item {\bf Experimental setting/details}
    \item[] Question: Does the paper specify all the training and test details (e.g., data splits, hyperparameters, how they were chosen, type of optimizer) necessary to understand the results?
    \item[] Answer: \answerYes{} 
    \item[] Justification: We provide detailed experimental settings in Sec.~\ref{sec:experiments} and Supp.~\ref{sec:sup-mat_exp}.

\item {\bf Experiment statistical significance}
    \item[] Question: Does the paper report error bars suitably and correctly defined or other appropriate information about the statistical significance of the experiments?
    \item[] Answer: \answerNo{} 
    \item[] Justification: Our work focuses on the evaluation metric itself, rerunning the evaluation with different random seeds does not affect the results as long as the same generated image set is evaluated.

\item {\bf Experiments compute resources}
    \item[] Question: For each experiment, does the paper provide sufficient information on the computer resources (type of compute workers, memory, time of execution) needed to reproduce the experiments?
    \item[] Answer: \answerYes{} 
    \item[] Justification: We state our computer resources in Sec.\ref{sec:experiments}.
    
\item {\bf Code of ethics}
    \item[] Question: Does the research conducted in the paper conform, in every respect, with the NeurIPS Code of Ethics \url{https://neurips.cc/public/EthicsGuidelines}?
    \item[] Answer: \answerYes{} 
    \item[] Justification: We confirm that our work complies with NeurIPS Code of Ethics.

\item {\bf Broader impacts}
    \item[] Question: Does the paper discuss both potential positive societal impacts and negative societal impacts of the work performed?
    \item[] Answer: \answerNo{} 
    \item[] Justification: We do not propose a new generative model or technique intended for broad public deployment. Instead, our work introduces an evaluation paradigm for analyzing generative models, which we believe is unlikely to have direct negative societal impact.
    
\item {\bf Safeguards}
    \item[] Question: Does the paper describe safeguards that have been put in place for responsible release of data or models that have a high risk for misuse (e.g., pre-trained language models, image generators, or scraped datasets)?
    \item[] Answer: \answerNA{} 
    \item[] Justification: Our work focuses on evaluation metrics and does not introduce new generative models or datasets with elevated misuse risks. We use publicly available datasets and official pretrained checkpoints released by the original authors. Therefore, no additional safeguards for model or data release are required beyond those already provided by the original sources.

\item {\bf Licenses for existing assets}
    \item[] Question: Are the creators or original owners of assets (e.g., code, data, models), used in the paper, properly credited and are the license and terms of use explicitly mentioned and properly respected?
    \item[] Answer: \answerYes{} 
    \item[] Justification: We properly credit all datasets, pretrained models, and codebases used in this work through citations and references to the original publications and repositories. We use publicly available assets in accordance with their respective licenses and terms of use.

\item {\bf New assets}
    \item[] Question: Are new assets introduced in the paper well documented and is the documentation provided alongside the assets?
    \item[] Answer: \answerYes{} 
    \item[] Justification: We provide documentation for the proposed metric, including implementation details, experimental settings, and usage descriptions in both the main paper, supplementary material and code. The code is released through an anonymous repository to support review.

\item {\bf Crowdsourcing and research with human subjects}
    \item[] Question: For crowdsourcing experiments and research with human subjects, does the paper include the full text of instructions given to participants and screenshots, if applicable, as well as details about compensation (if any)? 
    \item[] Answer: \answerYes{} 
    \item[] Justification: We provide detailed descriptions of the user study, including participant instructions and screenshots in Supp.\ref{subsec:sup-mat_userstudy}. No monetary compensation was provided to participants.

\item {\bf Institutional review board (IRB) approvals or equivalent for research with human subjects}
    \item[] Question: Does the paper describe potential risks incurred by study participants, whether such risks were disclosed to the subjects, and whether Institutional Review Board (IRB) approvals (or an equivalent approval/review based on the requirements of your country or institution) were obtained?
    \item[] Answer: \answerNA{} 
    \item[] Justification: The user study in our work involves only perceptual comparison of generated and real images and does not collect sensitive personal information or involve vulnerable populations. We do not anticipate significant risks to participants beyond those encountered in everyday computer use. Participation was voluntary, and participants were informed about the purpose and procedure of the study. According to our institution’s guidelines, this study did not require formal IRB or equivalent ethical approval.

\item {\bf Declaration of LLM usage}
    \item[] Question: Does the paper describe the usage of LLMs if it is an important, original, or non-standard component of the core methods in this research? Note that if the LLM is used only for writing, editing, or formatting purposes and does \emph{not} impact the core methodology, scientific rigor, or originality of the research, declaration is not required.
    \item[] Answer: \answerNA{} 
    \item[] Justification: LLMs were used only for text editing and grammar checking, and did not contribute to the core methodology, experiments, or scientific findings of this work.

\end{enumerate}

%% file: main.bib
@String(CVPR  = {IEEE Conf. Comput. Vis. Pattern Recog.})

@String(ICCV  = {Int. Conf. Comput. Vis.})

@String(AAAI  = {AAAI})

@String(ICASSP=	{ICASSP})

@String(CVPR  = {CVPR})

@String(ICCV  = {ICCV})

@inproceedings{Kingma2013auto,
  title        = {Auto-Encoding Variational Bayes},
  author       = {Diederik P. Kingma and Max Welling},
  booktitle    = {International Conference on Learning Representations},
  year         = {2014}
}

@inproceedings{karras2019style,
  title={A style-based generator architecture for generative adversarial networks},
  author={Karras, Tero and Laine, Samuli and Aila, Timo},
  booktitle={Proceedings of the IEEE/CVF conference on computer vision and pattern recognition},
  pages={4401--4410},
  year={2019}
}

@inproceedings{karras2020analyzing,
  title={Analyzing and improving the image quality of stylegan},
  author={Karras, Tero and Laine, Samuli and Aittala, Miika and Hellsten, Janne and Lehtinen, Jaakko and Aila, Timo},
  booktitle={Proceedings of the IEEE/CVF conference on computer vision and pattern recognition},
  pages={8110--8119},
  year={2020}
}

@inproceedings{karras2021alias,
  title={Alias-free generative adversarial networks},
  author={Karras, Tero and Aittala, Miika and Laine, Samuli and H{\"a}rk{\"o}nen, Erik and Hellsten, Janne and Lehtinen, Jaakko and Aila, Timo},
  booktitle={Advances in Neural Information Processing Systems},
  year={2021}
}

@inproceedings{sauer2021projected,
  title={Projected gans converge faster},
  author={Sauer, Axel and Chitta, Kashyap and M{\"u}ller, Jens and Geiger, Andreas},
  booktitle={Advances in Neural Information Processing Systems},
  year={2021}
}

@inproceedings{ho2020denoising,
  title={Denoising diffusion probabilistic models},
  author={Ho, Jonathan and Jain, Ajay and Abbeel, Pieter},
  booktitle={Advances in Neural Information Processing Systems},
  year={2020}
}

@inproceedings{song2020denoising,
title={Denoising Diffusion Implicit Models},
author={Jiaming Song and Chenlin Meng and Stefano Ermon},
booktitle={International Conference on Learning Representations},
year={2021}
}

@InProceedings{nichol2021glide,
  title = 	 {{GLIDE}: Towards Photorealistic Image Generation and Editing with Text-Guided Diffusion Models},
  author =       {Nichol, Alexander Quinn and Dhariwal, Prafulla and Ramesh, Aditya and Shyam, Pranav and Mishkin, Pamela and Mcgrew, Bob and Sutskever, Ilya and Chen, Mark},
  booktitle = 	 {Proceedings of the 39th International Conference on Machine Learning},
  pages = 	 {16784--16804},
  year = 	 {2022},
  volume = 	 {162},
  series = 	 {Proceedings of Machine Learning Research},
  month = 	 {17--23 Jul},
  publisher =    {PMLR}
}

@inproceedings{rezende2015variational,
  title={Variational inference with normalizing flows},
  author={Rezende, Danilo and Mohamed, Shakir},
  booktitle={International conference on machine learning},
  pages={1530--1538},
  year={2015},
  organization={PMLR}
}

@inproceedings{dinh2016density,
  title     = {Density Estimation Using Real NVP},
  author    = {Laurent Dinh and Jascha Sohl{-}Dickstein and Samy Bengio},
  booktitle = {International Conference on Learning Representations},
  year      = {2017}
}

@inproceedings{heusel2017gans,
  title={Gans trained by a two time-scale update rule converge to a local nash equilibrium},
  author={Heusel, Martin and Ramsauer, Hubert and Unterthiner, Thomas and Nessler, Bernhard and Hochreiter, Sepp},
  booktitle={Advances in Neural Information Processing Systems},
  year={2017}
}

@inproceedings{salimans2016improved,
  title={Improved techniques for training gans},
  author={Salimans, Tim and Goodfellow, Ian and Zaremba, Wojciech and Cheung, Vicki and Radford, Alec and Chen, Xi},
  booktitle={Advances in Neural Information Processing Systems},
  year={2016}
}

@inproceedings{sutherland2018demystifying,
  title={Demystifying mmd gans},
  author={Sutherland, JD and Arbel, Michael and Gretton, Arthur},
  booktitle={International conference for learning representations},
  volume={6},
  year={2018}
}

@inproceedings{kynkaanniemi2019improved,
  title={Improved precision and recall metric for assessing generative models},
  author={Kynk{\"a}{\"a}nniemi, Tuomas and Karras, Tero and Laine, Samuli and Lehtinen, Jaakko and Aila, Timo},
  booktitle={Advances in Neural Information Processing Systems},
  year={2019}
}

@inproceedings{sajjadi2018assessing,
  title={Assessing generative models via precision and recall},
  author={Sajjadi, Mehdi SM and Bachem, Olivier and Lucic, Mario and Bousquet, Olivier and Gelly, Sylvain},
  booktitle={Advances in Neural Information Processing Systems},
  year={2018}
}

@inproceedings{naeem2020reliable,
  title={Reliable fidelity and diversity metrics for generative models},
  author={Naeem, Muhammad Ferjad and Oh, Seong Joon and Uh, Youngjung and Choi, Yunjey and Yoo, Jaejun},
  booktitle={International conference on machine learning},
  pages={7176--7185},
  year={2020},
  organization={PMLR}
}

@inproceedings{radford2021learning,
  title={Learning transferable visual models from natural language supervision},
  author={Radford, Alec and Kim, Jong Wook and Hallacy, Chris and Ramesh, Aditya and Goh, Gabriel and Agarwal, Sandhini and Sastry, Girish and Askell, Amanda and Mishkin, Pamela and Clark, Jack and others},
  booktitle={International conference on machine learning},
  pages={8748--8763},
  year={2021},
  organization={PmLR}
}

@inproceedings{hessel2021clipscore,
  title={{CLIPScore:} A Reference-free Evaluation Metric for Image Captioning},
  author={Hessel, Jack and Holtzman, Ari and Forbes, Maxwell and Bras, Ronan Le and Choi, Yejin},
  booktitle={EMNLP},
  year={2021}
}

@InProceedings{kim2023attribute,
  title = 	 {Attribute Based Interpretable Evaluation Metrics for Generative Models},
  author =       {Kim, Dongkyun and Kwon, Mingi and Uh, Youngjung},
  booktitle = 	 {Proceedings of the 41st International Conference on Machine Learning},
  pages = 	 {24271--24293},
  year = 	 {2024},
  volume = 	 {235},
  series = 	 {Proceedings of Machine Learning Research},
  month = 	 {21--27 Jul},
  publisher =    {PMLR}
}

@inproceedings{kirillov2023segment,
  title={Segment anything},
  author={Kirillov, Alexander and Mintun, Eric and Ravi, Nikhila and Mao, Hanzi and Rolland, Chloe and Gustafson, Laura and Xiao, Tete and Whitehead, Spencer and Berg, Alexander C and Lo, Wan-Yen and others},
  booktitle={Proceedings of the IEEE/CVF international conference on computer vision},
  pages={4015--4026},
  year={2023}
}

@inproceedings{szegedy2016rethinking,
  title={Rethinking the inception architecture for computer vision},
  author={Szegedy, Christian and Vanhoucke, Vincent and Ioffe, Sergey and Shlens, Jon and Wojna, Zbigniew},
  booktitle={Proceedings of the IEEE conference on computer vision and pattern recognition},
  pages={2818--2826},
  year={2016}
}

@article{ghiasi2022vision,
  title={What do vision transformers learn? a visual exploration},
  author={Ghiasi, Amin and Kazemi, Hamid and Borgnia, Eitan and Reich, Steven and Shu, Manli and Goldblum, Micah and Wilson, Andrew Gordon and Goldstein, Tom},
  journal={arXiv preprint arXiv:2212.06727},
  year={2022}
}

@inproceedings{xu2022multi,
  title={Multi-class token transformer for weakly supervised semantic segmentation},
  author={Xu, Lian and Ouyang, Wanli and Bennamoun, Mohammed and Boussaid, Farid and Xu, Dan},
  booktitle={Proceedings of the IEEE/CVF conference on computer vision and pattern recognition},
  pages={4310--4319},
  year={2022}
}

@inproceedings{gao2021ts,
  title={Ts-cam: Token semantic coupled attention map for weakly supervised object localization},
  author={Gao, Wei and Wan, Fang and Pan, Xingjia and Peng, Zhiliang and Tian, Qi and Han, Zhenjun and Zhou, Bolei and Ye, Qixiang},
  booktitle={Proceedings of the IEEE/CVF international conference on computer vision},
  pages={2886--2895},
  year={2021}
}

@inproceedings{liu2015faceattributes,
  title = {Deep Learning Face Attributes in the Wild},
  author = {Liu, Ziwei and Luo, Ping and Wang, Xiaogang and Tang, Xiaoou},
  booktitle = {Proceedings of International Conference on Computer Vision (ICCV)},
  month = {December},
  year = {2015} 
}

@conference{kwon2022diffusion,
title = {Diffusion-based image translation using disentangled style and content representation},
author = {Gihyun Kwon and Ye, \{Jong Chul\}},
booktitle = {International Conference on Learning Representations},
year = {2023}
}

@inproceedings{he2016deep,
  title={Deep residual learning for image recognition},
  author={He, Kaiming and Zhang, Xiangyu and Ren, Shaoqing and Sun, Jian},
  booktitle={Proceedings of the IEEE conference on computer vision and pattern recognition},
  pages={770--778},
  year={2016}
}

@inproceedings{dosovitskiy2020image,
  title     = {An Image is Worth 16×16 Words: Transformers for Image Recognition at Scale},
  author    = {Alexey Dosovitskiy and Lucas Beyer and Alexander Kolesnikov and Dirk Weissenborn and Xiaohua Zhai and Thomas Unterthiner and Mostafa Dehghani and Matthias Minderer and Georg Heigold and Sylvain Gelly and Jakob Uszkoreit and Neil Houlsby},
  booktitle = {International Conference on Learning Representations},
  year      = {2021}
}

@inproceedings{lin2024tagclip,
  title={Tagclip: A local-to-global framework to enhance open-vocabulary multi-label classification of clip without training},
  author={Lin, Yuqi and Chen, Minghao and Zhang, Kaipeng and Li, Hengjia and Li, Mingming and Yang, Zheng and Lv, Dongqin and Lin, Binbin and Liu, Haifeng and Cai, Deng},
  booktitle={Proceedings of the AAAI Conference on Artificial Intelligence},
  volume={38},
  pages={3513--3521},
  year={2024}
}

@inproceedings{sun2022dualcoop,
  title={Dualcoop: Fast adaptation to multi-label recognition with limited annotations},
  author={Sun, Ximeng and Hu, Ping and Saenko, Kate},
  booktitle={Advances in Neural Information Processing Systems},
  year={2022}
}

@inproceedings{paszke2019pytorch,
  title={Pytorch: An imperative style, high-performance deep learning library},
  author={Paszke, Adam and Gross, Sam and Massa, Francisco and Lerer, Adam and Bradbury, James and Chanan, Gregory and Killeen, Trevor and Lin, Zeming and Gimelshein, Natalia and Antiga, Luca and others},
  booktitle={Advances in Neural Information Processing Systems},
  year={2019}
}

@inproceedings{Rombach2022ldm,
  title     = {High-Resolution Image Synthesis with Latent Diffusion Models},
  author    = {Rombach, Robin and Blattmann, Andreas and Lorenz, Dominik and Esser, Patrick and Ommer, Björn},
  booktitle = {IEEE/CVF Conference on Computer Vision and Pattern Recognition (CVPR)},
  year      = {2022},
  pages     = {10684--10695}
}

@inproceedings{deng2009imagenet,
  title={Imagenet: A large-scale hierarchical image database},
  author={Deng, Jia and Dong, Wei and Socher, Richard and Li, Li-Jia and Li, Kai and Fei-Fei, Li},
  booktitle={2009 IEEE conference on computer vision and pattern recognition},
  pages={248--255},
  year={2009},
  organization={Ieee}
}

@inproceedings{sauer2022stylegan,
  title={Stylegan-xl: Scaling stylegan to large diverse datasets},
  author={Sauer, Axel and Schwarz, Katja and Geiger, Andreas},
  booktitle={ACM SIGGRAPH 2022 conference proceedings},
  pages={1--10},
  year={2022}
}

@inproceedings{brock2018large,
title={Large Scale {GAN} Training for High Fidelity Natural Image Synthesis},
author={Andrew Brock and Jeff Donahue and Karen Simonyan},
booktitle={International Conference on Learning Representations},
year={2019}
}

@inproceedings{elizalde2023clap,
  title={Clap learning audio concepts from natural language supervision},
  author={Elizalde, Benjamin and Deshmukh, Soham and Al Ismail, Mahmoud and Wang, Huaming},
  booktitle={ICASSP 2023-2023 IEEE International Conference on Acoustics, Speech and Signal Processing (ICASSP)},
  pages={1--5},
  year={2023},
  organization={IEEE}
}

@inproceedings{yuksekgonul2023and,
  title     = {When and Why Vision-Language Models Behave Like Bags-of-Words, and What to Do About It?},
  author    = {Mert Yuksekgonul and Federico Bianchi and Pratyusha Kalluri and Dan Jurafsky and James Y. Zou},
  booktitle = {International Conference on Learning Representations},
  year      = {2023}
}

@inproceedings{jayasumana2024rethinking,
  title={Rethinking fid: Towards a better evaluation metric for image generation},
  author={Jayasumana, Sadeep and Ramalingam, Srikumar and Veit, Andreas and Glasner, Daniel and Chakrabarti, Ayan and Kumar, Sanjiv},
  booktitle={IEEE/CVF Conference on Computer Vision and Pattern Recognition},
  year={2024}
}

@inproceedings{kynkaanniemi2022role,
  title     = {The Role of ImageNet Classes in Fréchet Inception Distance},
  author    = {Tuomas Kynk\"a\"anniemi and Tero Karras and Miika Aittala and Timo Aila and Jaakko Lehtinen},
  booktitle = {International Conference on Learning Representations},
  year      = {2023}
}

@article{betzalel2022study,
  title={A study on the evaluation of generative models},
  author={Betzalel, Eyal and Penso, Coby and Navon, Aviv and Fetaya, Ethan},
  journal={arXiv preprint arXiv:2206.10935},
  year={2022}
}

@inproceedings{morozov2021self,
  title={On self-supervised image representations for gan evaluation},
  author={Morozov, Stanislav and Voynov, Andrey and Babenko, Artem},
  booktitle={International Conference on Learning Representations},
  year={2021}
}

@inproceedings{parmar2022aliased,
  title={On aliased resizing and surprising subtleties in gan evaluation},
  author={Parmar, Gaurav and Zhang, Richard and Zhu, Jun-Yan},
  booktitle={Proceedings of the IEEE/CVF conference on computer vision and pattern recognition},
  pages={11410--11420},
  year={2022}
}

@article{zhou2021deepvit,
  title={DeepViT: Towards Deeper Vision Transformer},
  author={Zhou, Daquan and Kang, Bingyi and Jin, Xiaojie and Yang, Linjie and Lian, Xiaochen and Hou, Qibin and Feng, Jiashi},
  journal={arXiv preprint arXiv:2103.11886},
  year={2021}
}

@inproceedings{uchida2025objectness,
  title     = {Objectness Similarity: Capturing Object-Level Fidelity in 3D Scene Evaluation},
  author    = {Yuiko Uchida and Ren Togo and Keisuke Maeda and Takahiro Ogawa and Miki Haseyama},
  booktitle = {Proceedings of the IEEE/CVF International Conference on Computer Vision (ICCV) 2025 Workshops},
  year      = {2025}
}

@article{ghosh2023geneval,
  title={Geneval: An object-focused framework for evaluating text-to-image alignment},
  author={Ghosh, Dhruba and Hajishirzi, Hannaneh and Schmidt, Ludwig},
  journal={Advances in Neural Information Processing Systems},
  volume={36},
  pages={52132--52152},
  year={2023}
}

@inproceedings{lin2014microsoft,
  title={Microsoft coco: Common objects in context},
  author={Lin, Tsung-Yi and Maire, Michael and Belongie, Serge and Hays, James and Perona, Pietro and Ramanan, Deva and Doll{\'a}r, Piotr and Zitnick, C Lawrence},
  booktitle={European conference on computer vision},
  pages={740--755},
  year={2014},
  organization={Springer}
}

@misc{mukherjee_furniture_dataset,
  author       = {Mukherjee, Uday Sankar},
  title        = {Furniture Image Dataset},
  year         = {2020},
  publisher    = {Kaggle},
  howpublished = {\url{https://www.kaggle.com/datasets/udaysankarmukherjee/furniture-image-dataset}}
}

@inproceedings{zhang2018unreasonable,
  title={The unreasonable effectiveness of deep features as a perceptual metric},
  author={Zhang, Richard and Isola, Phillip and Efros, Alexei A and Shechtman, Eli and Wang, Oliver},
  booktitle={Proceedings of the IEEE conference on computer vision and pattern recognition},
  pages={586--595},
  year={2018}
}

@inproceedings{zhai2023sigmoid,
  title={Sigmoid loss for language image pre-training},
  author={Zhai, Xiaohua and Mustafa, Basil and Kolesnikov, Alexander and Beyer, Lucas},
  booktitle={Proceedings of the IEEE/CVF international conference on computer vision},
  pages={11975--11986},
  year={2023}
}

@inproceedings{caron2021emerging,
  title={Emerging properties in self-supervised vision transformers},
  author={Caron, Mathilde and Touvron, Hugo and Misra, Ishan and J{\'e}gou, Herv{\'e} and Mairal, Julien and Bojanowski, Piotr and Joulin, Armand},
  booktitle={Proceedings of the IEEE/CVF international conference on computer vision},
  pages={9650--9660},
  year={2021}
}

@article{dhariwal2021diffusion,
  title={Diffusion models beat gans on image synthesis},
  author={Dhariwal, Prafulla and Nichol, Alexander},
  journal={Advances in neural information processing systems},
  volume={34},
  pages={8780--8794},
  year={2021}
}

@inproceedings{lorenz2022high,
  title={High-resolution image synthesis with latent diffusion models},
  author={Lorenz, D and Esser, Patrick and Ommer, Bjorn},
  booktitle={Proceedings of the IEEE/CVF Conference on Computer Vision and Pattern Recognition},
  year={2022}
}

@article{ramesh2022hierarchical,
  title={Hierarchical text-conditional image generation with clip latents},
  author={Ramesh, Aditya and Dhariwal, Prafulla and Nichol, Alex and Chu, Casey and Chen, Mark},
  journal={arXiv preprint arXiv:2204.06125},
  volume={1},
  number={2},
  pages={3},
  year={2022}
}

@inproceedings{abdelfattah2023cdul,
  title={Cdul: Clip-driven unsupervised learning for multi-label image classification},
  author={Abdelfattah, Rabab and Guo, Qing and Li, Xiaoguang and Wang, Xiaofeng and Wang, Song},
  booktitle={Proceedings of the IEEE/CVF international conference on computer vision},
  pages={1348--1357},
  year={2023}
}

@book{cohen1988statistical,
  title     = {Statistical Power Analysis for the Behavioral Sciences},
  author    = {Cohen, Jacob},
  edition   = {2},
  publisher = {Lawrence Erlbaum Associates},
  year      = {1988},
  address   = {Hillsdale, NJ}
}

@inproceedings{li2024genai,
  title={GenAI-bench: A holistic benchmark for compositional text-to-visual generation},
  author={Li, Baiqi and Lin, Zhiqiu and Pathak, Deepak and Li, Jiayao Emily and Xia, Xide and Neubig, Graham and Zhang, Pengchuan and Ramanan, Deva},
  booktitle={Synthetic Data for Computer Vision Workshop@ CVPR 2024},
  year={2024}
}

@article{huang2023t2i,
  title={T2i-compbench: A comprehensive benchmark for open-world compositional text-to-image generation},
  author={Huang, Kaiyi and Sun, Kaiyue and Xie, Enze and Li, Zhenguo and Liu, Xihui},
  journal={Advances in Neural Information Processing Systems},
  volume={36},
  pages={78723--78747},
  year={2023}
}

@inproceedings{hu2023tifa,
  title={Tifa: Accurate and interpretable text-to-image faithfulness evaluation with question answering},
  author={Hu, Yushi and Liu, Benlin and Kasai, Jungo and Wang, Yizhong and Ostendorf, Mari and Krishna, Ranjay and Smith, Noah A},
  booktitle={2023 ieee/cvf international conference on computer vision (iccv)},
  pages={20349--20360},
  year={2023},
  organization={IEEE}
}

@inproceedings{park2025know,
  title={Know “no” better: A data-driven approach for enhancing negation awareness in clip},
  author={Park, Junsung and Lee, Jungbeom and Song, Jongyoon and Yu, Sangwon and Jung, Dahuin and Yoon, Sungroh},
  booktitle={2025 IEEE/CVF International Conference on Computer Vision (ICCV)},
  pages={2825--2835},
  year={2025},
  organization={IEEE}
}

@inproceedings{nie2024out,
  title={Out-of-distribution detection with negative prompts},
  author={Nie, Jun and Zhang, Yonggang and Fang, Zhen and Liu, Tongliang and Han, Bo and Tian, Xinmei},
  booktitle={International Conference on Learning Representations},
  volume={2024},
  pages={46394--46413},
  year={2024}
}

@inproceedings{yao2011human,
  title={Human action recognition by learning bases of action attributes and parts.},
  author={Yao, Bangpeng and Jiang, Xiaoye and Khosla, Aditya and Lin, Andy Lai and Guibas, Leonidas J and Fei-Fei, Li},
  booktitle={ICCV},
  volume={2},
  pages={7},
  year={2011},
  organization={Barcelona, Spain}
}

@article{zhou2022learning,
  title={Learning to prompt for vision-language models},
  author={Zhou, Kaiyang and Yang, Jingkang and Loy, Chen Change and Liu, Ziwei},
  journal={International journal of computer vision},
  volume={130},
  number={9},
  pages={2337--2348},
  year={2022},
  publisher={Springer}
}
